\documentclass[journal]{IEEEtran}
\usepackage{amsmath,amsfonts}
\usepackage{algorithmic}
\usepackage{algorithm}
\usepackage{array}
\usepackage[caption=false,font=normalsize,labelfont=sf,textfont=sf]{subfig}
\usepackage{textcomp}
\usepackage{stfloats}
\usepackage{url}
\usepackage{verbatim}
\usepackage{graphicx}
\usepackage{cite}
\usepackage{lettrine}
\usepackage{hyperref}
\usepackage{multirow}
\usepackage{diagbox}
\begin{document}
\title{Image Copy-Move Forgery Detection and Localization Scheme: How to Avoid Missed Detection and False Alarm}

\author{Li Jiang, Zhaowei Lu, Yuebing Gao, Yifan Wang
		\thanks{Corresponding author: Zhaowei Lu}
		\thanks{L. Jiang and  Z. Lu et al. are with the School of Electrical and Information Engineering, Zhengzhou University, Zhengzhou, China (ieljiang@zzu.edu.cn, luzhaoweizzu@163.com).}
		}



\maketitle

\begin{abstract}
Image copy-move is an operation that replaces one part of the image with another part of the same image, which can be used for illegal purposes due to the potential semantic changes. Recent studies have shown that keypoint-based algorithms achieved excellent and robust localization performance even when small or smooth tampered areas were involved. However, when the input image is low-resolution, most existing keypoint-based algorithms are difficult to generate sufficient keypoints, resulting in more missed detections. In addition, existing algorithms are usually unable to distinguish between Similar but Genuine Objects (SGO) images and tampered images, resulting in more false alarms. This is mainly due to the lack of further verification of local homography matrix in forgery localization stage. To tackle these problems, this paper firstly proposes an excessive keypoint extraction strategy to overcome missed detection. Subsequently, a group matching algorithm is used to speed up the matching of excessive keypoints. Finally, a new iterative forgery localization algorithm is introduced to quickly form pixel-level localization results while ensuring a lower false alarm. Extensive experimental results show that our scheme has superior performance than state-of-the-art algorithms in overcoming missed detection and false alarm.
Our code is available at https://github.com/LUZW1998/CMFDL.
\end{abstract}


\section{Introduction}\label{introduction}
\lettrine{D}IGITAL image editing technology has developed rapidly over the past few decades. This technology poses huge challenges to the authenticity and integrity of images. Once this technology is used for illegal purposes, it may have a huge negative impact on the public opinion.

Copy-move, splicing and inpainting are three common semantic tampering operations in image editing technology. Compared with the latter two, copy-move is difficult to detect inconsistencies due to the homogeneity of the operation content. Therefore, content retrieval techniques are often used to determine the presence of copy-move operations. The classical Copy-Move Forgery Detection and Localization (CMFDL) flowchart consists of three stages \cite{1}: 1) feature extraction, 2) feature matching and 3) post-processing (forgery localization). The main ideas are 1) designing suitable features and generation keypoints covering the entire image, 2) quickly searching for similar features (keypoints) and 3) converting similar keypoints into binary localization results.

Traditionally, there are two different branches for CMFDL \cite{1}: dense-field (block-based algorithms) \cite{2,3,4,5,6,7,8,9,10,11,12,13,28,30,35} and sparse-field (keypoint-based algorithms) \cite{14,15,16,17,18,19,20,21,22,23,24,25,26,27,29,31}. The main difference is that the block-based algorithms use pixels from the entire image for matching, while the keypoint-based algorithms select important pixels for matching through certain strategies. In general, computer vision classifies these strategies into two categories, corner points and blob points \cite{900}. Correspondingly, these features are also divided into corn features and blob features. In particular, blob points are generated in scale space, which allows scale information (variable window) to be considered in the feature representation. Therefore, blob features are more robust to scaling attack. keypoint-based algorithms are robust to geometric transformations primarily because they extensively utilize blob features.

In recent years, deep learning methods \cite{38,39,40} have rapidly developed in CMFDL. This is primarily due to the fact that deep networks can capture features more efficiently. The main differences between traditional algorithms and deep learning methods are as follows:
\begin{itemize}
	\item Deep learning methods usually utilize end-to-end network training for parameters; while traditional algorithms heavily rely on manual parameter setting.
	\item Deep learning methods can achieve multi-task forensics. For instance, they can detect copy-move operations while distinguishing between the source and target; however, traditional algorithms struggle to accomplish multi-task forensics.
\end{itemize}

Although deep learning methods have achieved great performance, they cannot directly process high-resolution images. Typically, high-resolution images need to be reduced in resolution before being fed into a deep network. This will lose some information \cite{27}, resulting in poor performance. Therefore, it is necessary to continue the development of traditional algorithms. Recent advances in traditional CMFDL have shown that great performance has been achieved for tampering involving small or smooth regions, especially keypoint-based algorithms. However, it is observed that keypoint-based algorithms still suffer from the following common problems:

\begin{itemize}
	\item The phenomenon of missed detections is severe in low-resolution images. This is mainly due to the fact that the complete semantics of low-resolution images has a smaller number of pixels. It is difficult for the keypoint detection algorithm to generate 4 matches in these semantic patches \cite{18} (the RANSAC estimation \cite{901} needs at least 4 correct matches).
	\item The phenomenon of false alarms is serious in SGO images \cite{17}. This is primarily due to the fact that most keypoint-based algorithms utilize less information to represent complete semantics. Under ideal scenarios, the tampered area can be determined by geometric relationships. However, once SGOs present in an image, the geometric relationship established by a small number of matches may be caused by self-similarity, resulting in false alarms.
\end{itemize}

It can be concluded that most of state-of-the-art keypoint-based frameworks lack further extraction of keypoint. As a consequence, these algorithms exhibit poor detection performance in low-resolution or SGO images. To tackle these problems, three stages of CMFDL are researched. The main contributions of this paper are as follows:
\begin{itemize}
	\item A feature extraction algorithm based on excessive keypoint strategy is proposed. The essence of the keypoint-based algorithm is analyzed to ensure that the complete semantic patches have a sufficient number of keypoints, which is conducive to reducing the missed detections of ICMFDL.
	\item A fast group matching algorithm is developed. In this stage, a grouping strategy that combines ANN and clustering is used for feature matching. Although it may lose some correct matches, the number of correct matches generated under the excessive keypoint strategy far exceeds that of existing algorithms.
	\item A new iterative forgery localization algorithm is improved. This stage firstly removes the models of the current iteration one by one, which helps to separate spatially closer models. Subsequently, the minimum number of inliers is used to distinguish the source of the local homography matrix. Finally, a robust grayscale statistics are exploited for dense-field localization.
\end{itemize}

The remainder of this paper is organized as follows. Related work is reported in Section \ref{related work}. Our proposed method is introduced in Section \ref{proposed method}. The experimental results of our method are presented in Section \ref{experiment results}, and a short conclusion
is finally drawn in Section \ref{conclusion}.

\section{Related work}\label{related work}
\subsection{Feature Extraction Stage}
Over the past decades, a significant amount of research has been concentrated on hand-crafted feature, mainly considering robustness against pre-processing attacks (geometric transformation) and post-processing attacks (such as brightness change, color reduction). These features can generally be divided into three categories in terms of describing the shape of the window: 1) square window, such as Discrete Cosine Transformation (DCT) \cite{2,10,11,13}, Histogram of Orientated Gradients (HoG) \cite{7}, color intensity \cite{9,35}, Discrete Wavelet Transform (DWT) \cite{11,13}; 2) circular window, such as Radon Transform (RT)\cite{3}, Zernike Moments (ZM) \cite{4}, Polar Cosine Transform (PCT) \cite{5,6,12,18}, Polar Complex Exponential Transform (PCET) \cite{8}, Bessel-Fourier Moments (BFM) \cite{25,27}; 3) variable window, such as Scale Invariant Feature Transform (SIFT) \cite{14,15,16,19,20,22,24,26}, Speeded-Up Robust Feature (SURF) \cite{17}, KAZA \cite{19}. Existing studies have indicated that SIFT algorithm \cite{902} exhibited excellent performance in computational efficiency and robustness to geometric transformations. Therefore, this paper develops a complete CMFDL framework using the SIFT algorithm. However, the classic SIFT algorithm usually fails to cover low-contrast (smooth) areas or struggles to generate sufficient keypoints when the semantic patch has few pixels. These problems have received widespread attention in the past decade.

\begin{figure}[ht]
	\centering
	\begin{tabular}{cc}
		\includegraphics[width=0.45\linewidth]{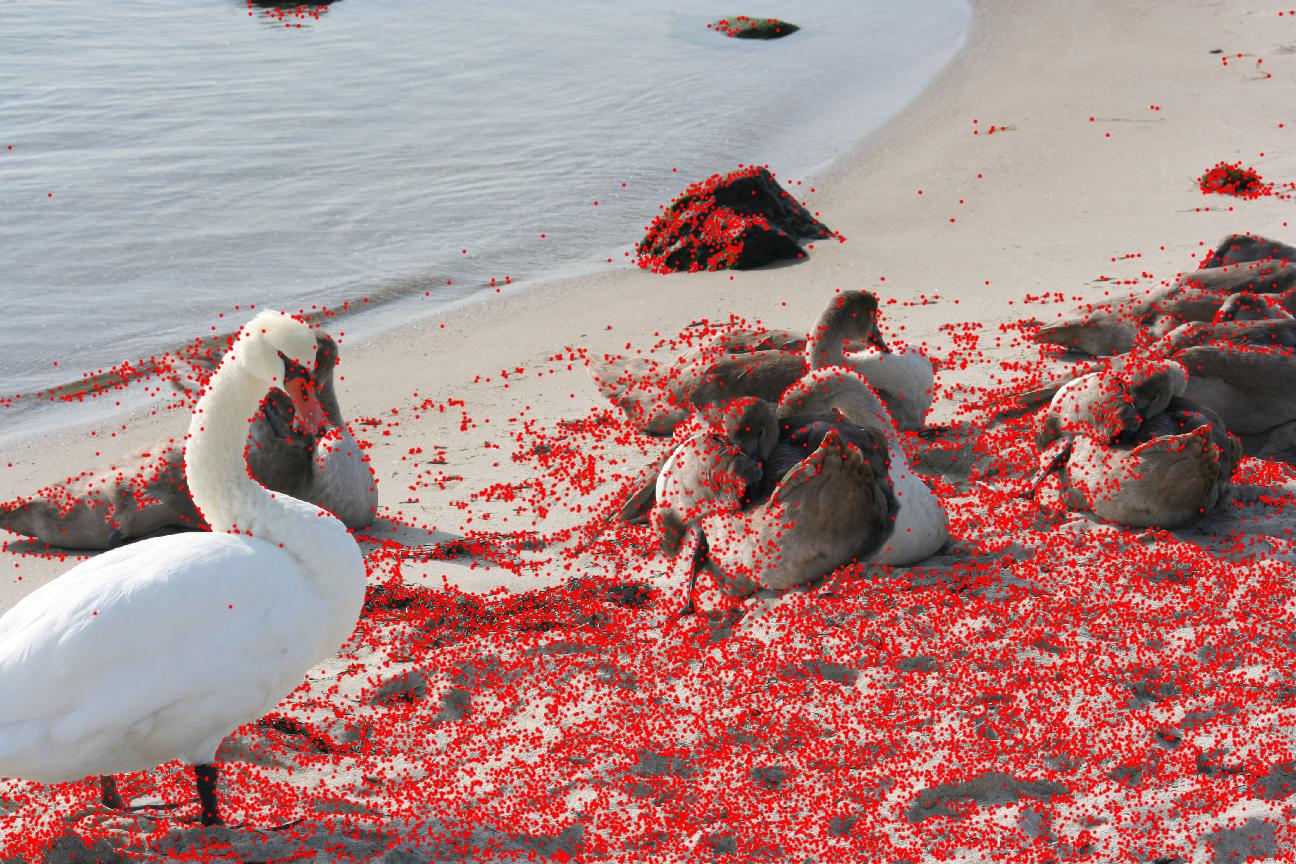} &
		\includegraphics[width=0.45\linewidth]{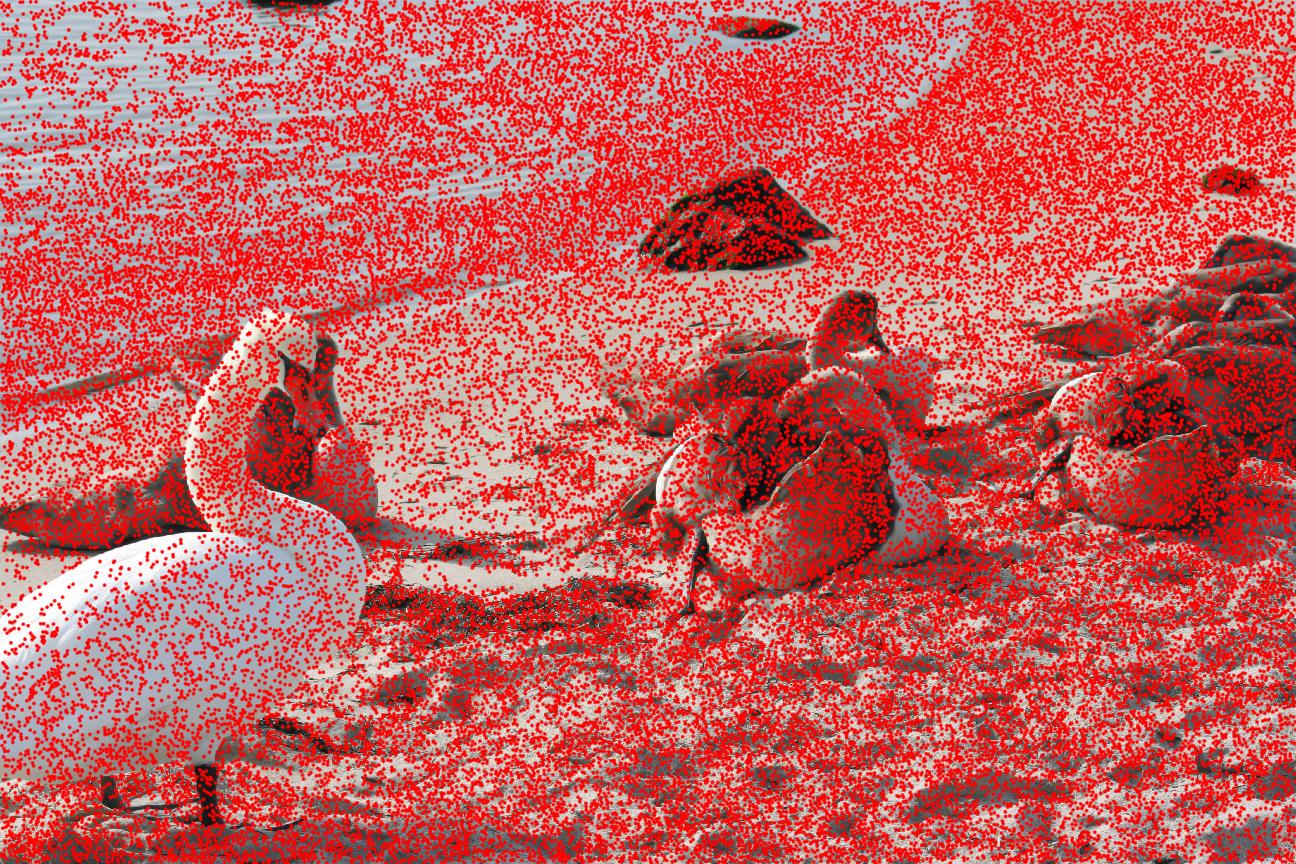} \\
		(a) & (b) \\
		\includegraphics[width=0.45\linewidth]{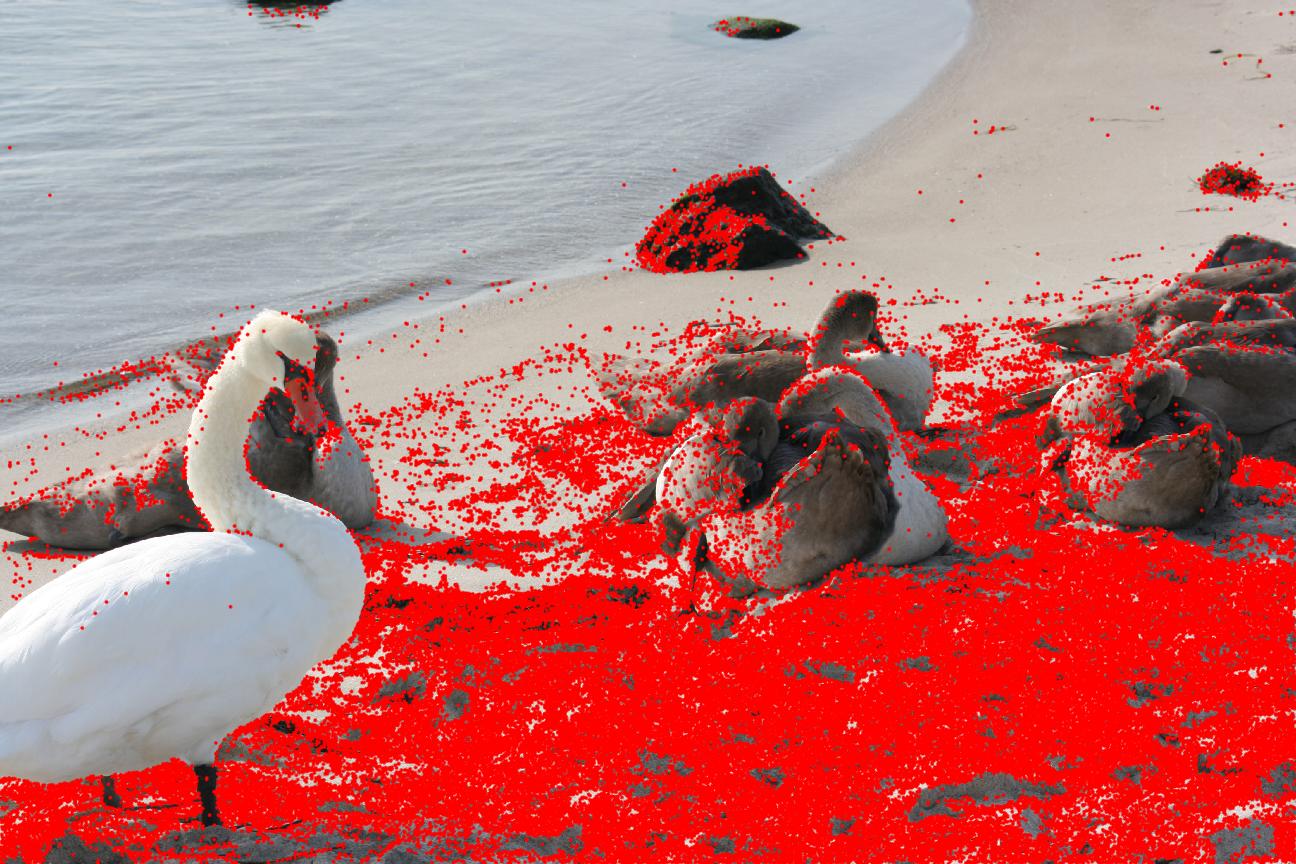} &
		\includegraphics[width=0.45\linewidth]{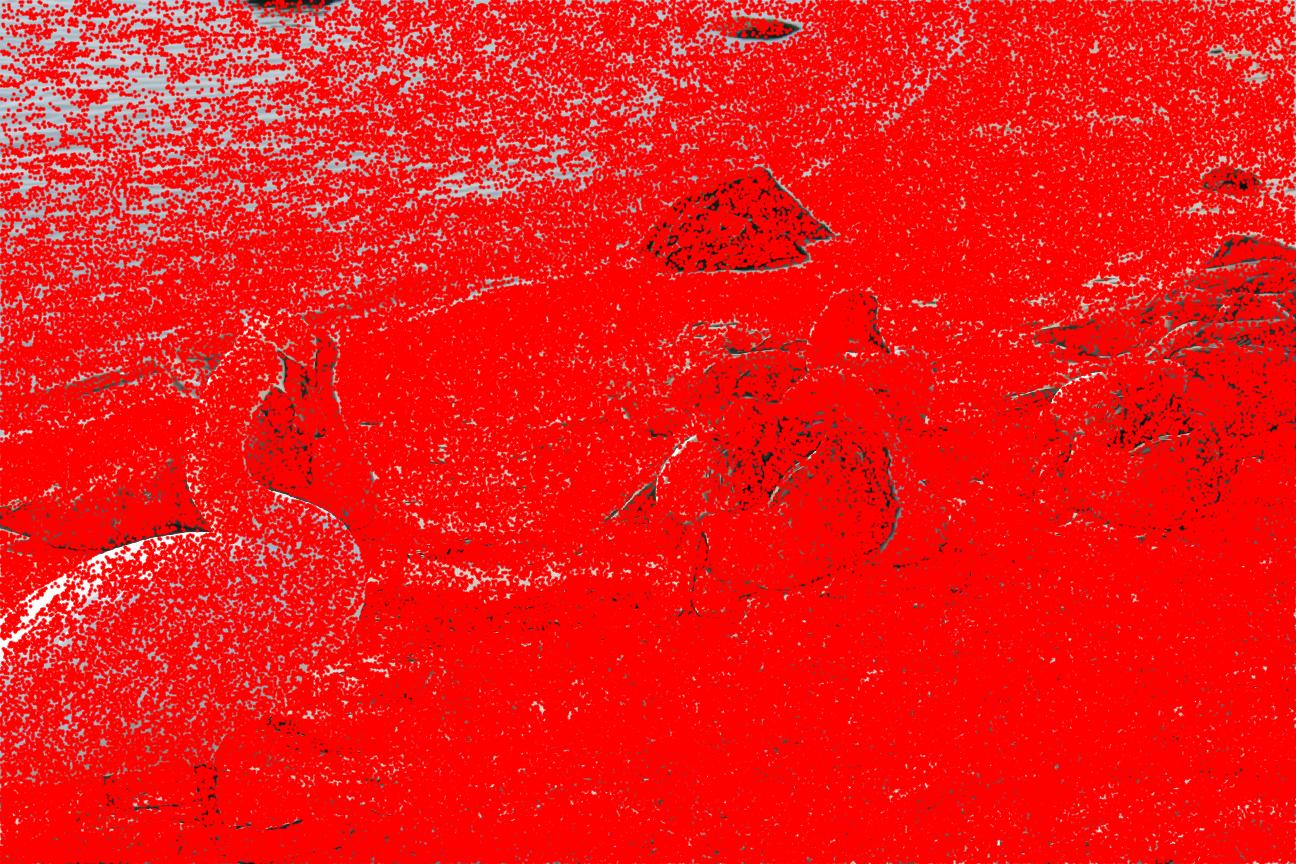} \\
		(c) & (d)
	\end{tabular}
	\caption{Keypoint detection using (a) classical contrast threshold; (b) reducing contrast threshold; (c) upsampling and classical contrast threshold; (d) upsampling and reducing contrast threshold.}
	\label{Fig1}
\end{figure}

As illustrated in Fig. \ref{Fig1}, the SIFT algorithm has two strategies commonly used to detect smooth or small tampered patches. The first strategy is to reduce the contrast threshold, which aims to control the generation of keypoints in low-contrast areas. For instance, Li \cite{22} trained contrast threshold on smooth semantic patches to ensure stable keypoints; Niu \cite{25} and Wang \cite{27} set the contrast threshold to 0 to obtain all discernible keypoints; Jin \cite{20} obtained stable keypoints from discernible keypoints through Non-Maximum Suppression (NMS). The second strategy is image upsampling, which aims to control the number of keypoints generated for a certain number of pixels (relative to the original image). For instance, Li \cite{22} used single upsampling to ensure a sufficient number of keypoints in small patches, and Wang \cite{27} normalized the long edge of the low-resolution image to 3000 pixels, without changing the aspect ratio.

In summary, most algorithms strive to use fewer keypoints to complete CMFDL, which may result in these algorithms lacking necessary verification information when facing SGO images. In this paper, a feature extraction algorithm based on the excessive keypoint strategy is proposed to increase the number of matches and ensure that forgery localization algorithm can further verify self-similarity or tampering.

\subsection{Feature Matching Stage}
The essence of feature matching is searching for the Nearest Neighbor (NN) of the query feature. Over the past few decades, matching accuracy and efficiency have received widespread attention in the CMFDL field.

To accurately identify NNs, there are two types of NN tests: absolute distance NN tests and relative distance NN tests. The absolute distance NN tests compute whether the absolute distance between features is less than a predefined threshold. These methods face several challenges: if a high threshold is used for matching, it may result in an increased number of false positives. Conversely, setting a low threshold could lead to missing some correct matches. To overcome these problems, Zandi \cite{28} proposed an adaptive threshold that adjusted the threshold by estimating the Standard Deviation (SD). The relative distance NN tests compute whether the ratio between the $ i $-th neighbor distance and the $ (i+1) $-th nearest neighbor distance is less than a predetermined threshold. A common relative distance test is the 2NN test \cite{14}. Additionally, to fit multiple copy-move forgery detection, Amerini \cite{15} proposed the Generalized 2NN (G2NN). Subsequently, in order to solve the premature termination of G2NN, Wang \cite{29} proposed the Reversed Generalized 2NN (RG2NN). However, RG2NN often requires a large computational cost. To mitigate this problem, Yang \cite{19} proposed Improved Generalized 2NN (IG2NN). Recently, Wang \cite{27} combined the absolute distance test and the relative distance test to form Improved 2NN (I2NN) test, with much fewer mismatches and complexity.

To speed up the matching processing, the most common techniques were Approximate Nearest Neighbor (ANN). These techniques were mainly based on PatchMatch (PM) \cite{6,9,12,30}, lexicographic sorting \cite{2,7,10,28}, k-d tree \cite{3} and hashing \cite{4,5,12}. In particular, PM-based algorithm were proposed for dense-field matching algorithm, which fully considered the spatial correlation of the matching area. Besides, clustering matching \cite{8,22,24,25,27} showed excellent performance keypoint-based algorithms.

In summary, existing matching algorithms can be performed efficiently, even in situations with high computational burden. However, once the number of features increases significantly, it becomes difficult for the algorithms to maintain matching accuracy, especially keypoint-based algorithms. This paper employs grouping techniques to accelerate matching. our method combines clustering and ANN techniques, achieving a great balance between accuracy and efficiency.

\subsection{Forgery Localization Stage}
Since the basic information of the copy-move operation (homography matrix, number of tampering) cannot be known in advance, the core of the forgery positioning stage is to recover the basic information of the copy-move operation, and then locate the forgery regions in dense-fields.

The RANSAC algorithm can estimate a reliable homography matrix even if there is a lot of noise in the set. However, the RANSAC algorithm may fail when there are multiple homography matrices (models) in the image. To overcome this problem, clustering is often used to separate models, which typically fall into two categories: space-based and concept-based clustering. Space-based clustering utilizes the spatial relationships among matches, such as Hierarchical Agglomerative Clustering (HAC) \cite{15}, Density-Based Spatial Clustering of Applications with Noise (DBSCAN) \cite{26}, etc. On the other hand, concept-based clustering leverages the constraint relationships among the matches, such as J-linkage \cite{16}, offset vector \cite{25}, etc. In general, the latter outperforms the former in terms of model separation, especially for separated models that are close to each other.

To locate forgery regions in the dense-field, the forgery localization stage requires the identification of suspicious regions. Early localization algorithms \cite{14,16} often consider the entire image as a single suspicious area and apply Zero-Mean Normalized Cross Correlation (ZNCC) to all pixels. In terms of localization performance, these methods can ensure fewer missed detections, but it may also bring many false alarms. To overcome this problem, utilizing matching information to constrain the suspicious region has become a mainstream solution in recent years. For example, Xiang‑yang Wang \cite{31} computed ZNCC based on Regions Of Interest (ROI); Pun \cite{32,33} introduced Simple Linear Iterative Clustering (SLIC) to determine suspicious regions; Zandi \cite{18}, Hai-peng Chen \cite{24} and Chien-Chang Chen \cite{41} proposed region growing algorithm to optimize localization results.

In summary, most forgery localization algorithms focus on how to separate models and validate the model estimated by RANSAC algorithm. However, keypoint-based algorithms use stable keypoints for matching, which makes these algorithms assume that all matches are formed by tampering. This makes the performance of scheme degrade when facing SGO images. In this paper, the iterative forgery localization algorithm \cite{22} is further developed into a new iterative forgery localization algorithm, which reduces the false alarm rate.

\section{Proposed method}\label{proposed method}
\begin{figure}[ht]
	\centering
	\includegraphics[width=1\linewidth]{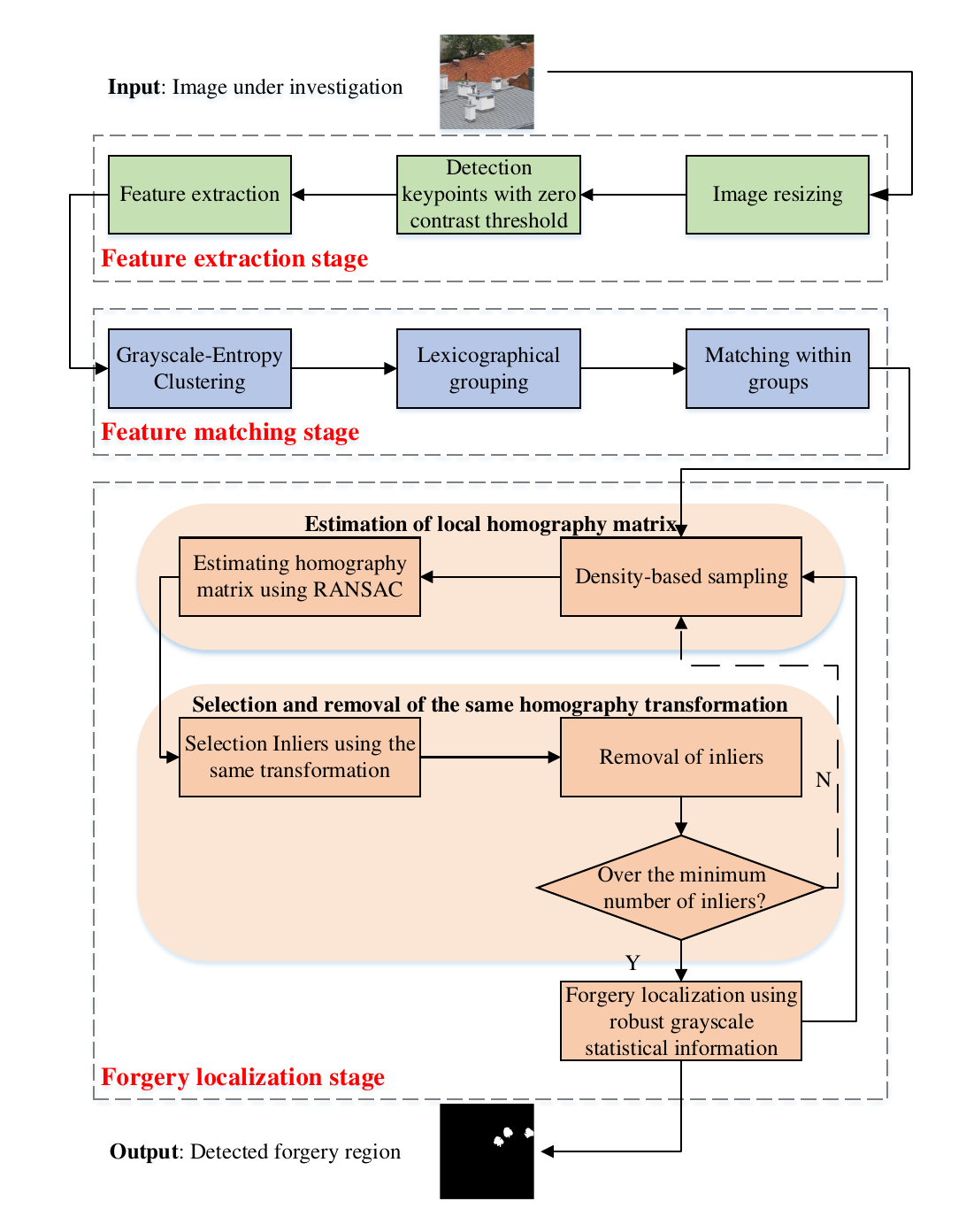}
	\caption{Framework of the proposed scheme.}
	\label{Fig2}
\end{figure}

To overcome missed detection in low-resolution images and false alarms in SGO images, this paper investigates the three stages of copy-move forgery detection. The proposed scheme framework is shown in Fig. \ref{Fig2}. Our scheme follows the classical CMFDL flowchart. In the feature extraction stage, the core is to ensure keypoint coverage to reduce missed detections; in the feature matching stage, the core is to maintain a balance between precision and efficiency under the excessive keypoint strategy; in the forgery localization stage, the core is to reduce false alarms and accurately locate tampering areas in dense-fields.

\subsection{Feature Extraction Stage}\label{proposed method1}
Over the past few years, an excessive amount of research effort has been focused on ensuring the number of keypoints within a complete semantic, especially for smooth or small semantic patch. For instance, Li \cite{22} extracted patches with the minimum variance and then ensured that there were 4 keypoints within 1200 pixels; Wang \cite{27} ensured that the number of keypoints in the tampering region was no less than 600. It must be acknowledged that these works exhibited excellent performance on public datasets. However, a neglected question is: what are the normal working limits of the keypoint-based algorithms? How to avoid missed detection problems as much as possible?

In keypoint-based CMFDL, local feature descriptors are used to measure the similarity between two local patches. When the size of the tampered region is smaller than the local descriptor (feature) window size, the information captured by the features is not exclusively derived from the tampering. Consequently, even if feature matching occurs, there is insufficient evidence to assert the presence of a copy-move operation. Based on this assertion, this paper suggests that the distribution of keypoints in CMFDL should ensure that there are at least 4 keypoints within the feature window size (The RANSAC algorithm requires 4 pairs of matches to determine a homography matrix, and these 4 pairs of matches necessitate at least 4 keypoints). The adopted window size is $ 16 \times 16 $, which is the smallest window size of SIFT algorithm (at the lowest octave). To implement this suggestion, the original image needs to be resized. However, resizing will greatly increase the number of keypoints, making it difficult to implement subsequent stages, especially for those high-resolution images. To balance efficiency and coverage, a lower scaling factor $ s $ is adopted for higher resolution images. Even so, the number of keypoints generated during the feature extraction stage of this paper still far exceeds that of existing algorithms. Therefore, the suggestion used in the feature extraction stage of this paper is named the \textbf{excessive keypoint strategy}. In this paper, the scaling factor $ s $ is defined as:
\begin{equation}
	\label{Eq1}
	s = \left\{ \begin{array}{l}
		4,\frac{{(h + w)}}{2} < 1024\\
		2,\frac{{(h + w)}}{2} \ge 1024
	\end{array} \right.,
\end{equation}
where, $ h \times w $ represents the resolution of the input image $ I $. Subsequently, a set of keypoints $ {\bf{K}} = \{ {{\bf{k}}_1},{{\bf{k}}_2}, \cdots ,{{\bf{k}}_n}\} $ and their corresponding features $ {\bf{F}} = \{ {{\bf{f}}_1},{{\bf{f}}_2}, \cdots ,{{\bf{f}}_n}\} $ are extracted using the SIFT algorithm by reducing contrast threshold $ T_{con} $ ($ T_{con}=0 $). Assuming that $ {{\bf{k}}_i} $ $ (i \in [1,n]) $ is a keypoint in $ \bf{K} $, it can be expressed as:
\begin{equation}
	\label{Eq2}
	{{\bf{k}}_i} = ({x_i},{y_i},{\sigma _i},{\theta _i}),
\end{equation}
where $ ({x_i},{y_i}) $, $ {\sigma _i} $ and $ {\theta _i} $ denote coordinates, scale and dominant orientation, respectively. As for $ \bf{F} $, all features are converted in normalized form ($ \left\| {\bf{f}}_i \right\|_{2}\ = 1, \forall i \in [1,n] $).

\subsection{Feature Matching Stage}
The greater the number of keypoints, the higher the computational burden. Therefore, keypoint-based algorithms rarely attempted excessive keypoint matching (in other words, they use a normal number of keypoints). Although a normal number of keypoints can ensure that the algorithms maintain a good performance in terms of true positive rate, these methods rely heavily on the homography matrix determined by a small number of keypoints. For example, when the contrast threshold of the SIFT algorithm is removed, even with a normal number of keypoints, some authentic regions (such as windows) tend to produce multiple matches (typically more than 4). It is enough to generate a homography matrix, and then enables dense-field localization. Most algorithms do not have enough information to further verify the generated homography matrix, which is the real reason for the high false alarm rate of SGO images.

To overcome this problem, it is necessary to increase the redundancy of matching information, thereby enhancing the ability to distinguish the source of the homography matrix (self-similarity or tampering) in the forgery localization stage. This is another reason to apply the excessive keypoint strategy. However, efficiently and accurately matching an excessive of keypoints is very challenging. This is because clustering-based matching is difficult to perform efficiently, while ANN-based matching is difficult to form an accurate index. This paper forms a \textbf{fast group matching algorithm} by combining the above two methods. Although this method may lose some correct matches, under the premise of excessive keypoints, the number of correct matches is still much higher than the existing algorithms. In particular, the proposed group matching algorithm consists of the following steps:

\begin{itemize}
	\item \textbf{Step1:} Grayscale-entropy clustering;
	\item \textbf{Step2:} Lexicographical grouping (LG);
	\item \textbf{Step3:} Matching within groups.
\end{itemize}

\begin{figure}[ht]
	\centering
	\includegraphics[width=0.5\linewidth]{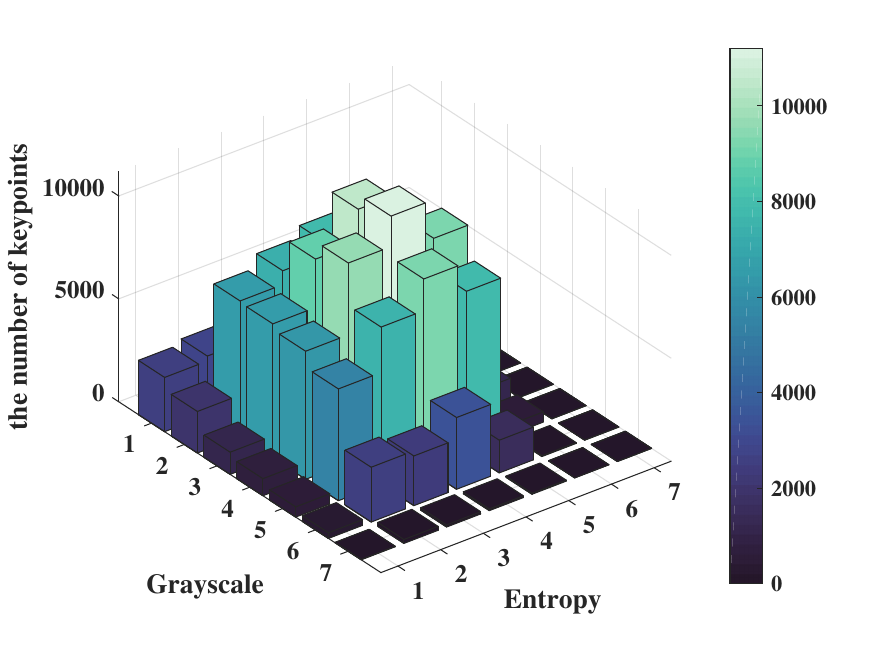}
	\caption{Illustration of the grayscale-entropy clustering.}
	\label{Fig3}
\end{figure}

\textbf{Step1:} \textbf{\textit{Grayscale-Entropy Clustering}}. This step groups keypoints into clusters based on grayscale and entropy information. As illustrated in Fig. \ref{Fig3}, the grayscale-entropy clustering can be essentially seen as an overlapped grid partition in a two-dimensional space. It enhances the similarity of features within each group, which is beneficial to constructing an excellent ANN index in the subsequent step.

In this paper, the overlapped gray level clustering proposed by Li \cite{22} is firstly improved, which is defined as:
\begin{equation}
	\label{Eq3}
	\left\{ \begin{array}{l}
		{{\bf{C}}^u} = \left\{ {k{_i}|{I_{gray}}(k{_i}) \in [a_l^u,a_h^u],k{_i} \in {\bf{K}}} \right\}\\
		a_l^u = (u - 1) \cdot (ste{p_1} - ste{p_2})\\
		a_h^u = \min (a_l^u + ste{p_1},255)
	\end{array} \right.,
\end{equation}
where, $ I_{gray} $ denotes the grayscale image converted from the input image $ I $, and $ step_1 $ and $ step_2 $ represent the fixed step size and overlapped step size of the gray level clustering, respectively. The most significant difference from Li's algorithm \cite{22} is that the grayscale of keypoints is used to replace the average grayscale of the local patch. This is primarily because the excessive keypoints strategy can express information more fully and accurately.
It should be noted that $ step_1 = 40, step_2 = 10 $, $ u \in [1,{N_u}] $. Formally, $ N_u $ is defined as:
\begin{equation}
	\label{Eq4}
	{N_u} = \left\lceil {\frac{{255 - ste{p_1}}}{{ste{p_1} - ste{p_2}}}} \right\rceil.
\end{equation}

The overlapped gray level clustering \cite{22} demonstrated excellent performance, which could improve matching efficiency without losing correct matches. Inspired by it, the overlapped entropy clustering is further forms. To better define the overlapped entropy clustering, the method for calculating entropy is defined as:
\begin{equation}
	\label{Eq5}
	{\mathop{\rm E}\nolimits} (x,y) = \sum\limits_{i = 0}^{255} {{P_i}(x,y) \cdot {{\log }_2}(} {P_i}(x,y)),
\end{equation}
where, $ {P_i}(x,y) $ is the probability of gray level $ i $ within the $ B_E \times B_E $ region centered at $ (x,y) $. Subsequently, the overlapped entropy clustering is defined as:
\begin{equation}
	\label{Eq6}
	\left\{ \begin{array}{l}
		{{\bf{C}}^{u,v}} = \left\{ {{k_i}|{\mathop{\rm E}\nolimits} ({k_i}) \in [a_l^{u,v},a_h^{u,v}],{k_i} \in {{\bf{C}}^u}} \right\}\\
		a_l^{u,v} = \max (0,(v - 1) \cdot ste{p_3} - ste{p_4})\\
		a_h^{u,v} = \min (7,v \cdot ste{p_3} + ste{p_4})
	\end{array} \right.,
\end{equation}
where, $ step_3 $ and $ step_4 $ represent the fixed step size and overlapped step size of the entropy clustering, respectively ($ step_3 > step_4 $, $ v \in [1,{N_{u,v}}] $). Mathematically, $ N_{u,v} $ is defined as:
\begin{equation}
	\label{Eq7}
	{N_{u,v}} = \left\lceil {\frac{{7 - ste{p_4}}}{{ste{p_3}}}} \right\rceil.
\end{equation}

The design of Equation (\ref{Eq6}) takes into account the theoretical upper bound of entropy. In this paper, $ B_E $ is set to 9 (MATLAB default setting). Combining knowledge from information theory, it is known that the maximum value of entropy is $ - {\log _2}(1/81) \approx 6.34 $. This paper rounds this maximum value up, hence it is 7. Furthermore, Equation (\ref{Eq6}) allows entropy clustering to overlap, which is beneficial to clustering without losing correct matches. Since an excessive number of keypoints are formed during the feature extraction stage, losing some correct matches has limited impact on subsequent stages. Therefore, this paper adopts a non-overlapping settings with $ step_3=1 $, $ step_4=0 $.

\textbf{Step2:} \textbf{\textit{Lexicographical Grouping}}. This step divides clusters with a high number of keypoints into several small groups with a similar number of keypoints, which is conducive to efficient matching.

Traditional ANN-based matching methods often rely on independent queries. In the CMFDL matching task, there is inevitably a certain amount of feature similarity information that can be shared among multiple consecutive queries. For example, suppose $ \left\{ {{{\bf{f}}_{(1)}},{{\bf{f}}_{(2)}},\; \cdots ,{{\bf{f}}_{(n)}}} \right\} $ is a set of feature that undergoes lexicographic sorting. When searching for the NN of a feature $ {{\bf{f}}_{(i)}} $ (the $ i $-th feature after lexicographic sorting), $ {{\bf{f}}_{(i)}} $ will calculate the similarity with the nearest $ step_L $ features, and then consider the feature with the highest similarity as the NN for $ {{\bf{f}}_{(i)}} $. When the NN of feature $ {{\bf{f}}_{(i+1)}} $ is queried next time, $ {{\bf{f}}_{(i+1)}} $ must be recalculated with $ {{\bf{f}}_{(i)}} $, and so on. In particular, in the case of excessive keypoints, a larger $ step_L $ needs to be used, which will inevitably intensify repeated calculations and greatly consume resources.

Grouping is used to overcome this problem. Suppose that there are $ n $ features and divide them into $ m $ groups. It is hoped that the number of feature similarity calculations will be as small as possible, so the objective function of the number of feature similarity calculations is:
\begin{equation}
	\label{Eq8}
	{\mathop{\rm argmin}\nolimits} \sum\limits_{i = 1}^m {\frac{{{n_i}(1 + {n_i})}}{2}} ,\sum\limits_{i = 1}^m {{n_i} = n},
\end{equation}
where, Equation (\ref{Eq8}) is equivalent to:
\begin{equation}
	\label{Eq9}
	{\mathop{\rm argmin}\nolimits} \sum\limits_{i = 1}^m {n_i^2} ,\sum\limits_{i = 1}^m {{n_i} = n}.
\end{equation}

Clearly, Equation (\ref{Eq9}) can be resolved using the Cauchy-Schwarz inequality. The result is that the objective function has a minimum value when each group has the same number of elements. However, most of the existing common grouping methods, such as clustering and hash grouping, cannot guarantee that the number of elements in each group is similar. This paper uses the LG to approximate this limit, which includes the following steps:

1) Obtain the feature set $ {{\bf{F}}_{{{\bf{C}}^{u,v}}}} $ corresponding to $ {{\bf{C}}^{u,v}} $, which is defined as:
\begin{equation}
	\label{Eq10}
	{{\bf{F}}_{{{\bf{C}}^{u,v}}}} = {\mathop{\rm des}\nolimits} ({{\bf{C}}^{u,v}}) = \{ {\bf{f}}{}_{{\bf{C}}_1^{u,v}},{{\bf{f}}_{{\bf{C}}_2^{u,v}}}, \cdots ,{{\bf{f}}_{{\bf{C}}_{{n_{u,v}}}^{u,v}}}\},
\end{equation}
where, $ {\mathop{\rm des}\nolimits} ( \bullet ) $ represents the relationship between keypoints and corresponding features, and $ n_{u,v} $ denotes the number of features in $ {{\bf{C}}^{u,v}} $.

2) Sort $ {{\bf{F}}_{{{\bf{C}}^{u,v}}}} $ by lexicographic order, which is defined as:
\begin{equation}
	\label{Eq11}
	\begin{array}{ll}
		{{\bf{S}}^{u,v}}&= {\mathop{\rm sortrows}\nolimits} ({{\bf{F}}_{{{\bf{C}}^{u,v}}}})\\
		&= \{ {{\bf{f}}_{{\bf{C}}_{(1)}^{u,v}}},{{\bf{f}}_{{\bf{C}}_{(2)}^{u,v}}}, \cdots ,{{\bf{f}}_{{\bf{C}}_{({n_{u,v}})}^{u,v}}}\}
	\end{array},
\end{equation}
where, $ {{\bf{S}}^{u,v}} $ denotes the lexicographic feature set, and $ {\mathop{\rm sortrows}\nolimits} ( \bullet ) $ represents the lexicographic sorting operation. Additionally, $ {{\bf{S}}_{i}^{u,v}} $ ($ {{\bf{f}}_{{\bf{C}}_{(i)}^{u,v}}} $) is defined as the $ i $-th feature of $ {{\bf{S}}^{u,v}} $.

3) Obtain the LG feature set $ {{\bf{S}}^{u,v,w}} $ and LG keypoint set $ {{\bf{C}}^{u,v,w}} $, which are defined as:
\begin{equation}
	\label{Eq12}
	\left\{ \begin{array}{l}
		{{\bf{S}}^{u,v,w}} = \{ {\bf{S}}_{(i)}^{u,v}|i \in [a_l^{u,v,w},a_h^{u,v,w}]\} \\
		a_h^{u,v,w} = \min ({n_{u,v}},w \cdot ste{p_5})\\
		a_l^{u,v,w} = \max (1,(w - \beta ) \cdot ste{p_5})
	\end{array} \right.,
\end{equation}
\begin{equation}
	\label{Eq13}
	{{\bf{C}}^{u,v,w}} = {{\mathop{\rm des}\nolimits} ^{ - 1}}({{\bf{S}}^{u,v,w}}),
\end{equation}
where, $ step_{5} $ denotes the fixed step size of the LG, and $ \beta $ represents the overlapped factor of the LG ($ \beta \in [1,2] $, $ w \in [1,{N_{u,v,w}}] $). Here, $ {N_{u,v,w}} $ can be defined as:
\begin{equation}
	\label{Eq14}
	{N_{u,v,w}} = \left\lceil {\frac{{{n_{u,v}}}}{{ste{p_5}}}} \right\rceil.
\end{equation}

To provide recommendation on the overlapped factor $ \beta $, an analysis is conducted on the number of similarity calculations for both LG and lexicographic matching. For a set of $ n $-dimensional features, when the number of similarity calculations of the LG matching is less than that of lexicographic matching, it can be approximately expressed as:
\begin{equation}
	\label{Eq15}
	\frac{n}{{ste{p_5}}} \cdot \frac{{\beta  \cdot ste{p_5}(1 + \beta  \cdot ste{p_5})}}{2} \le n \cdot ste{p_L}.
\end{equation}
Through simplification, it is equivalent to:
\begin{equation}
	\label{Eq16}
	\beta  + {\beta ^2} \cdot ste{p_5} \le 2 \cdot ste{p_L},
\end{equation}
where, $ step_L $ represents the ANN interval of lexicographic matching. Under the premise that $ \beta  \ll 2 \cdot ste{p_L} $, it can be further approximated as:
\begin{equation}
	\label{Eq17}
	ste{p_5} \le \frac{{2ste{p_L}}}{{{\beta ^2}}}.
\end{equation}

When $ step_5 =step_L $, the condition that the number of feature similarity calculations of the LG is less than that of lexicographic matching is $ \beta  \in [1,\sqrt 2 ] $. In this paper, $ step_{5} = 500 $, $ \beta = 1.1 $ is adopted.

\textbf{Step3:} \textbf{\textit{Matching within Groups}}. In order to deal with possible multiple copy-move forgeries, this step adopts the G2NN matching strategy.

Suppose that there is a LG feature set $ {{\bf{S}}^{u,v,w}} $, which contains $ n_{u, v,w} $ features. For a certain feature $ {\bf{f}}_{i} $ in $ {{\bf{S}}^{u,v,w}} $, its distance from other features in increasing order can be expressed as $ \{ d_{(1)}, d_{(2)}, \cdots ,d_{(n_{u,v,w}-1)} \} $, and then matches will be recognized under the following conditions:
\begin{equation}
	\label{Eq18}
	\frac{{{d_{(j)}}}}{{{d_{(j + 1)}}}} < {T_{match}},j = 1, \cdots ,{n_{u,v,w - 1}},
\end{equation}
where, $ T_{match} $ denotes the relative threshold of the G2NN. $ T_{match} = 0.5 $ is adopted in this paper. The matching set in the grouped set ${{\bf{C}}^{u,v,w}}$ is represented as ${{\bf{M}}^{u,v,w}}$. Finally, the matching set \textbf{M} is expressed as:
\begin{equation}
	\label{Eq19}
	\begin{array}{ll}
		{\bf{M}} &= \bigcup\limits_{u = 1}^{{N_u}} {\bigcup\limits_{v = 1}^{{N_{u,v}}} {\bigcup\limits_{w = 1}^{{N_{u,v,w}}} {{{\bf{M}}^{u,v,w}}} } } \\
		&=\{ < {{\bf{k}}_{l_{i}}},{{\bf{k}}_{r_{i}}} >| i \in [1,n_{\textbf{M}}]\}
	\end{array},
\end{equation}
where, $ {{\bf{k}}_{l}} $ and $ {{\bf{k}}_{r}} $ denote the left and right keypoint sets of the matching set \textbf{M}, respectively. $ < \bullet > $ indicates a direct match. The number of directed matches in the matching set \textbf{M} is denoted by $ n_\textbf{M} $.

\subsection{Forgery Localization Stage}
\begin{figure*}[ht]
	\centering
	\begin{tabular}{cccccc}
		{\rotatebox{90}{\textbf{Iteration 1}}} & \includegraphics[width=0.15\linewidth]{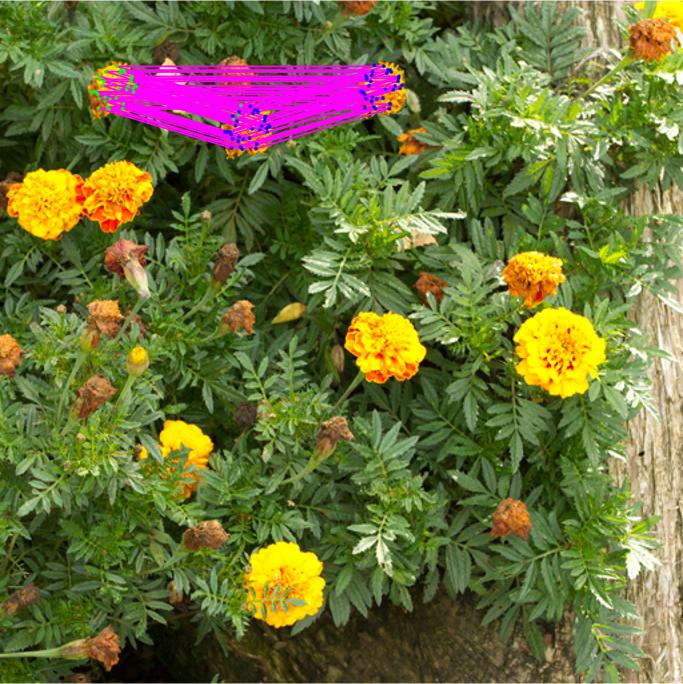} &
		\includegraphics[width=0.15\linewidth]{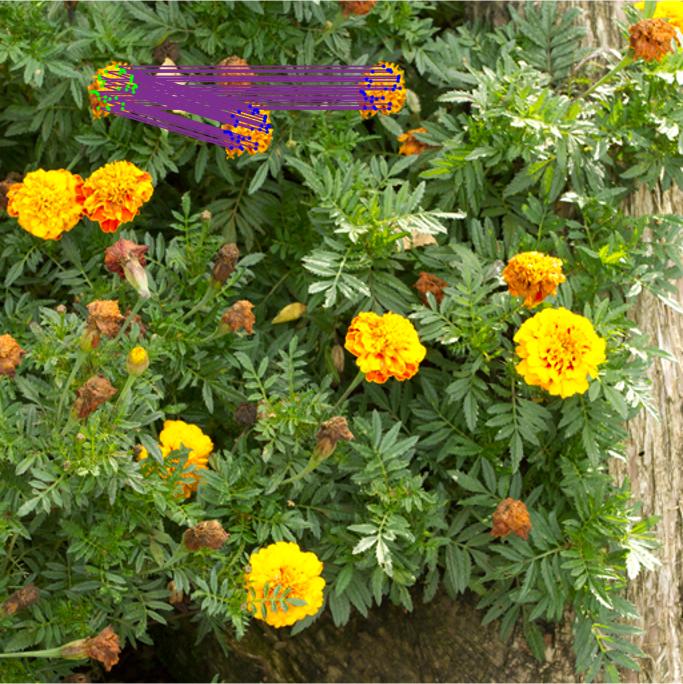} &
		\includegraphics[width=0.15\linewidth]{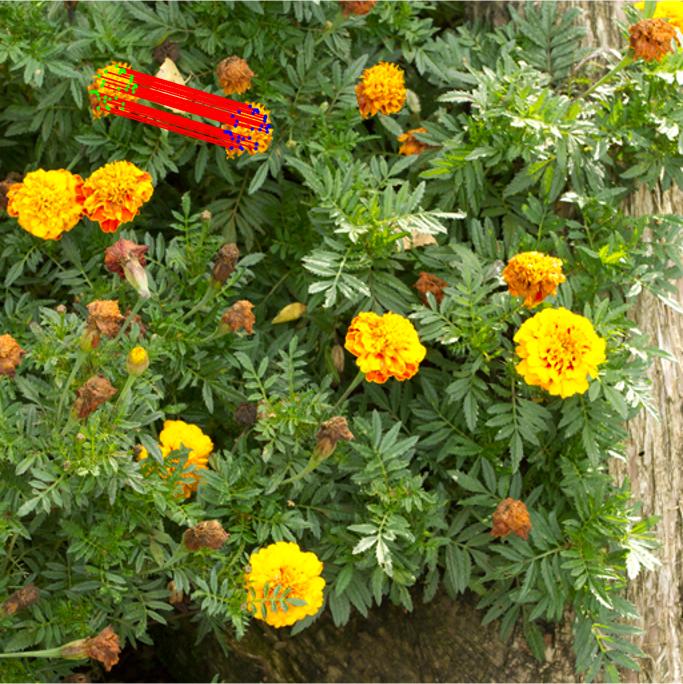} &
		\includegraphics[width=0.15\linewidth]{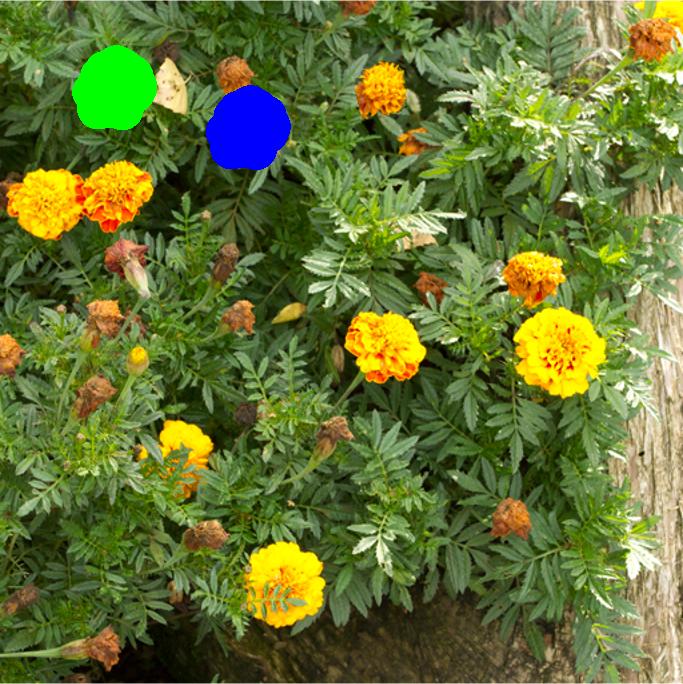} &
		\includegraphics[width=0.15\linewidth]{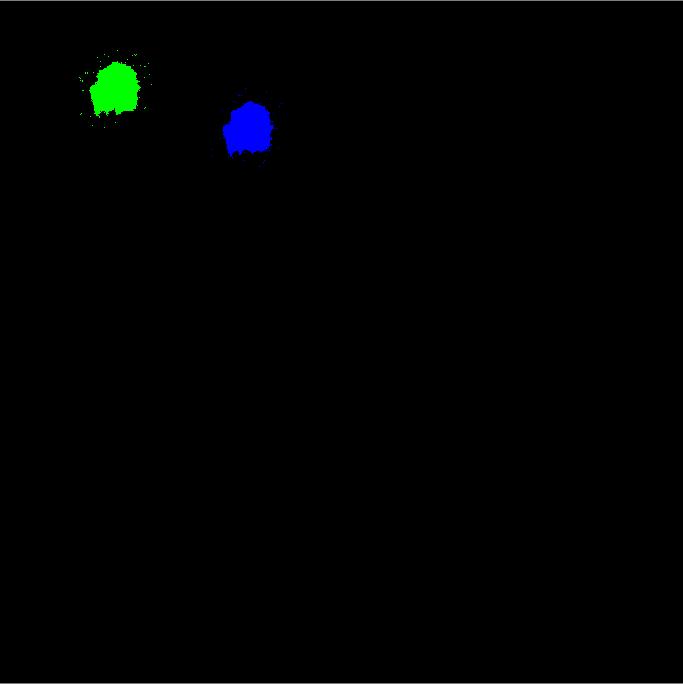} \\
		{\rotatebox{90}{\textbf{Iteration 2}}} & \includegraphics[width=0.15\linewidth]{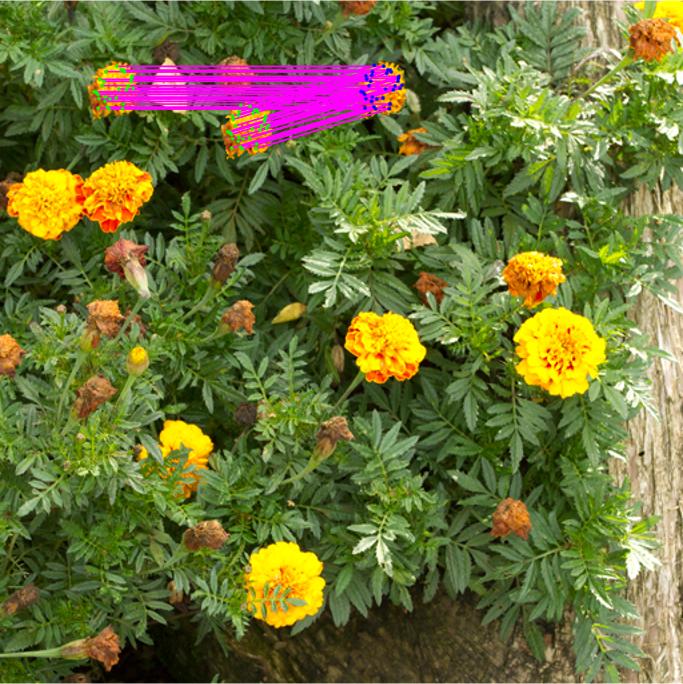} &
		\includegraphics[width=0.15\linewidth]{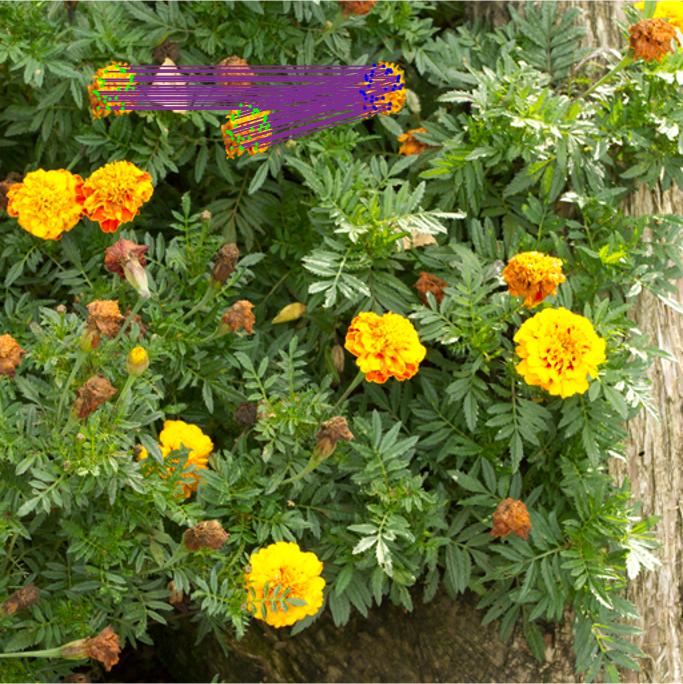} &
		\includegraphics[width=0.15\linewidth]{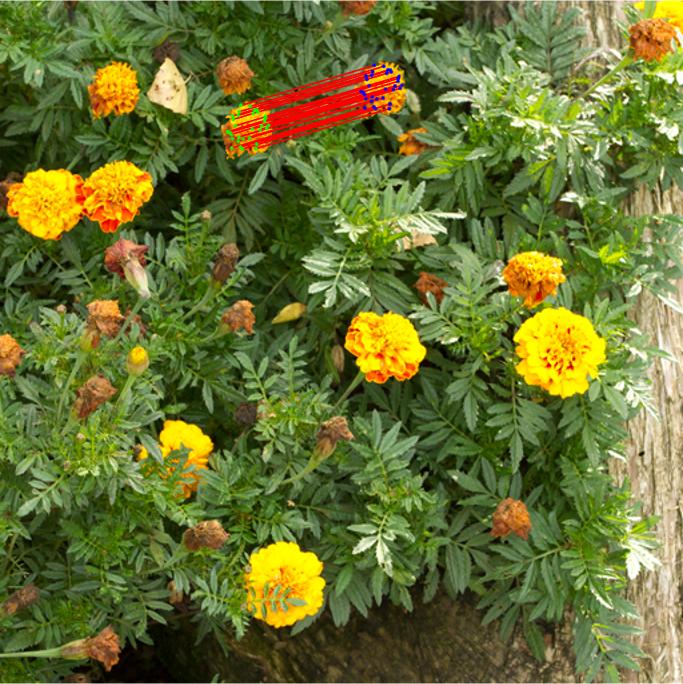} &
		\includegraphics[width=0.15\linewidth]{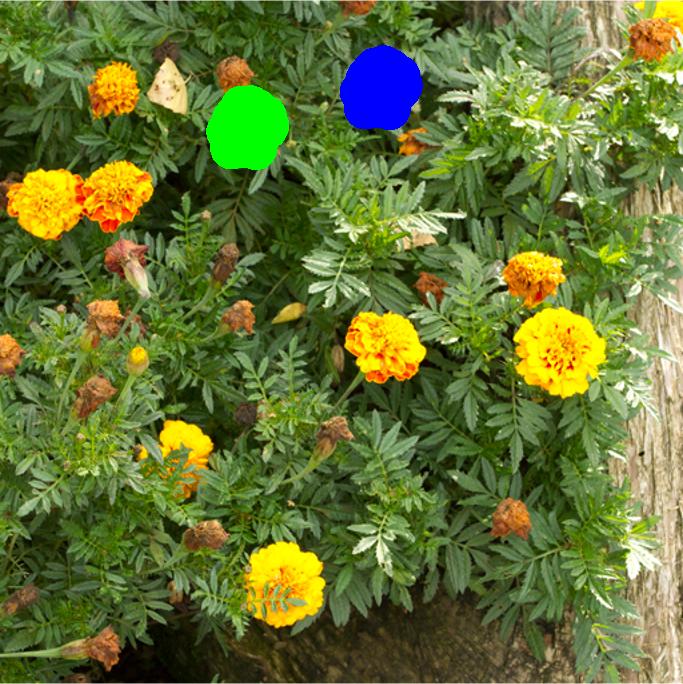} &
		\includegraphics[width=0.15\linewidth]{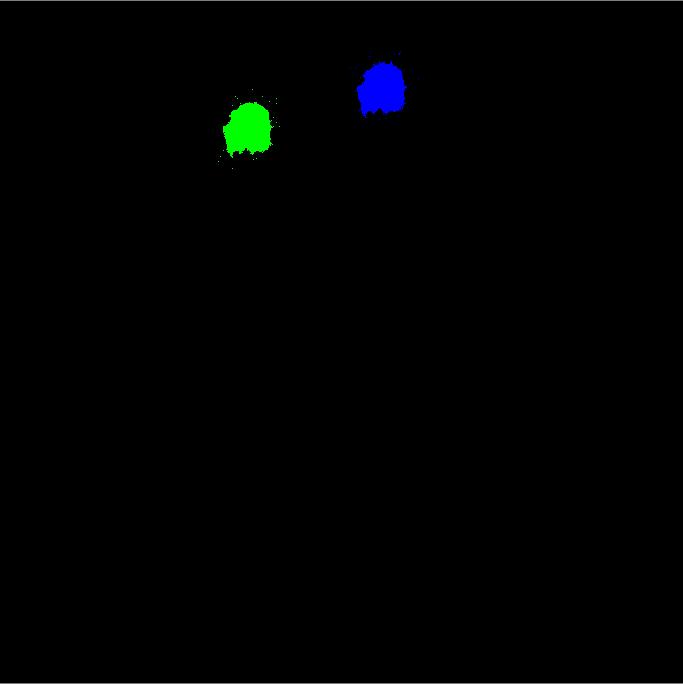} \\
		{\rotatebox{90}{\textbf{Iteration 3}}} & \includegraphics[width=0.15\linewidth]{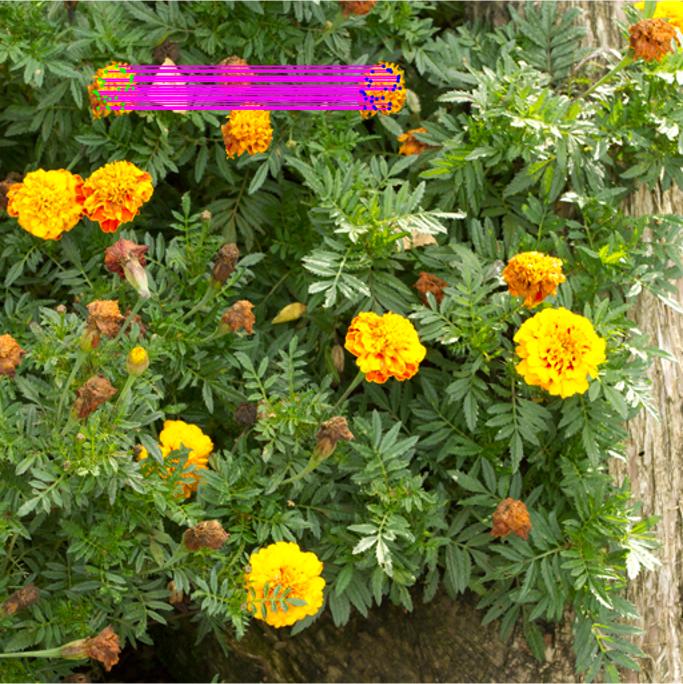} &
		\includegraphics[width=0.15\linewidth]{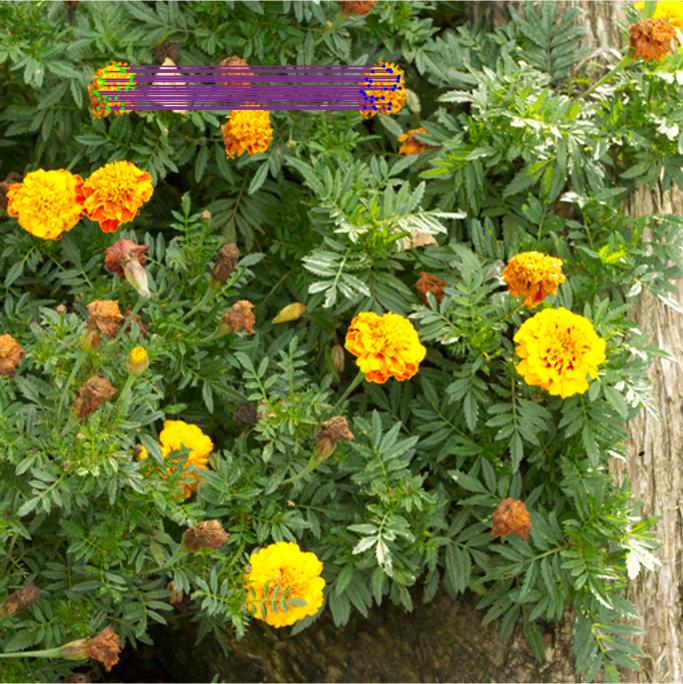} &
		\includegraphics[width=0.15\linewidth]{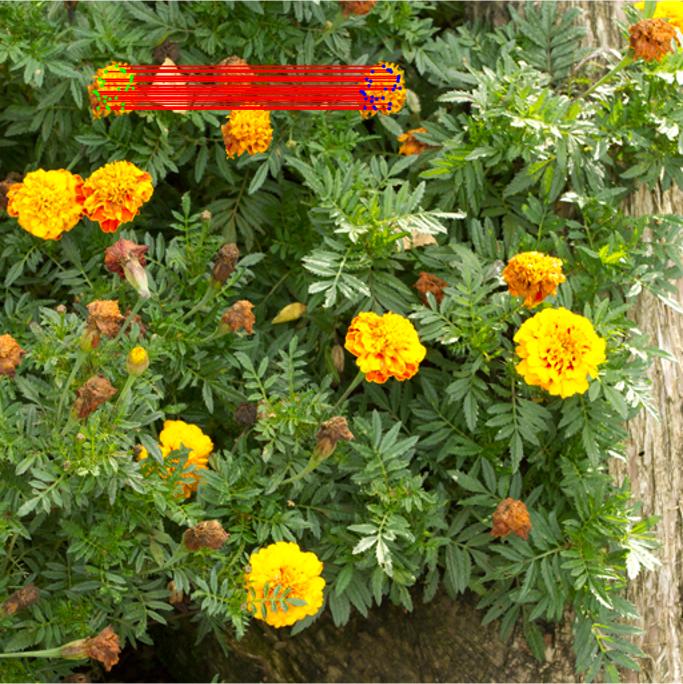} &
		\includegraphics[width=0.15\linewidth]{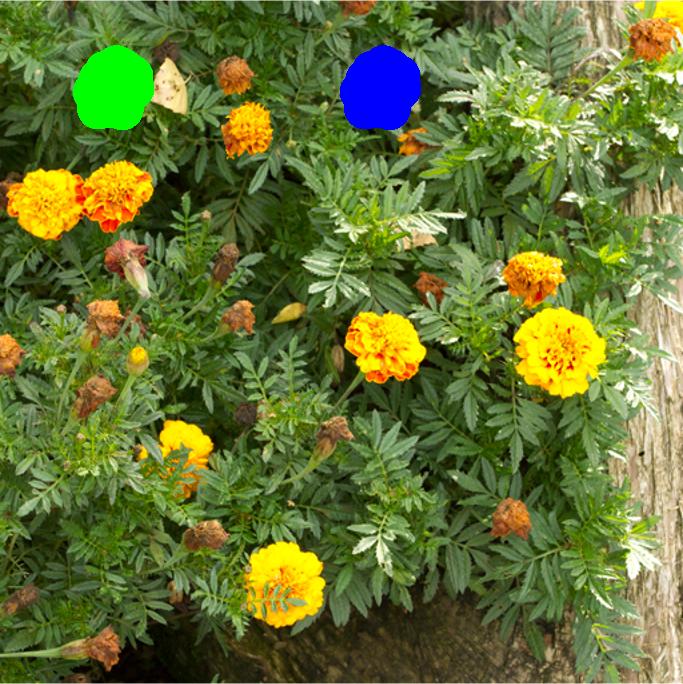} &
		\includegraphics[width=0.15\linewidth]{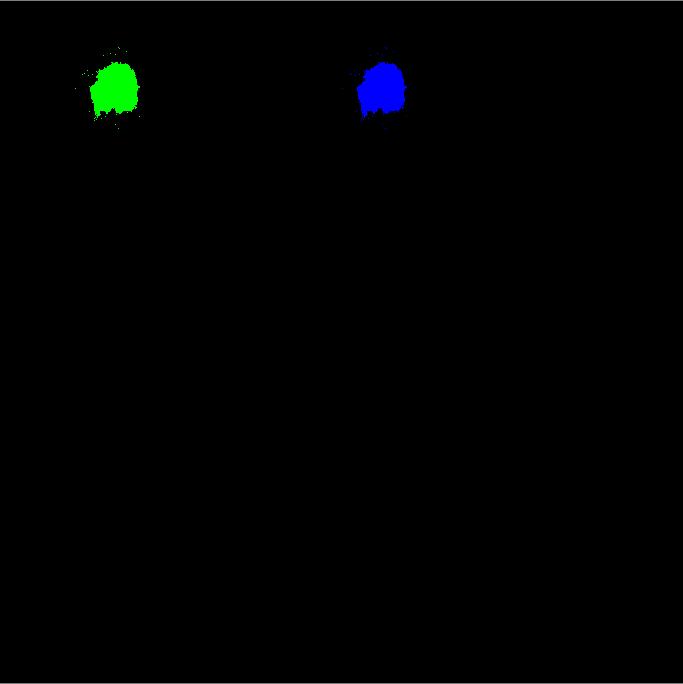} \\
		& (a) & (b) & (c) & (d) & (e) \\
	\end{tabular}
	\caption{Illustration of the new iterative forgery localization algorithm. (a) unvisited matches; (b) sampling set; (c) inliers with the same homography transformation; (d) suspicious region and (e) localization result in the iteration.}
	\label{Fig4}
\end{figure*}

The feature matching stage based on the keypoint algorithm provides a rough judgment result of similar areas. Such judgment results usually have many problems. On the one hand, when an excessive number of keypoints are extracted, although the missed detections of the scheme is effectively alleviated, it suffers from a series of false alarms. On the other hand, the feature matching results based on keypoints cannot provide precise pixel-level location results. This paper notices that the iterative forgery localization algorithm \cite{22} has significant advantages in accuracy and complexity. This is mainly due to: 1) Compared to non-restrictive suspicious region algorithms, it uses the scale information of keypoints to constrain the region, significantly reducing false detections in SGO images and minimizing unnecessary computation. 2) Compared with segmentation-based algorithms, it avoids the resources consumed by segmentation. Moreover, it is challenging to find a universal algorithm to handle unknown tampering. However, there are still several problems with the iterative forgery localization algorithm, mainly including: 1) Its sampling results cannot guarantee the extraction of all homography matrices (models). 2) Its sampling set may contain multiple models. The consequence of the above problems is that the iterative forgery location algorithm performs poorly in the case of multiple tampering, especially for detecting one-to-many copy-move. To overcome these problems, a new iterative forgery localization algorithm is developed, and a typical illustration is shown in Fig. \ref{Fig4}. Overall, the new iterative forgery localization algorithm is divided into three steps:
\begin{itemize}
	\item \textbf{Step1:} Estimation of local homography matrix;
	\item \textbf{Step2:} Selection and removal of the same homography transformation;
	\item \textbf{Step3:} Forgery localization using robust grayscale statistical information.
\end{itemize}

\textbf{Step1:} \textbf{\textit{Estimation of Local Homography Matrix}}. This step firstly forms a spatially close sampling set and then estimates its homography matrix.

To obtain a sampling set, the most common method is random sampling, such as Same Affine Transformation Selection (SATS) \cite{34} and iterative forgery localization \cite{22}. However, this method is similar to the iterative estimation of the RANSAC algorithm. If you are not lucky enough, you may not get all the models in the match. To alleviate this problem, \textbf{density-based sampling} is used. This method ensures that all models are sampled through iteration.

Firstly, define a directed set \textbf{Unvisited}, which is equivalent to \textbf{M}. In this paper, \textbf{Unvisited} is used to record the unvisited matches in the iteration. $ \forall <{\bf{k}}_{l_{i}},{\bf{k}}_{r_{i}}> \in \textbf{Unvisited} $ ($ i \in [1,n_\textbf{M}] $), the left keypoint sampling set $ {\bf{Sam}}_{l_{i}} $ and the right keypoint sampling set $ {\bf{Sam}}_{r_{i}} $ are defined as:
\begin{equation}
	\label{Eq20}
	{\bf{Sam}}_{l_{i}} = \{ {\bf{k}}|\forall {\bf{k}} \in {{\bf{Unvisited}}},{\mathop{\rm Dis}\nolimits} ({\bf{k}},{{\bf{k}}_{l_{i}}})<R_{sam}\},
\end{equation}
\begin{equation}
	\label{Eq21}
	{\bf{Sam}}_{r_{i}} = \{ {\bf{k}}|\forall {\bf{k}} \in {{\bf{Unvisited}}},{\mathop{\rm Dis}\nolimits} ({\bf{k}},{{\bf{k}}_{r_{i}}})<R_{sam}\},
\end{equation}
where, $ {\mathop{\rm Dis}\nolimits} ( \bullet ) $ is used to calculate the spatial distance of keypoints; $ R_{sam} $ denotes the sampling radius.
Subsequently, merge $ {\bf{Sam}}_{l_{i}} $ and $ {\bf{Sam}}_{r_{i}} $ to obtain a sampling set $ {\bf{Sam}}_{i} $, which is defined as:
\begin{equation}
	\label{Eq22}
	{\bf{Sa}}{{\bf{m}}_{{{\bf{k}}_i}}} = {\bf{Sam}}_{l_{i}} \bigcup {\bf{Sam}}_{r_{i}}.
\end{equation}

Then, the final sampling set $ {\bf{Sam}}_{\textbf{k}} $ of this iteration is determined by the following conditions:
\begin{equation}
	\label{Eq23}
	{\bf{Sa}}{{\bf{m}}_{\bf{k}}} = {\mathop{\rm argmax}\nolimits} \left\| {{\bf{Sa}}{{\bf{m}}_{{{\bf{k}}_i}}}} \right\|,i \in [1,{n_{\bf{M}}}],
\end{equation}
where, $ \left\|  \bullet  \right\| $ is used to calculate the number of elements in the set.

Finally, RANSAC algorithm is performed on the sampling set $ {\bf{Sam}}_{\textbf{k}} $ to obtain the local homography matrix $ {\bf{H}}_{\bf{k}} $.

\textbf{Step2:} \textbf{\textit{Selection and Removal of the Same Homography Transformation}}. This step firstly validates the homography matrix. Then, the inliers of the same homography transformation are selected and removed. Finally, the source of the homography matrix is distinguished.

For homography validation, dominant orientation is used to check the homography matrix formed by the sampling set $ {\bf{Sam}}_{\textbf{k}} $ \cite{22}. Subsequently, a set of inliers \textbf{Inlier} with the same homography matrix is selected, which is defined as:
\begin{equation}
	\label{Eq24}
	\begin{array}{ll}
		\textbf{Inlier} = \{  < {{\bf{k}}_{{l_i}}},{{\bf{k}}_{{r_i}}} > &| \forall < {{\bf{k}}_{{l_i}}},{{\bf{k}}_{{r_i}}} >  \in {\bf{M}},\\
		& {\left\| {{{\bf{H}}_{\bf{k}}} \cdot {{\bf{k}}_{{l_i}}} - {{\bf{k}}_{{r_i}}}} \right\|_2} < {T_{in}}\} 
	\end{array},
\end{equation}
where, $ T_{in} $ denotes the inliers selection threshold.

\begin{figure}[ht]
	\centering
	\includegraphics[width=0.5\linewidth]{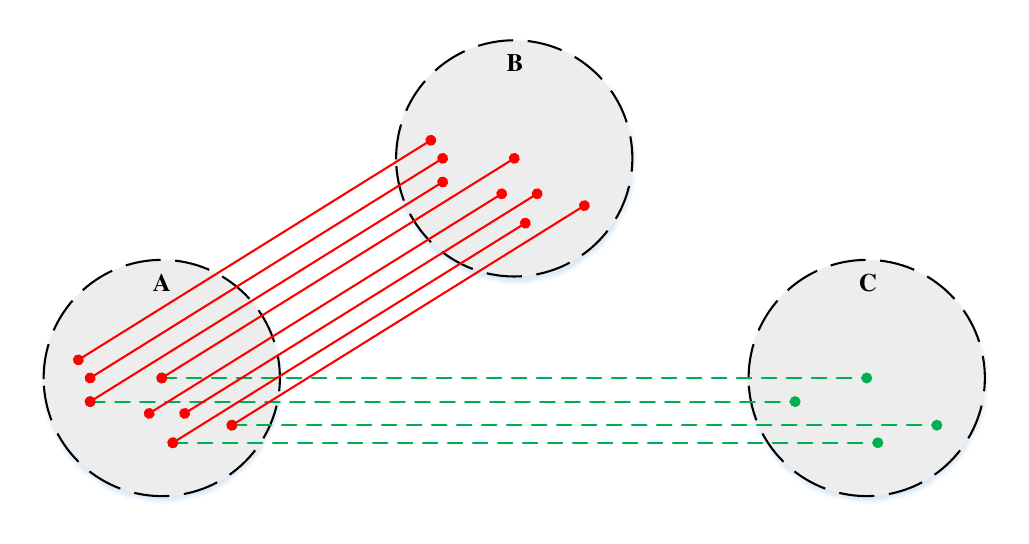}
	\caption{Illustration of a sampling set with multiple models.}
	\label{Fig5}
\end{figure}

In \textbf{Step1}, the sampling set $ {\bf{Sam}}_{\textbf{k}} $ is determined using spatial relationships. This method cannot guarantee that there is only a single model in the sampling set $ {\bf{Sam}}_{\textbf{k}} $. Assume that region A is moved to regions B and C after two duplications. Fig. \ref{Fig5} shows the sampling results when region A has the highest density. Clearly, the sampling result inevitably contains two models. However, the RANSAC algorithm can only identify a single model, in which case, the model represented by the green dashed line will be ignored. This may lead to the loss of some tampering areas in forgery localization stage. To overcome this problem, the \textbf{inlier removal strategy} is implemented. The most important role of this strategy is to change the number of models in a local region. Mathematically, the inlier removal strategy can be expressed as:
\begin{equation}
	\label{Eq25}
	{\bf{Unvisited}} = {\bf{Unvisited}} - {\bf{Inlier}},
\end{equation}
where, - denotes the set difference. The actual meaning of Equation (\ref{Eq25}) is that the set \textbf{Unvisited} only retains elements that do not belong to the set \textbf{Inlier}.

Finally, the source of the local homography matrix are distinguished. This paper notes that block-based algorithms usually set a minimum number of matches that satisfy the same affine transformation \cite{1,34} to reduce false alarms. This is because matches caused by tampering tend to be denser than matches caused by SGO's self-similarity. Therefore, under the premise of the excessive keypoint strategy, the \textbf{minimum number of inliers $ N_{in} $} is used to determine whether the local homography matrix $ {\bf{H}}_{\bf{k}} $ is generated by tampering, which satisfies:
\begin{equation}
	\label{Eq26}
	\left\| {{\bf{Inlier}}} \right\| > {N_{in}}.
\end{equation}

\textbf{Step3:} \textbf{\textit{Forgery Localization Using Robust Grayscale Statistical Information}}. In this step, the suspicious region is firstly divided, and then the suspicious region is verified to obtain the final localization result.

For determining suspicious regions, this paper uses the scale information of inliers \cite{22} to construct suspicious areas. Here, the suspicious areas formed by keypoints on the left and right sides of the inliers are defined as $ \textbf{SR}_{l} $ and $ \textbf{SR}_{r} $, respectively.

To accommodate various complex scenarios, a suspicious region verification method based on \textbf{robust grayscale statistical information} is proposed. In this paper, grayscale statistical information set \textbf{diff} is defined as:
\begin{equation}
	\label{Eq27}
	\begin{array}{ll}
	{\bf{diff}} &= \{ {I_{gray}}({{\bf{k}}_{{l_i}}}) - {I_{gray}}({{\bf{k}}_{{l_i}}})| \\
& \qquad \forall  < {{\bf{k}}_{{l_i}}},{{\bf{k}}_{{r_i}}} >  \in {\bf{Inlier}}\}
	\end{array},
\end{equation}
where, \textbf{diff} means the matching grayscale difference. Normally, the grayscale difference of matches and the grayscale differences at corresponding coordinates of the tampered area follow an independent and identical distribution. Therefore, suspicious areas can be verified using grayscale differences. In order to reduce the impact of noise, the Inter Quartile Range (IQR) is further used for \textbf{diff} to obtain a robust grayscale difference set \textbf{Diff}, which is defined as:
\begin{equation}
	\label{Eq28}
	\begin{array}{ll}
		{\bf{Diff}} &= \{ di{f_i}| \forall di{f_i} \in {\bf{diff}}, \\ & \qquad di{f_i} \in [{Q_1} - 1.5IQR,{Q_3} + 1.5IQR]\}
	\end{array},
\end{equation}
where, $ Q_{1} $ and $ Q_{3} $ represent the first quartile and the third quartile of the data, respectively. IQR is equivalent to $ Q_{3} - Q_{1} $.

Subsequently, the verification threshold of grayscale difference is determined as:
\begin{equation}
	\label{Eq29}
	D_{l} = \min ({\bf{Diff}}),
\end{equation}
\begin{equation}
	\label{Eq30}
	D_{h} = \max ({\bf{Diff}}),
\end{equation}

Then, the suspicious areas $ \textbf{SR}_{l} $ and $ \textbf{SR}_{r} $ are verified, and the corresponding localization result $ \textbf{map}_{l} $ and $ \textbf{map}_{r} $ are obtained, which can be expressed as:
\begin{equation}
	\label{Eq31}
	\begin{array}{ll}
		{\bf{ma}}{{\bf{p}}_l} & = \{  < {\bf{k}},{\bf{k}}' > |\forall {\bf{k}} \in {\bf{S}}{{\bf{R}}_l}, \\
		& \qquad {I_{gray}}({\bf{k}}) - {I_{gray}}({\bf{k}}') \in [{D_l},{D_h}]\}
	\end{array},
\end{equation}
\begin{equation}
	\label{Eq32}
	\begin{array}{ll}
		{\bf{ma}}{{\bf{p}}_r} & = \{  < {\bf{k}}'',{\bf{k}} > |\forall {\bf{k}} \in {\bf{S}}{{\bf{R}}_r}, \\
		& \qquad {I_{gray}}({\bf{k}}'') - {I_{gray}}({\bf{k}}) \in [{D_l},{D_h}]\}
	\end{array},
\end{equation}
where, $ {\bf{k}}' = {{\bf{H}}_{\bf{k}}} \cdot {\bf{k}} $; $ {\bf{k}}'' = {\bf{H}}_{\bf{k}}^{ - 1} \cdot {\bf{k}} $.

Finally, the localization result \textbf{map} obtained in this iteration is defined as:
\begin{equation}
	\label{Eq33}
	{\bf{map}} = {\bf{ma}}{{\bf{p}}_l}\bigcup {{\bf{ma}}{{\bf{p}}_r}}.
\end{equation}

In the new iterative forgery localization, \textbf{Step1} to \textbf{Step3} will continuously iteratively generate localization results, and these results will be continuously merged to form the final binary positioning result.

\section{Experiment results}\label{experiment results}
In this section, a series of experiments are conducted to evaluate our work, and then compare it with state-of-the-art methods. All the experiments are done using MATLAB R2018a under Microsoft Windows 11. The PC used for testing has 2.30 GHz CPU and 16 GB RAM.

\subsection{Datasets}

\begin{figure*}[ht]
	\centering
	\begin{tabular}{cccccc}
		\includegraphics[width=0.13\linewidth]{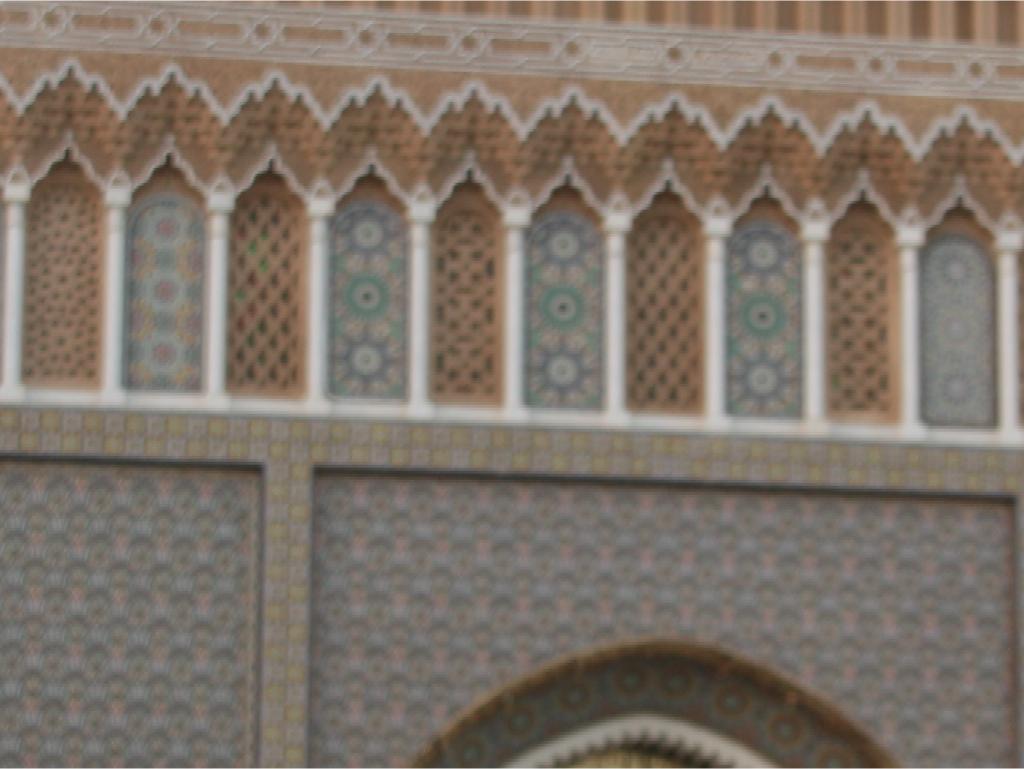} &
		\includegraphics[width=0.13\linewidth]{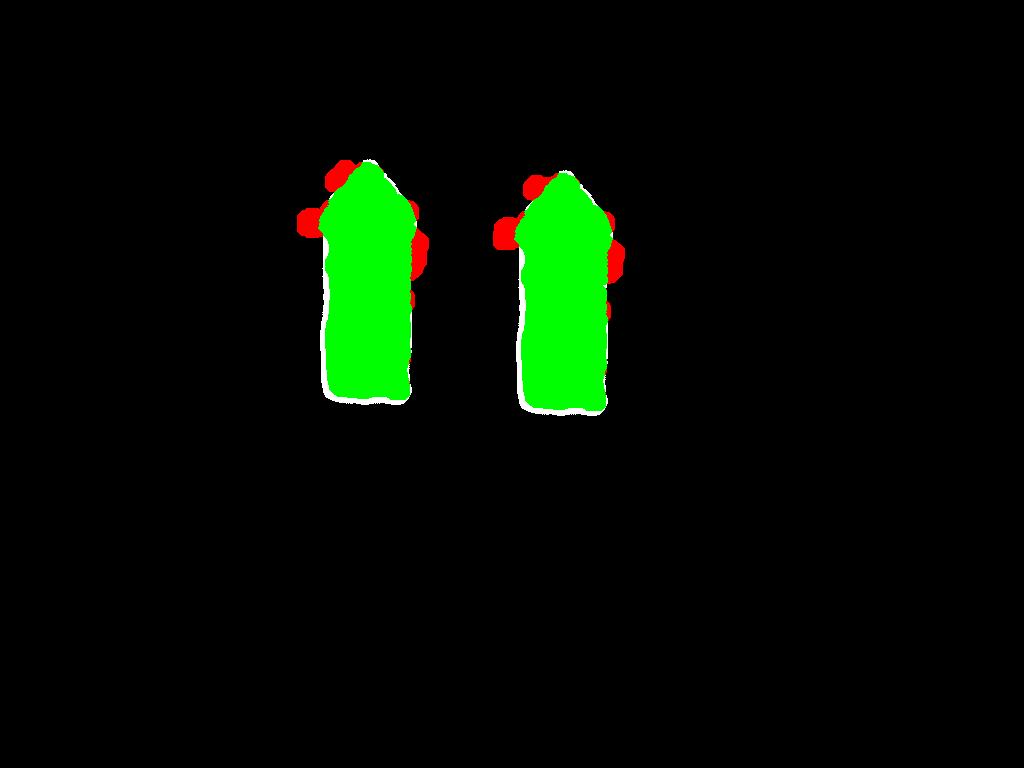} &
		\includegraphics[width=0.13\linewidth]{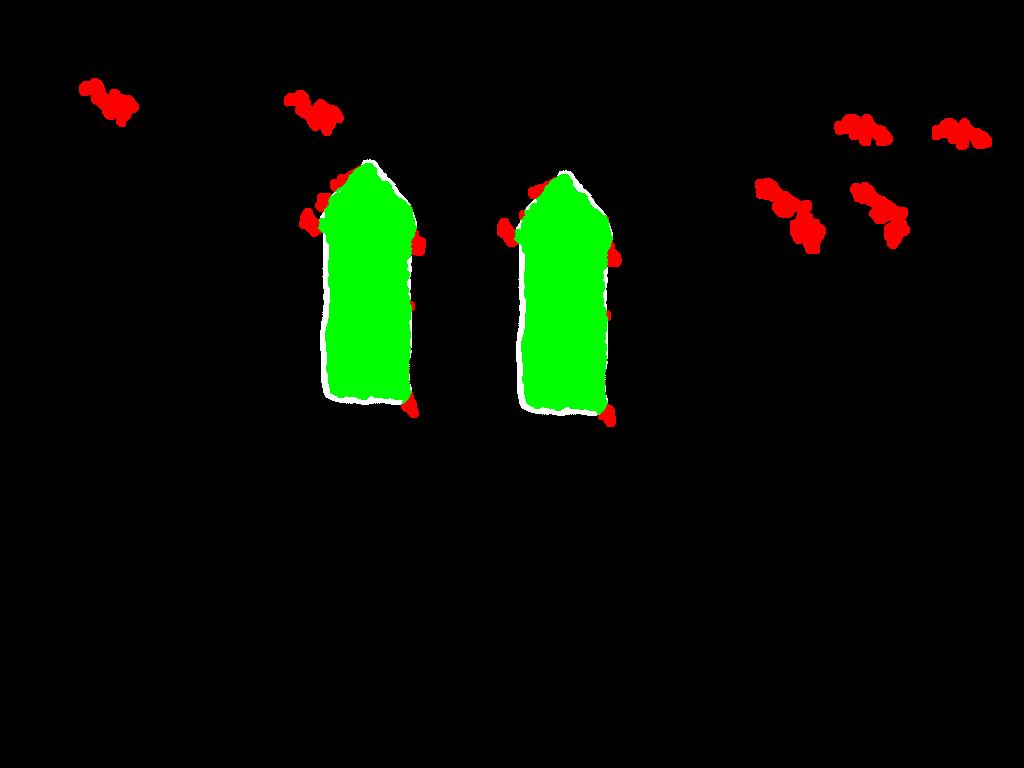} &
		\includegraphics[width=0.13\linewidth]{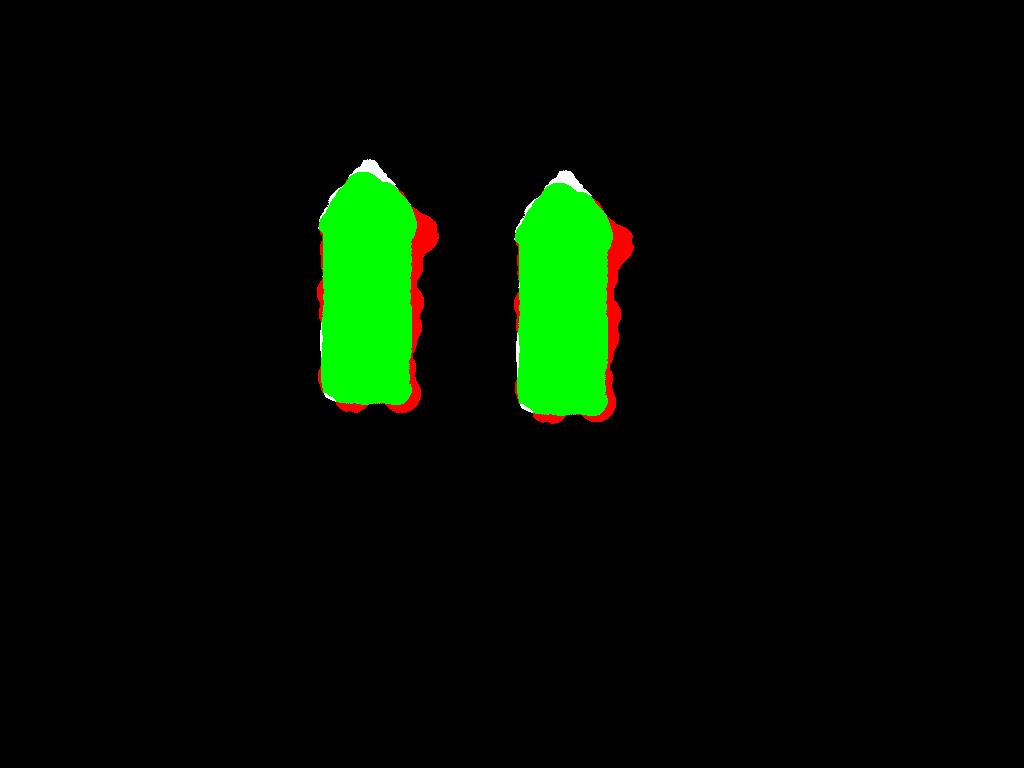} &
		\includegraphics[width=0.13\linewidth]{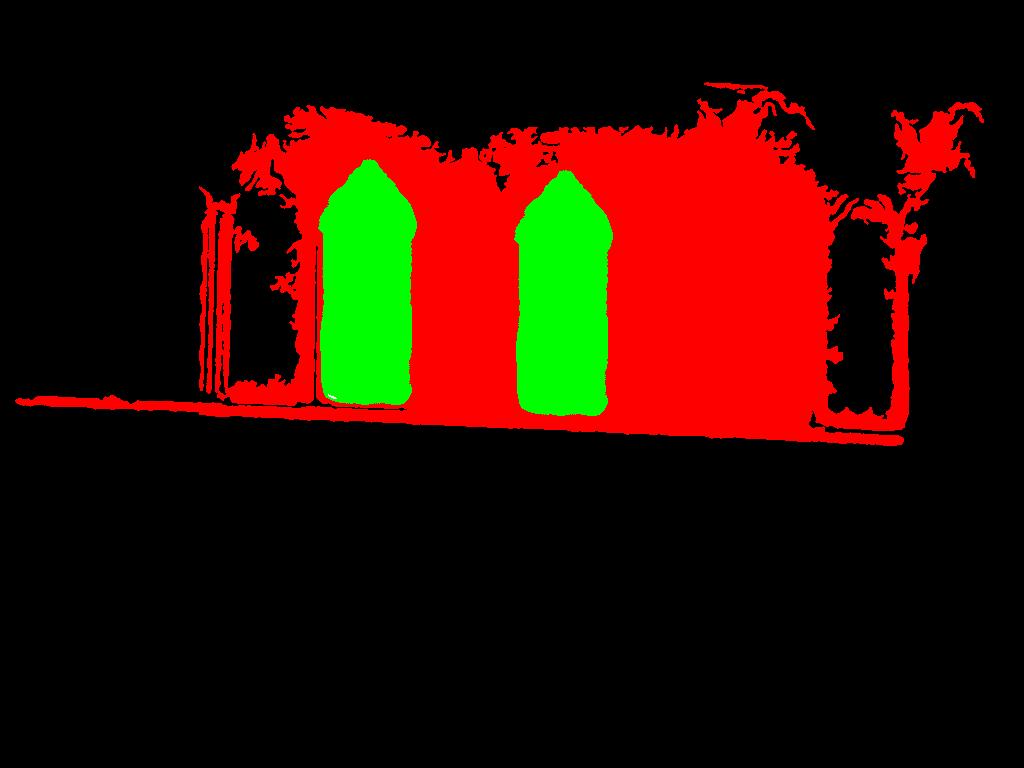} &
		\includegraphics[width=0.13\linewidth]{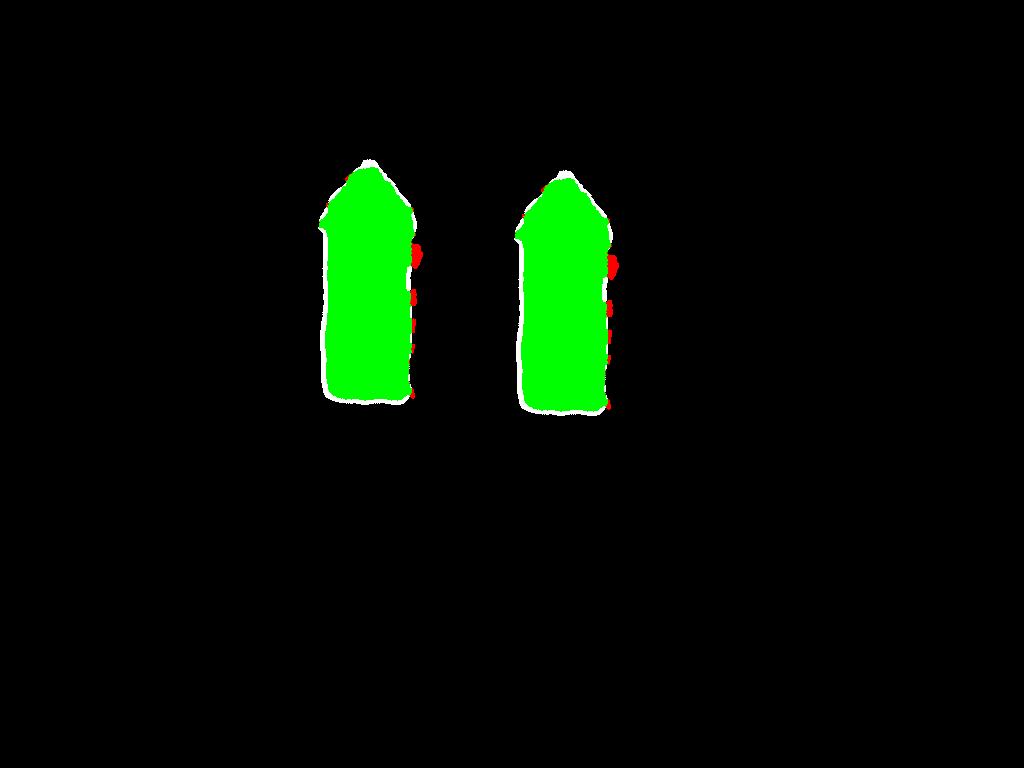} \\
		(a1) & (b1) & (c1) & (d1) & (e1) & (f1) \\
		\includegraphics[width=0.13\linewidth]{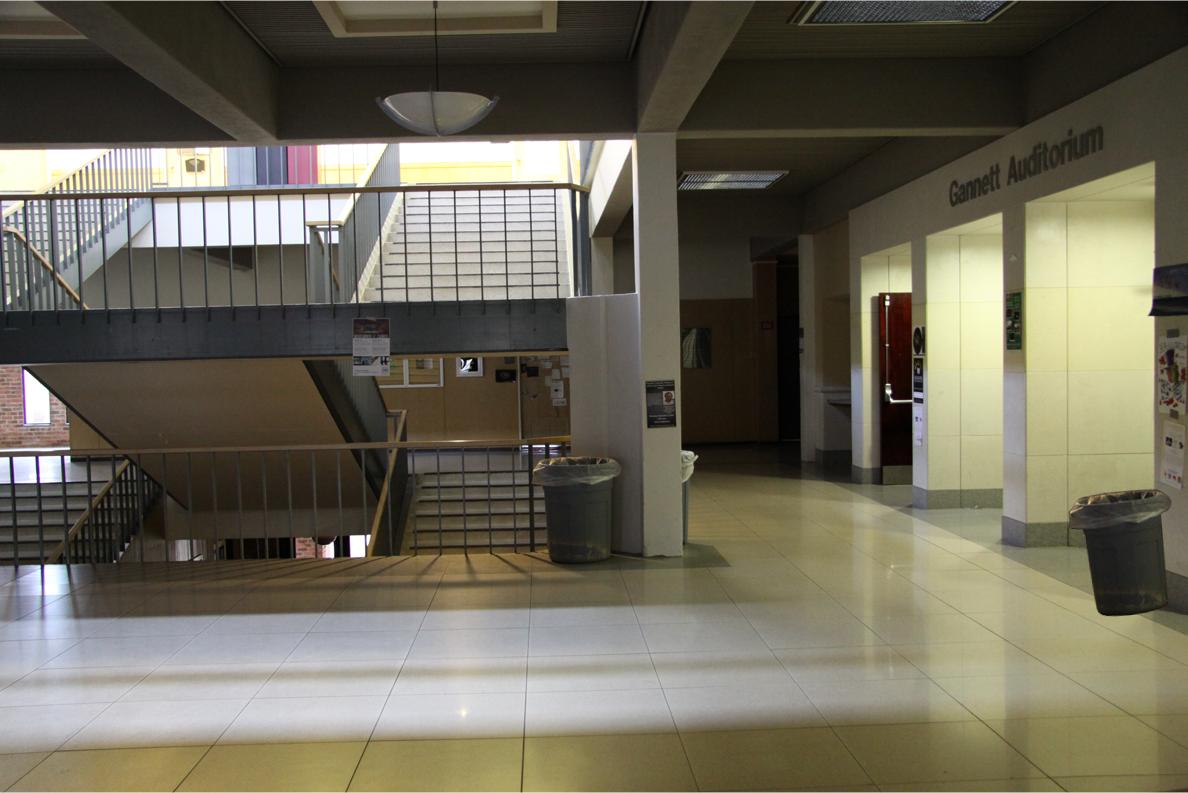} &
		\includegraphics[width=0.13\linewidth]{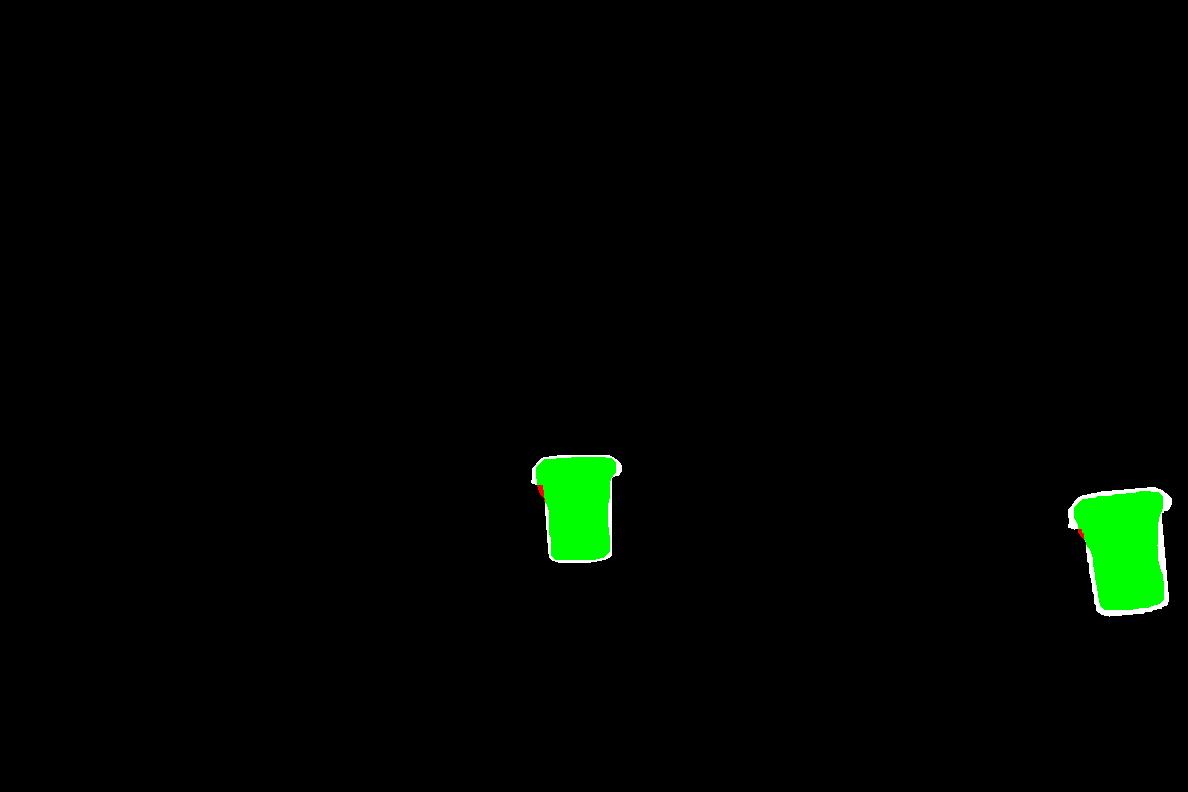} &
		\includegraphics[width=0.13\linewidth]{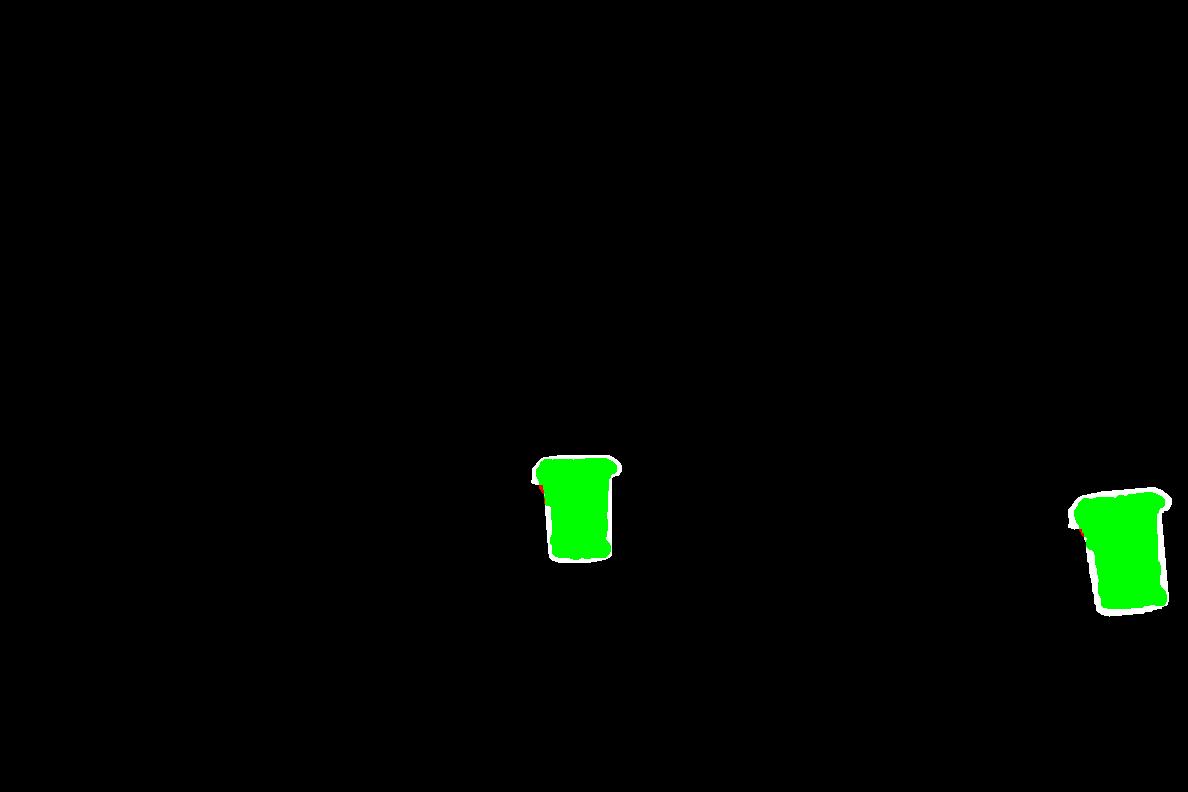} &
		\includegraphics[width=0.13\linewidth]{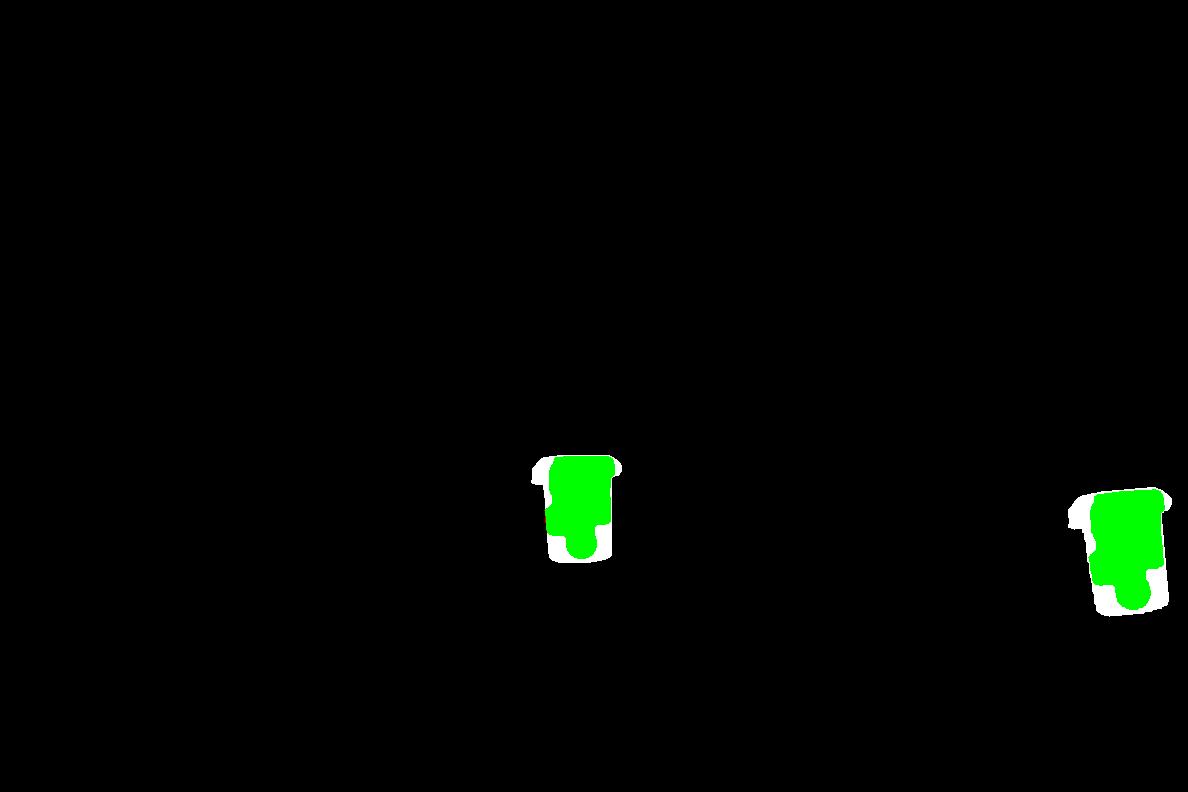} &
		\includegraphics[width=0.13\linewidth]{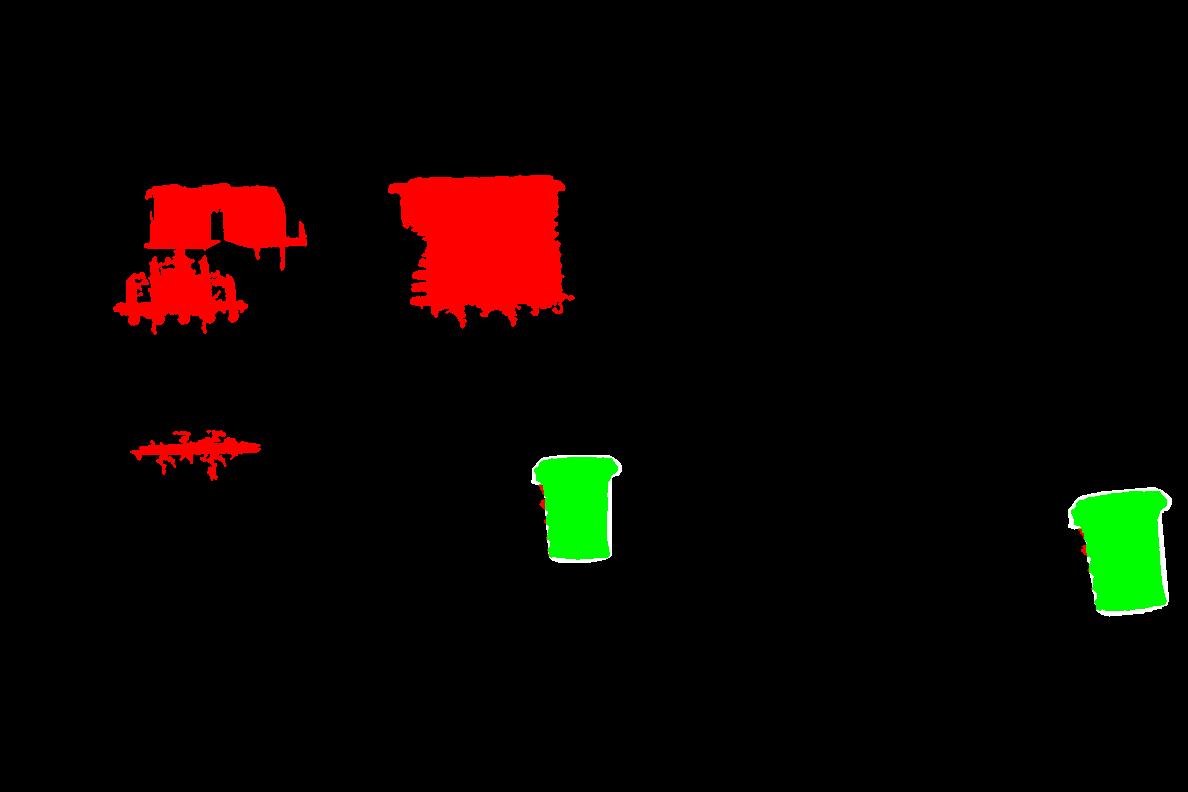} &
		\includegraphics[width=0.13\linewidth]{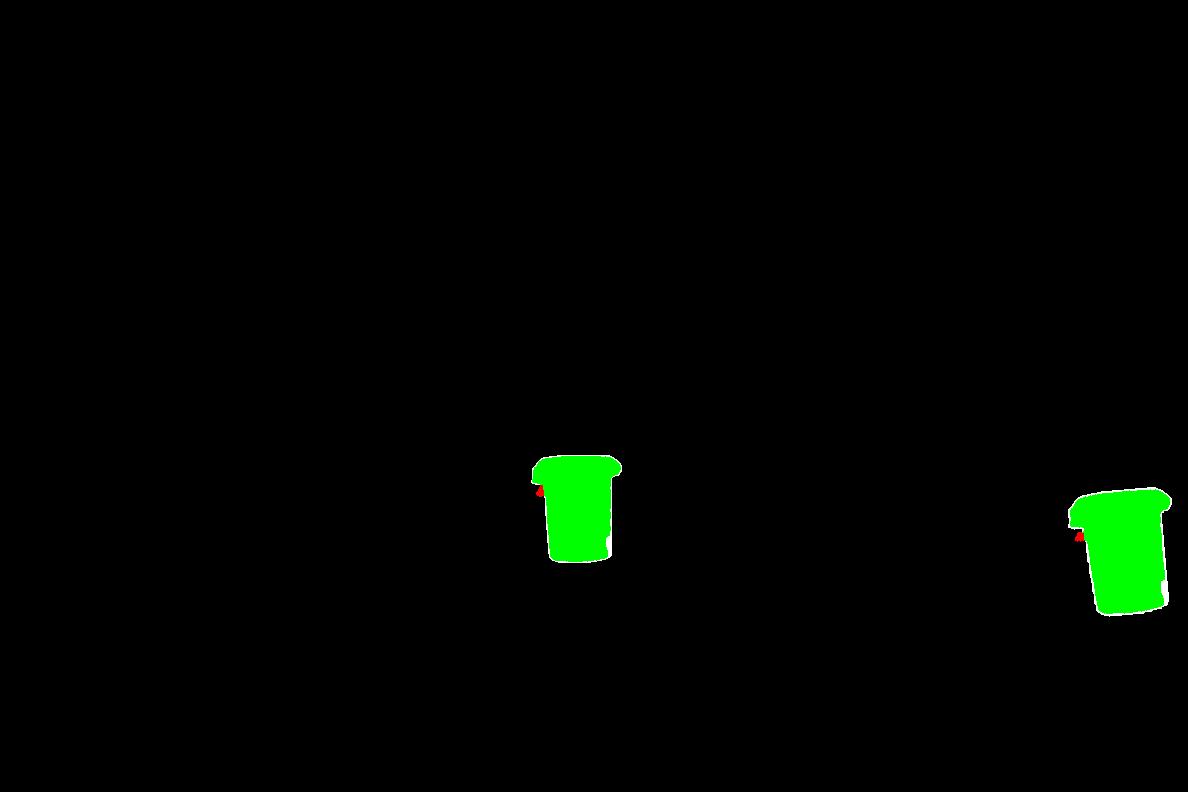} \\
		(a2) & (b2) & (c2) & (d2) & (e2) & (f2) \\
		\includegraphics[width=0.13\linewidth]{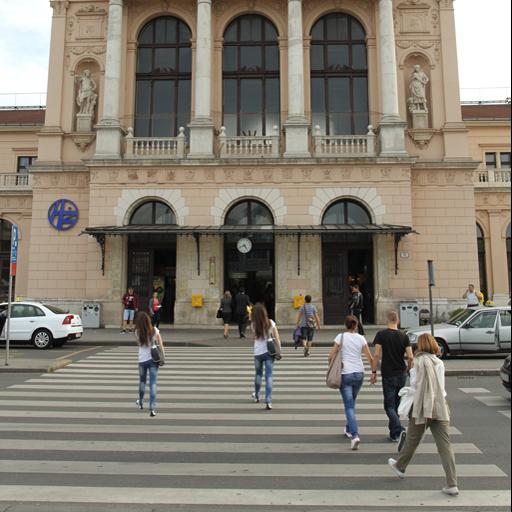} &
		\includegraphics[width=0.13\linewidth]{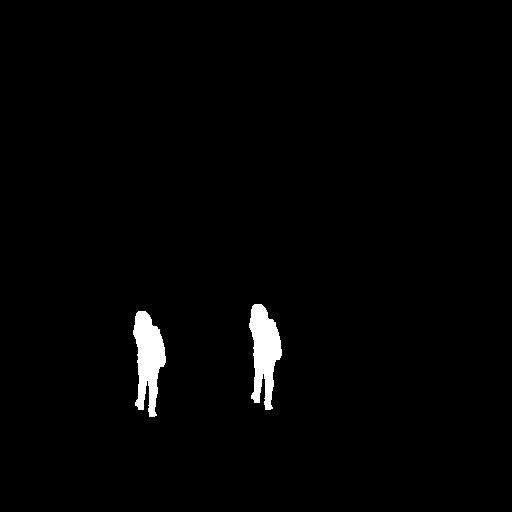} &
		\includegraphics[width=0.13\linewidth]{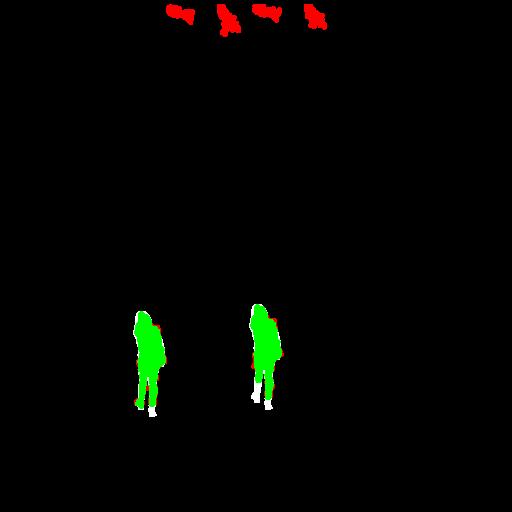} &
		\includegraphics[width=0.13\linewidth]{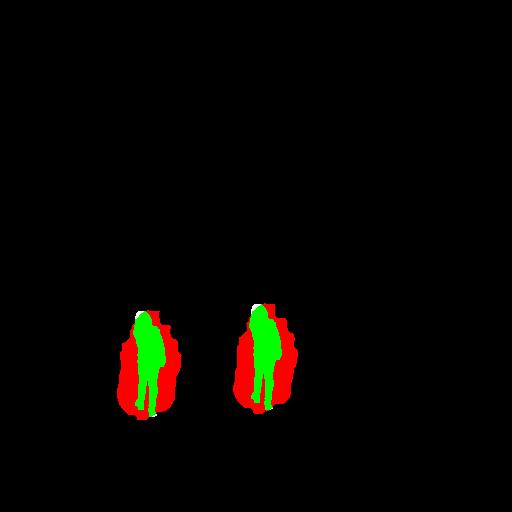} &
		\includegraphics[width=0.13\linewidth]{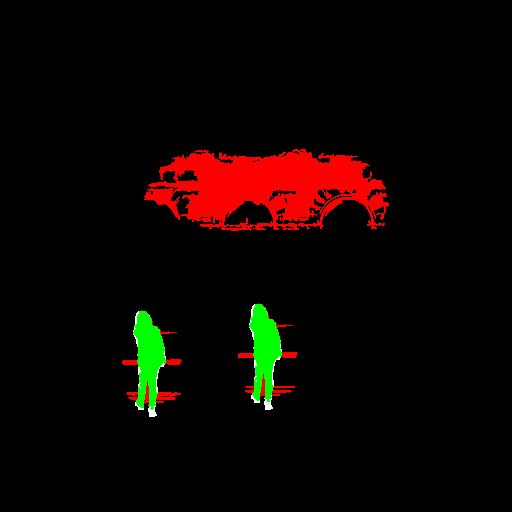} &
		\includegraphics[width=0.13\linewidth]{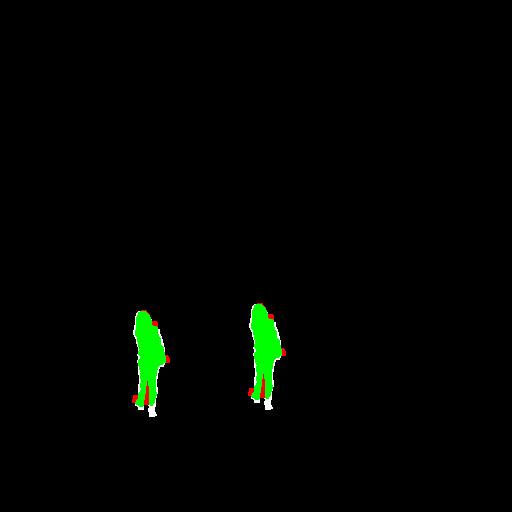} \\
		(a3) & (b3) & (c3) & (d3) & (e3) & (f3) \\
		\includegraphics[width=0.13\linewidth]{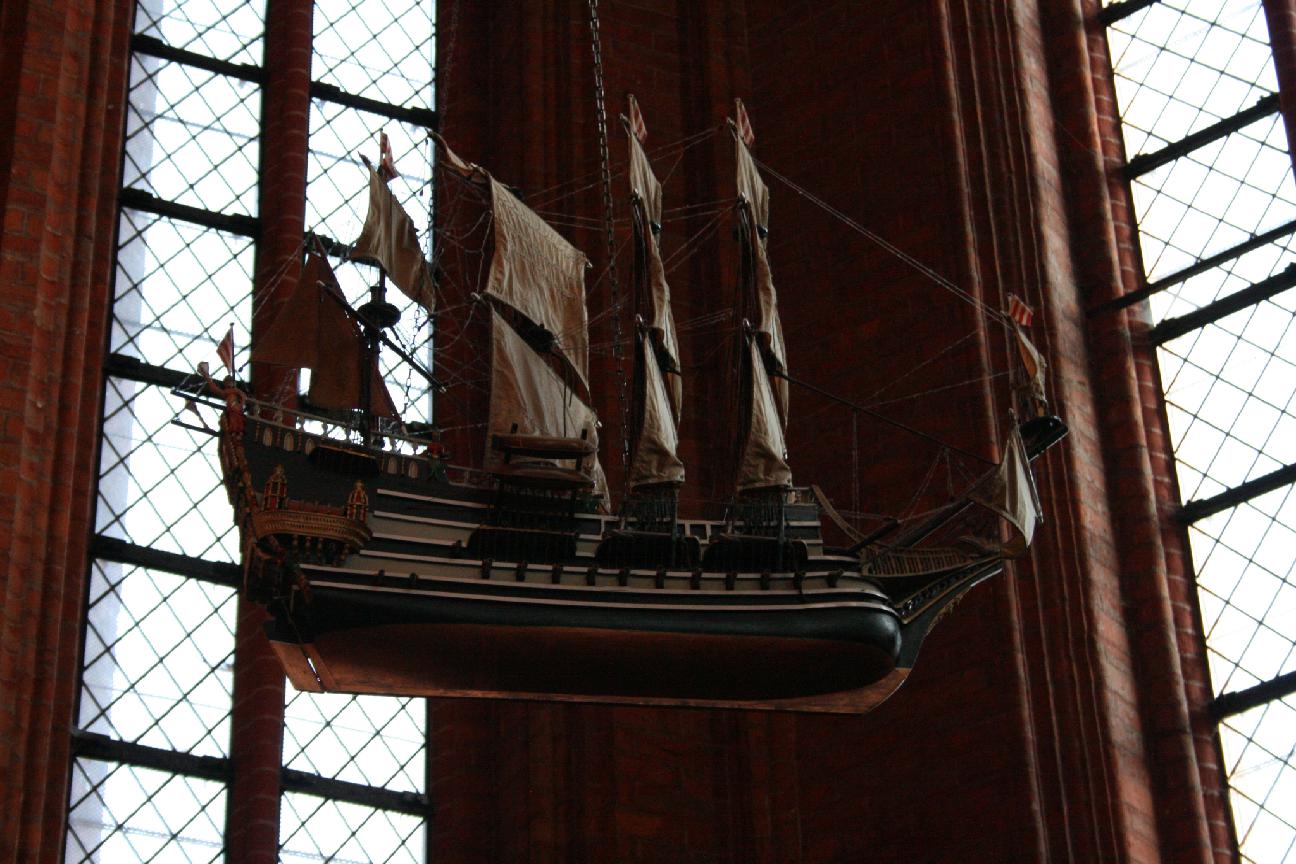} &
		\includegraphics[width=0.13\linewidth]{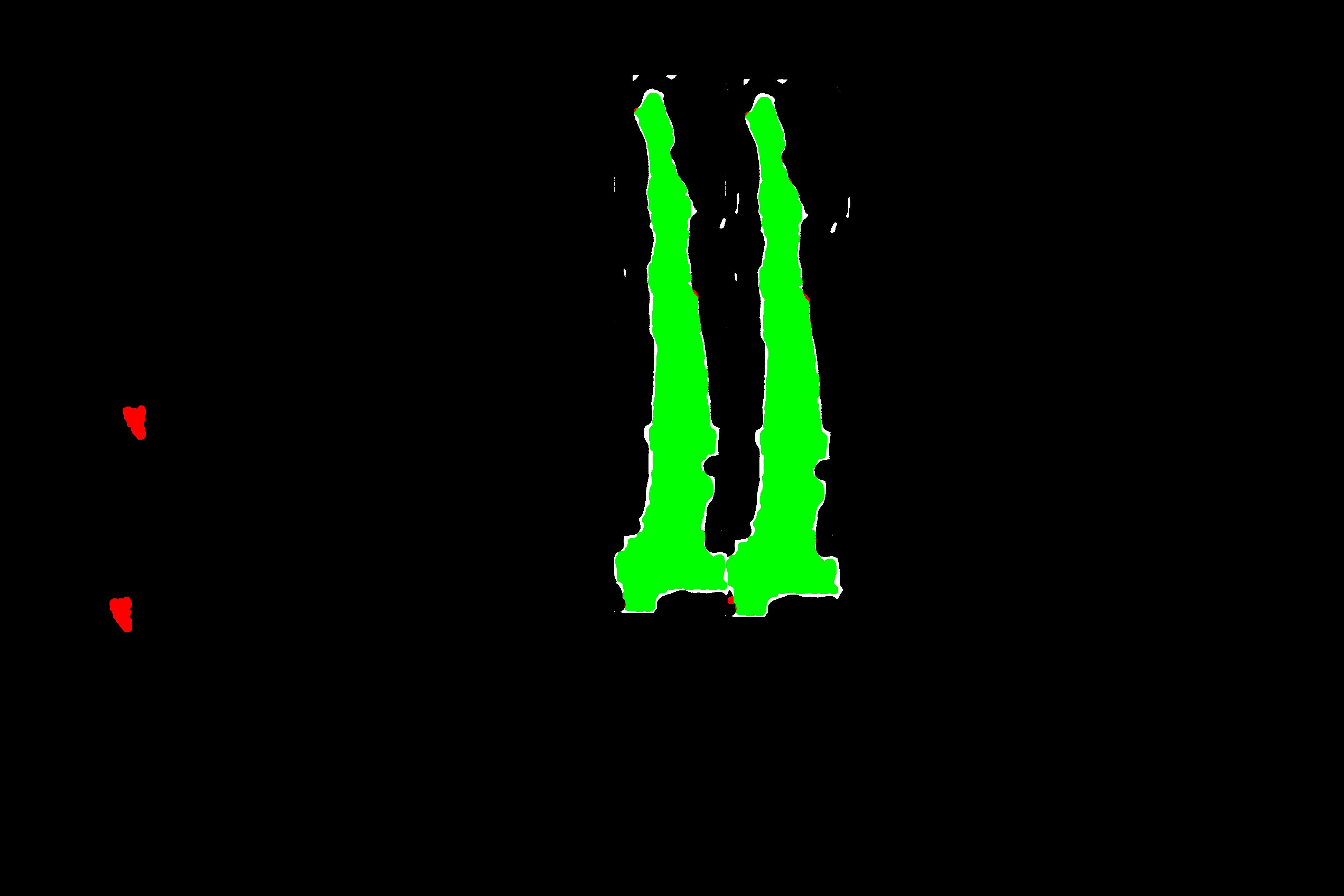} &
		\includegraphics[width=0.13\linewidth]{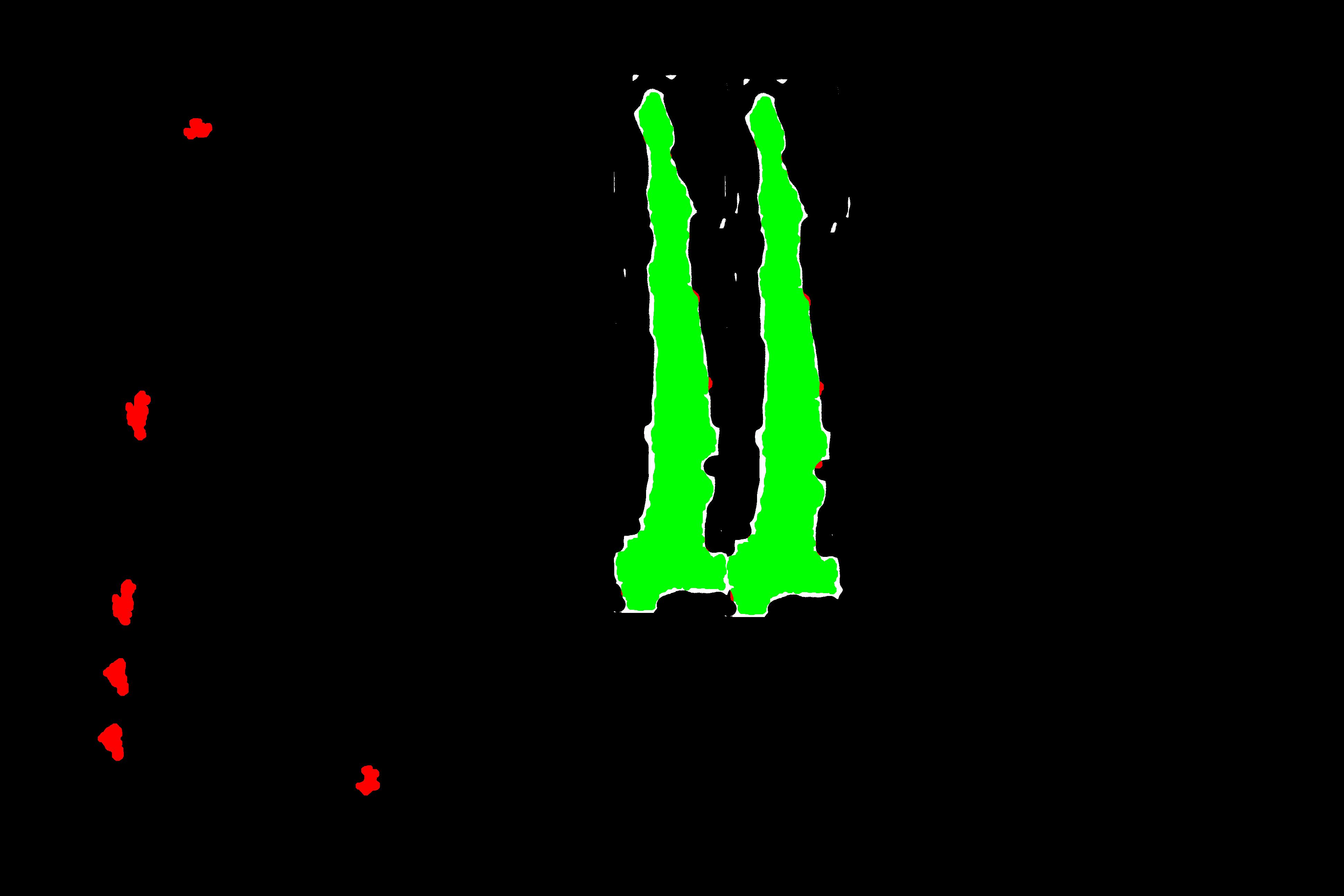} &
		\includegraphics[width=0.13\linewidth]{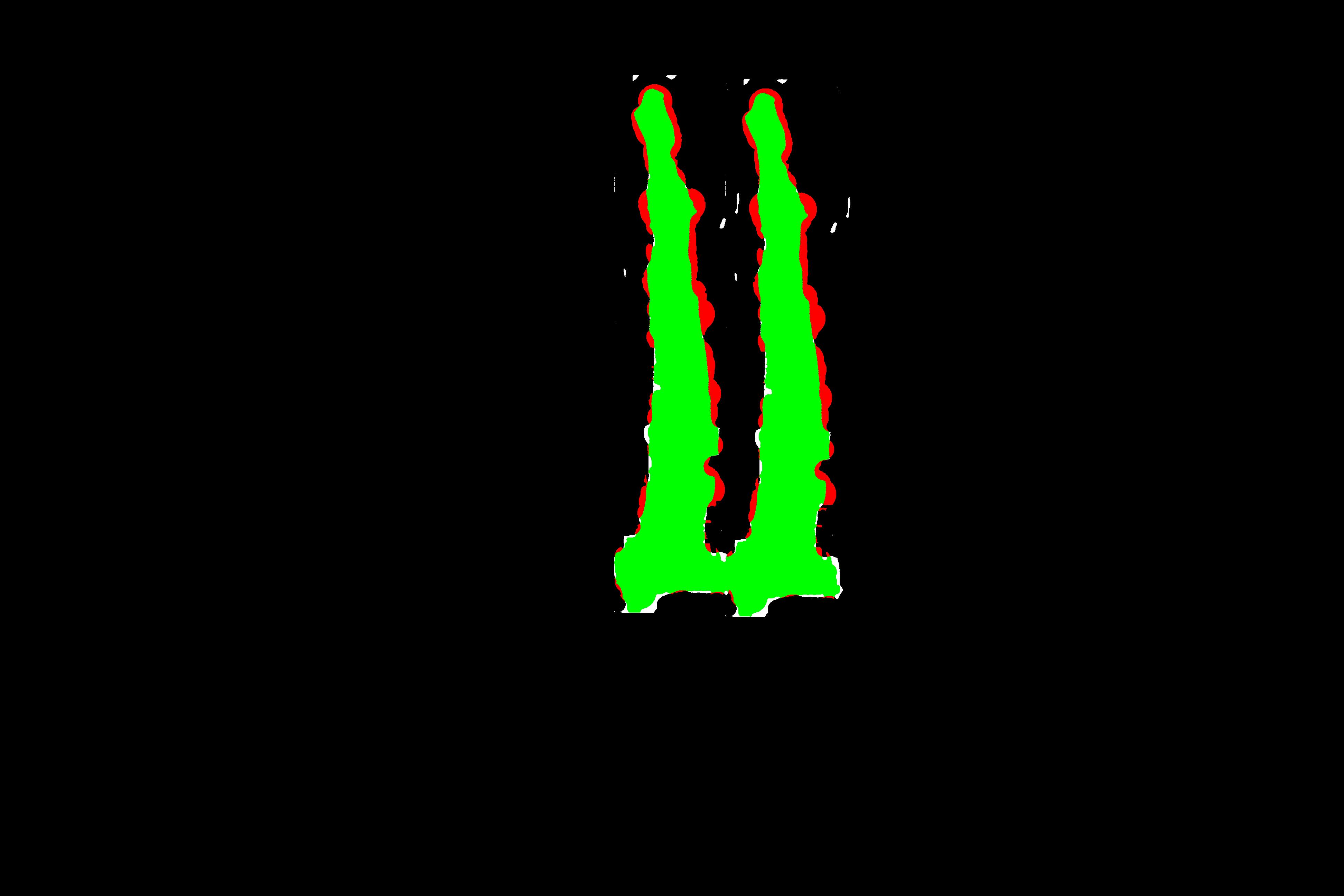} &
		\includegraphics[width=0.13\linewidth]{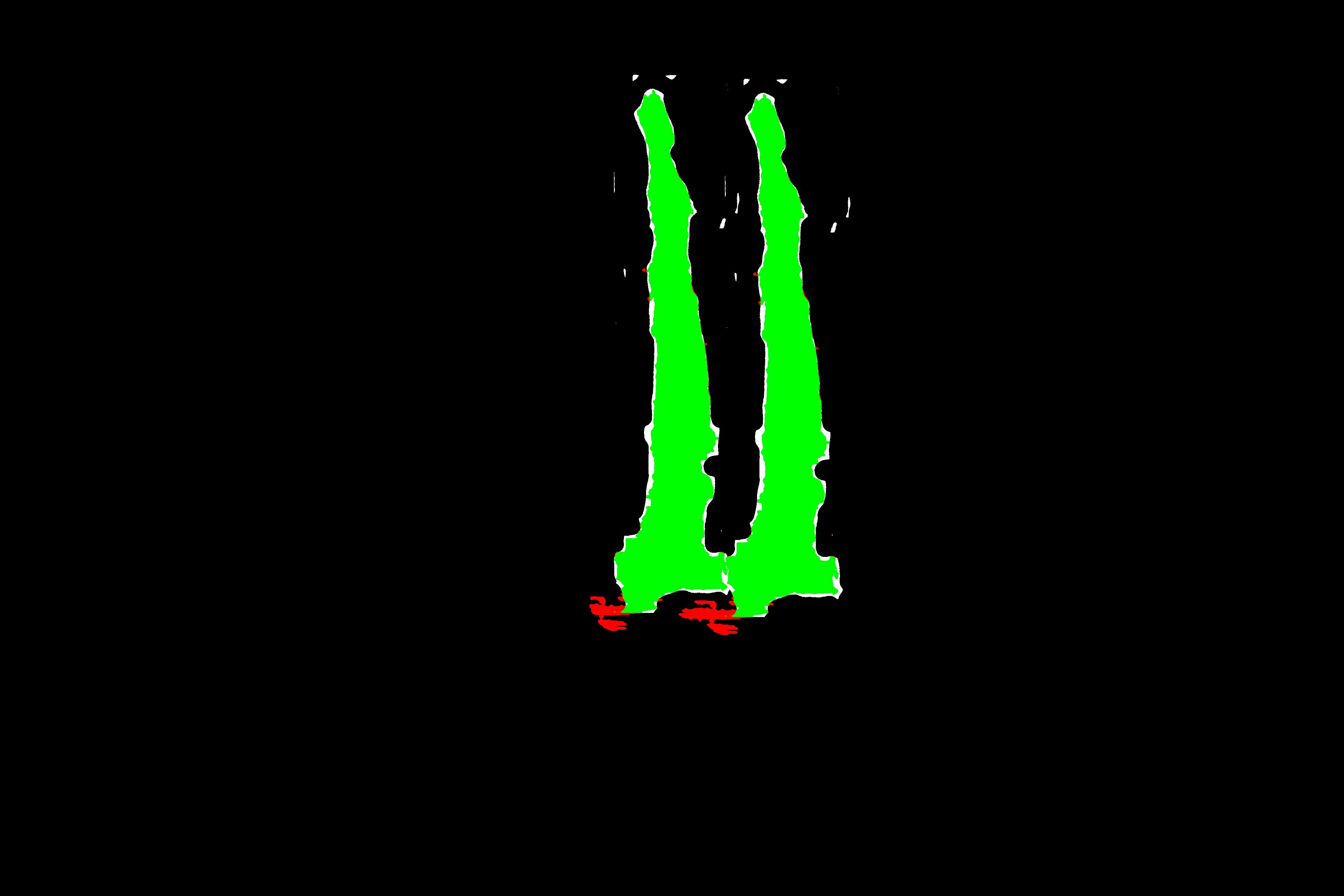} &
		\includegraphics[width=0.13\linewidth]{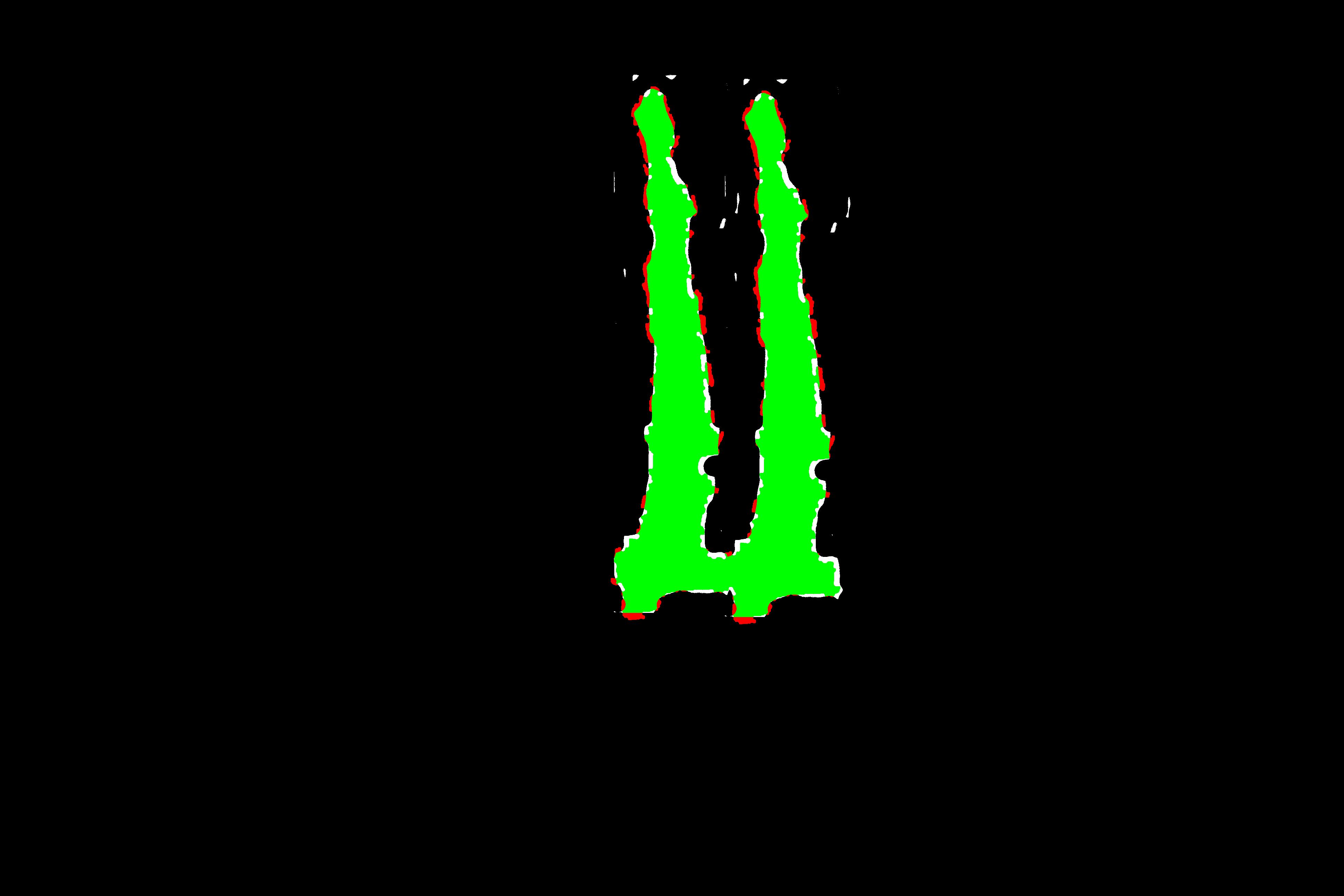} \\
		(a4) & (b4) & (c4) & (d4) & (e4) & (f4) \\
		\includegraphics[width=0.13\linewidth]{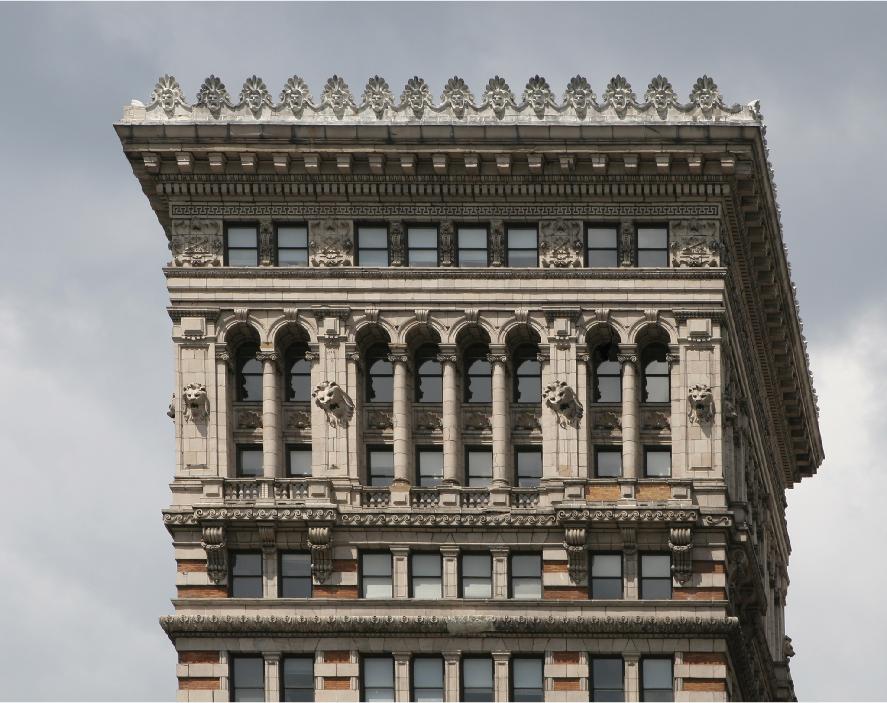} &
		\includegraphics[width=0.13\linewidth]{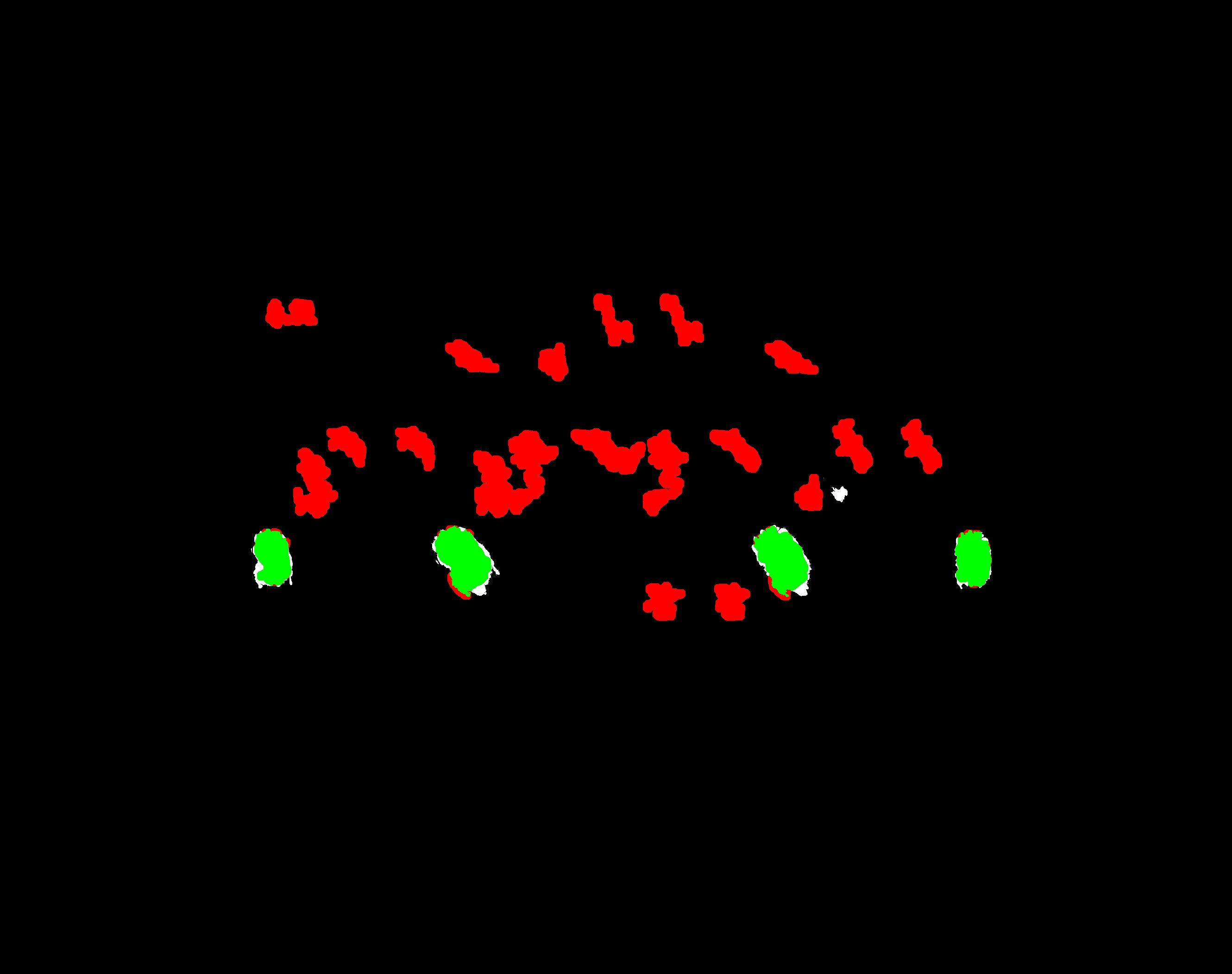} &
		\includegraphics[width=0.13\linewidth]{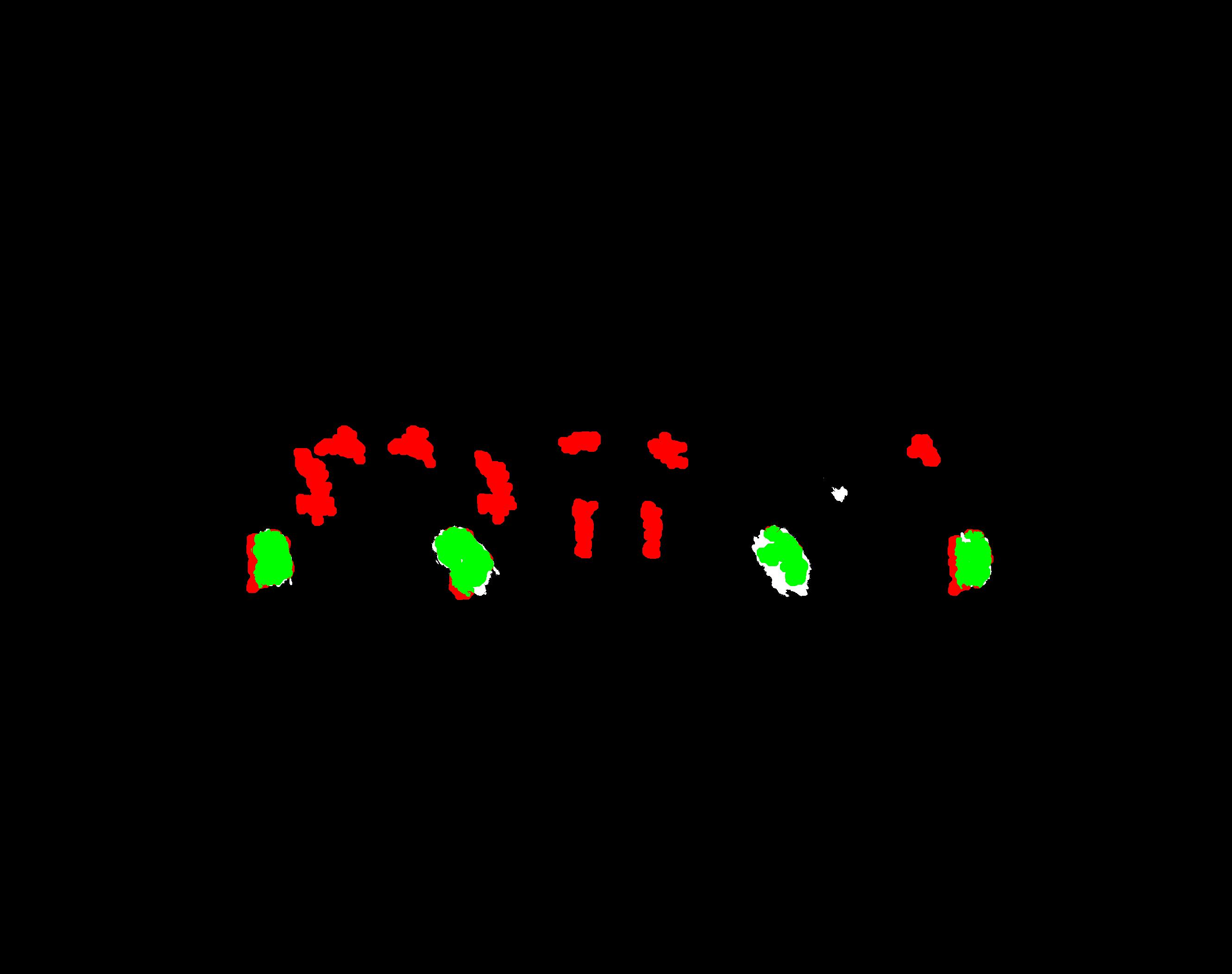} &
		\includegraphics[width=0.13\linewidth]{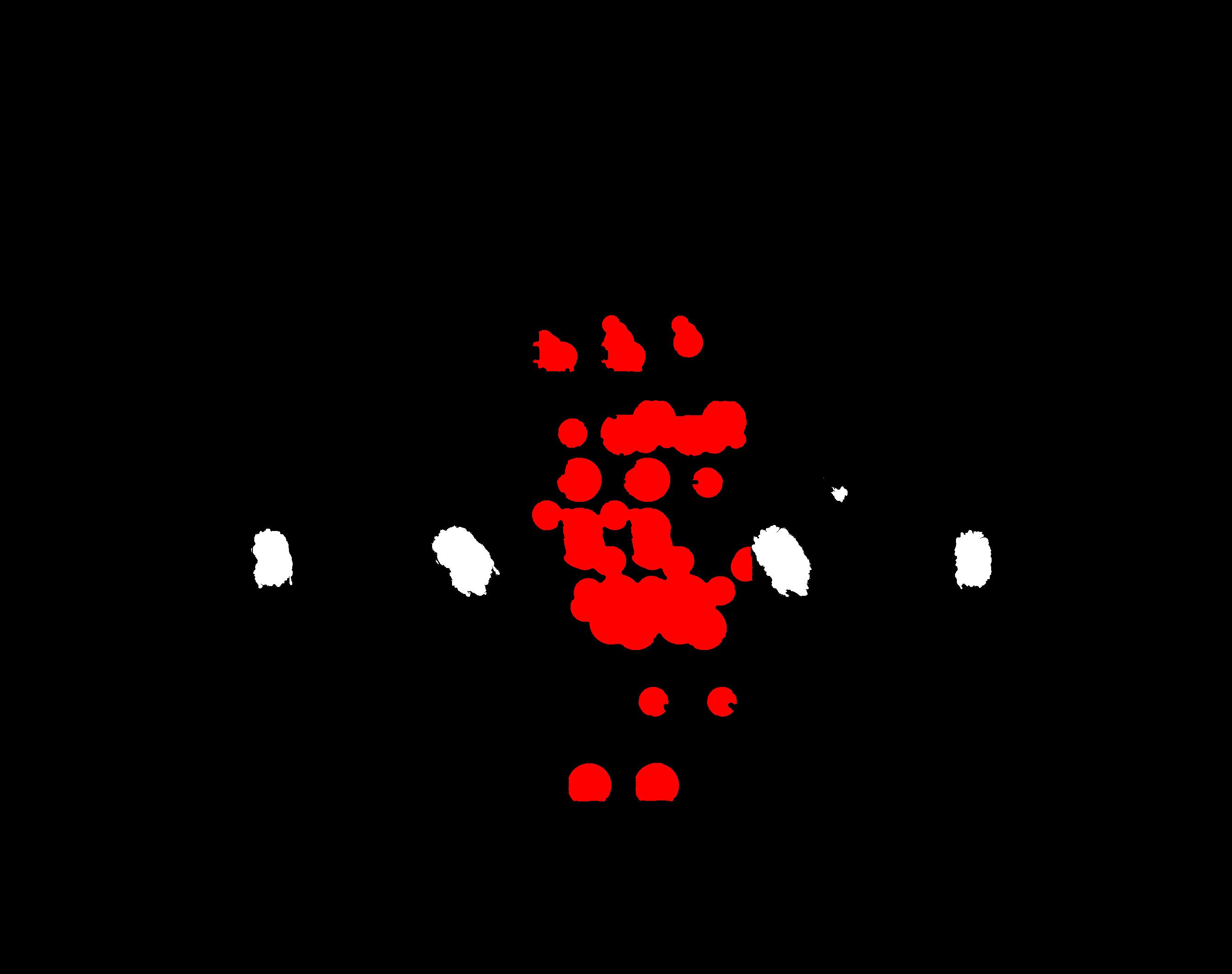} &
		\includegraphics[width=0.13\linewidth]{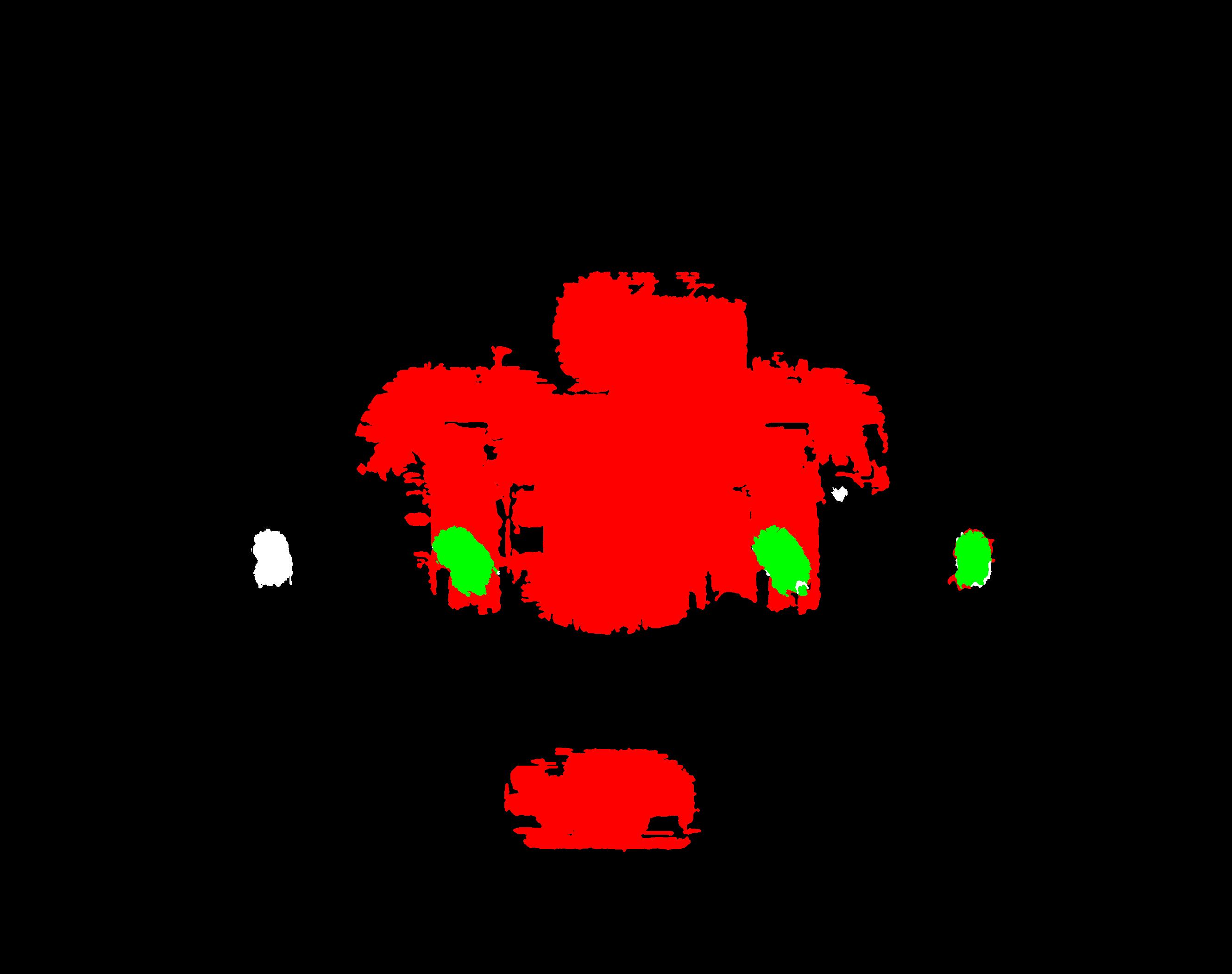} &
		\includegraphics[width=0.13\linewidth]{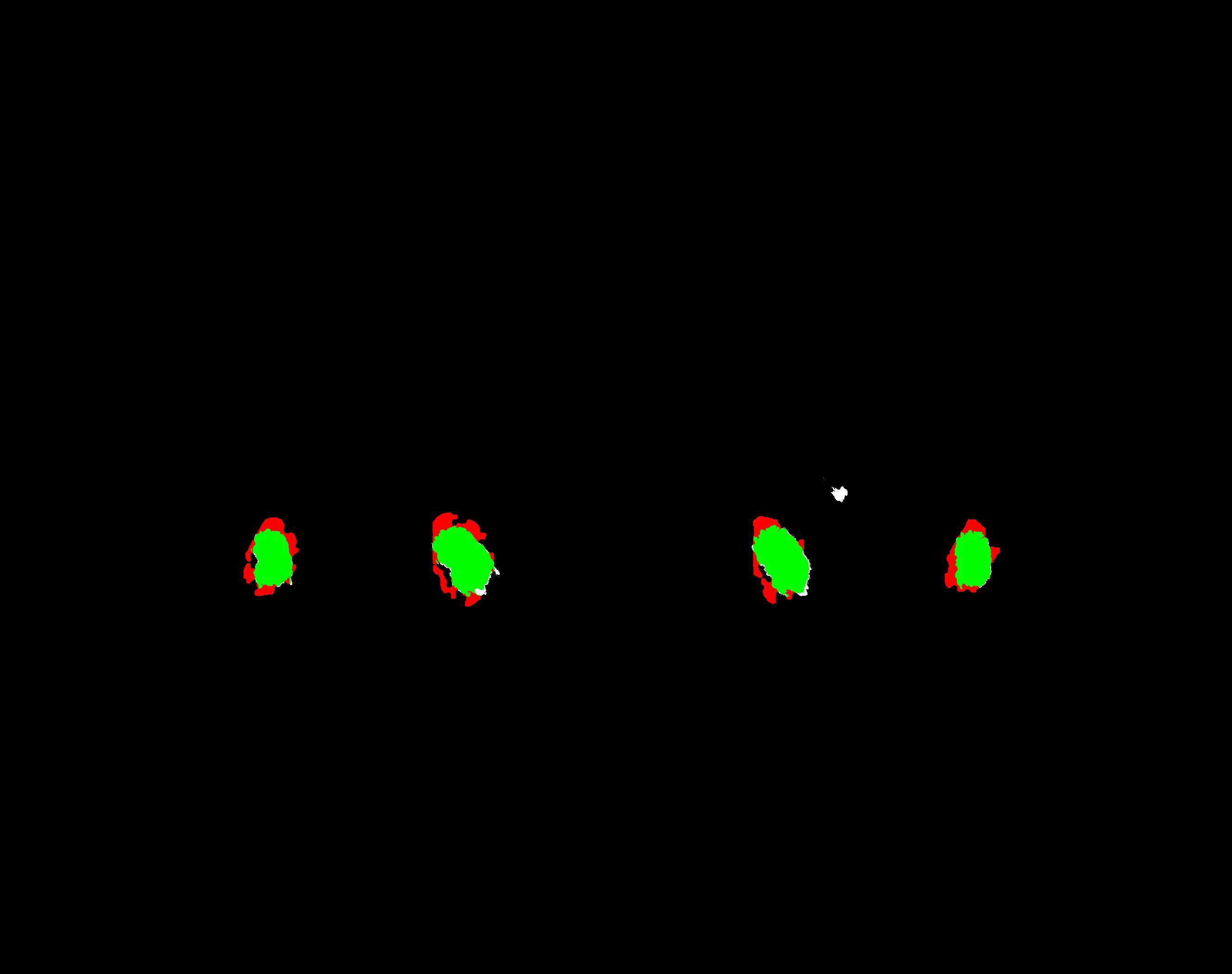} \\
		(a5) & (b5) & (c5) & (d5) & (e5) & (f5) \\
		\includegraphics[width=0.13\linewidth]{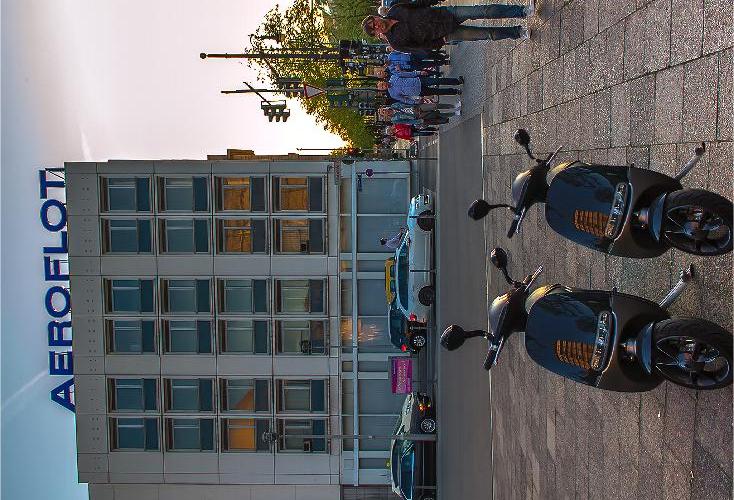} &
		\includegraphics[width=0.13\linewidth]{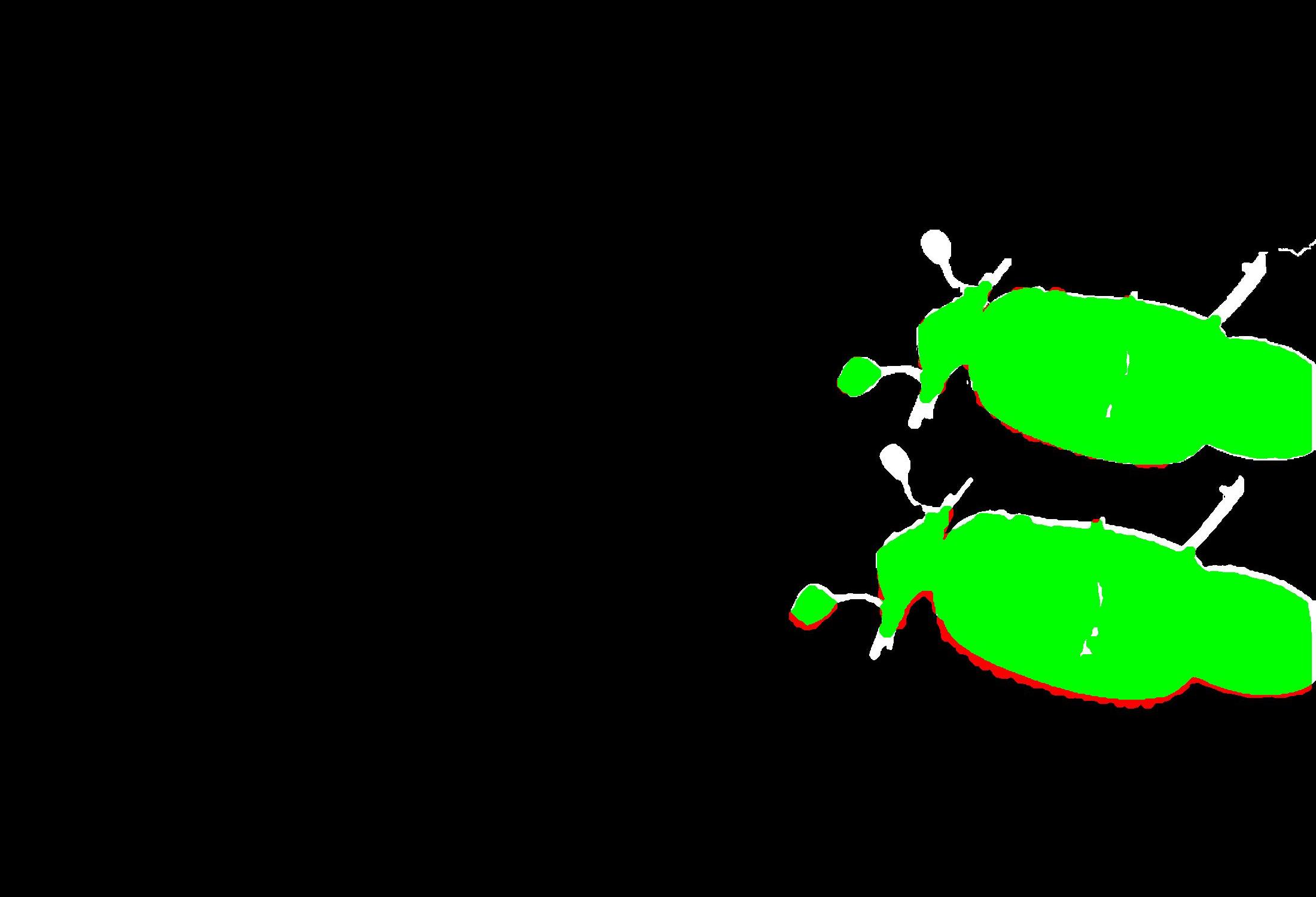} &
		\includegraphics[width=0.13\linewidth]{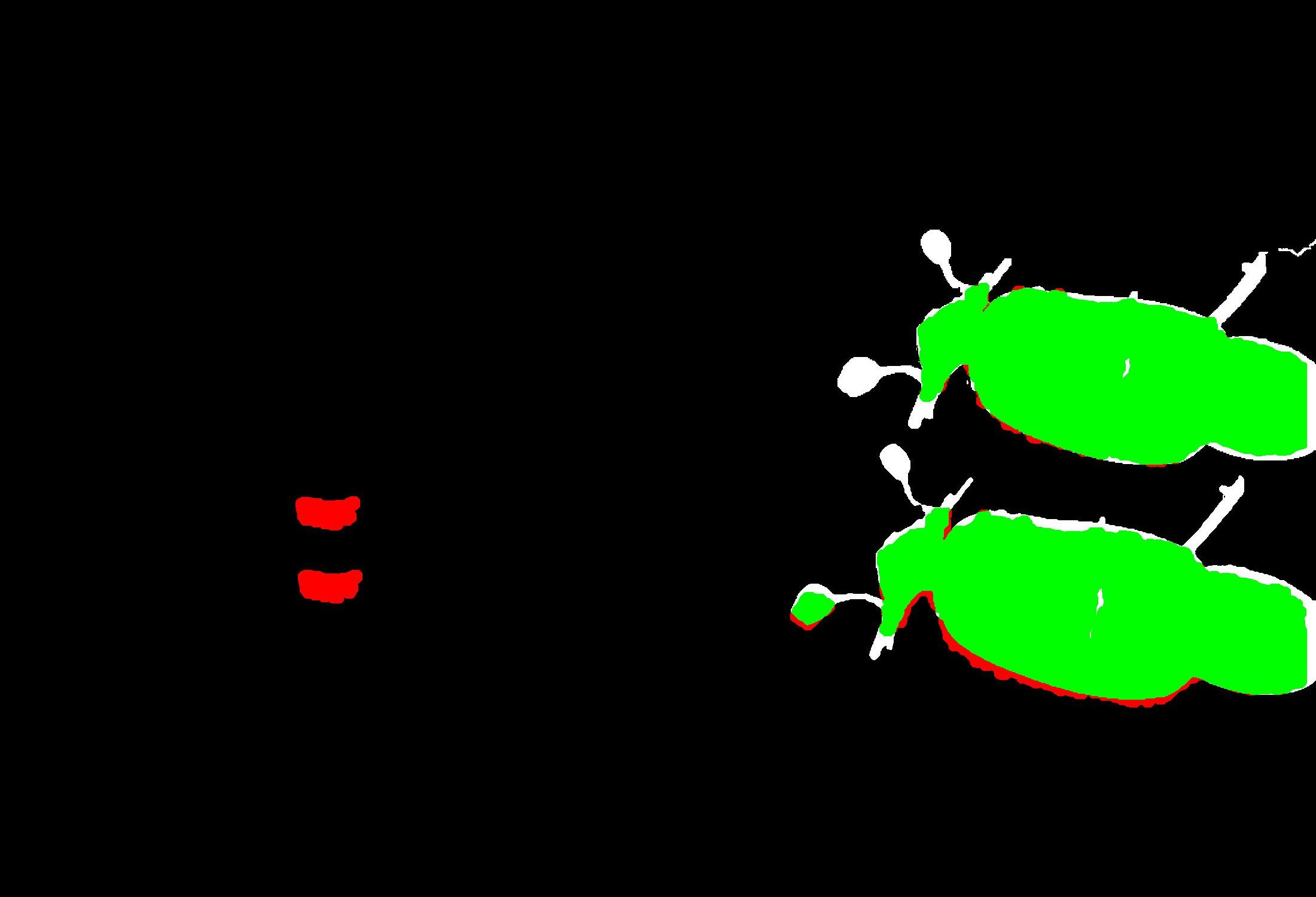} &
		\includegraphics[width=0.13\linewidth]{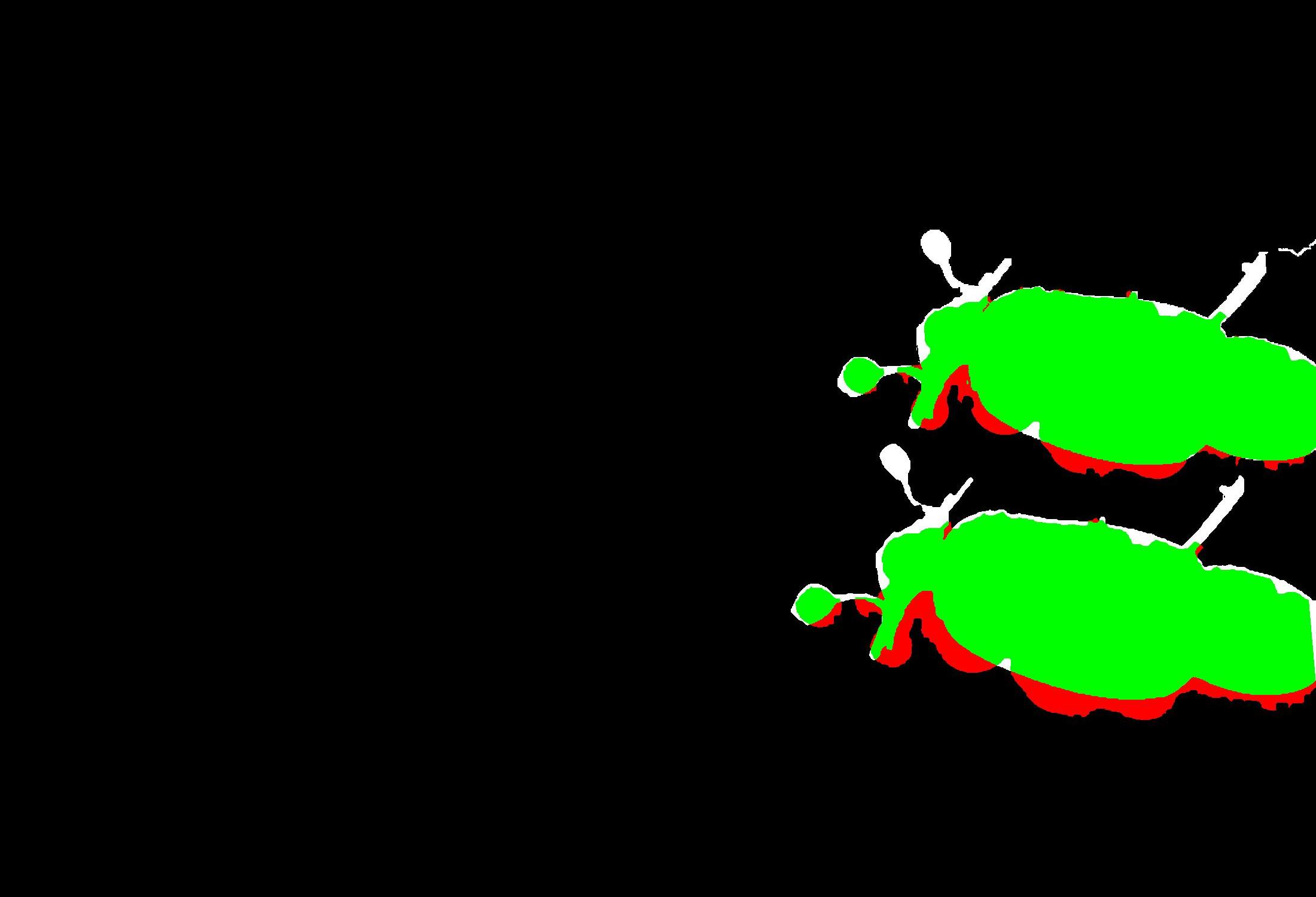} &
		\includegraphics[width=0.13\linewidth]{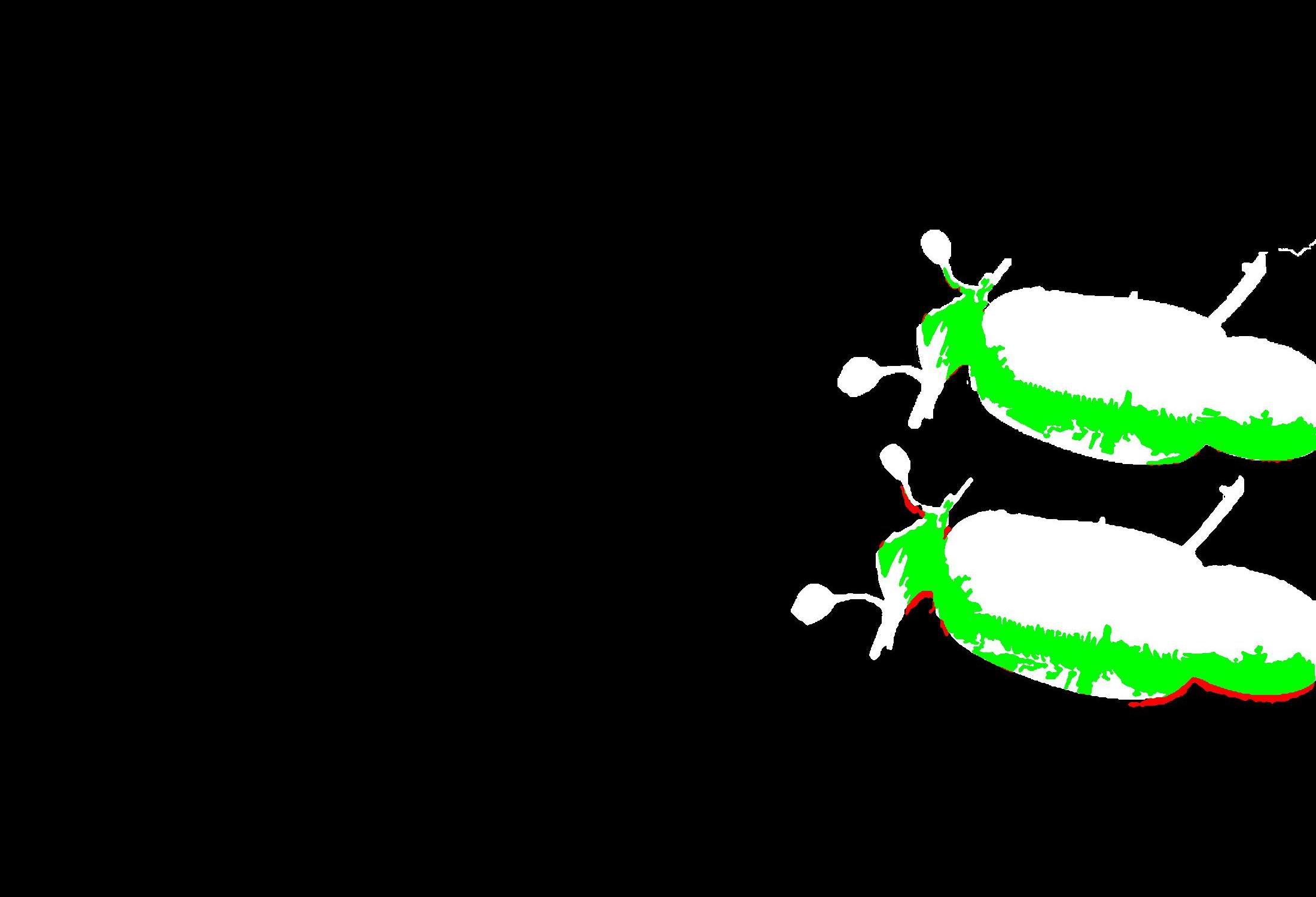} &
		\includegraphics[width=0.13\linewidth]{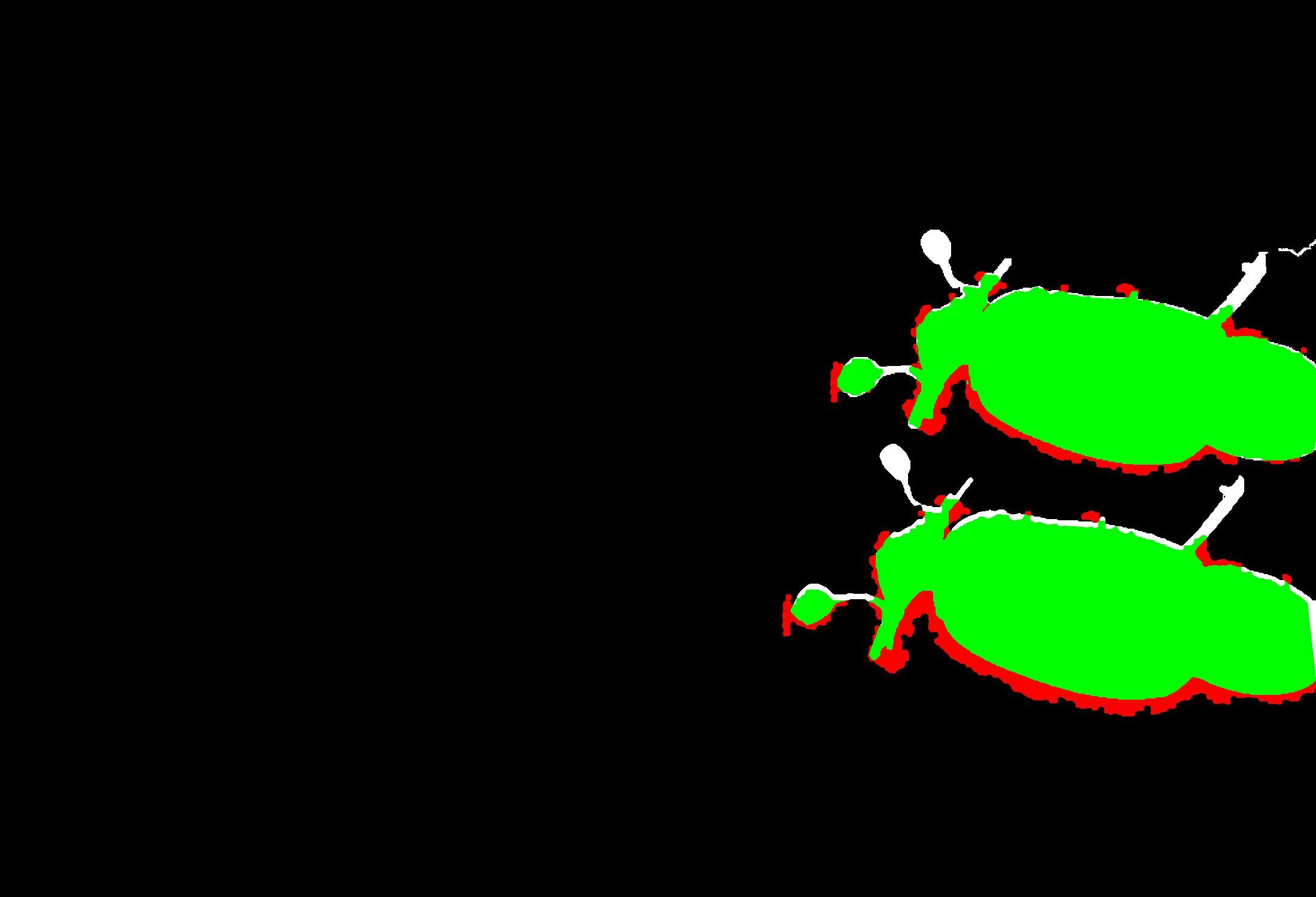} \\
		(a6) & (b6) & (c6) & (d6) & (e6) & (f7) \\
	\end{tabular}
	\caption{Several examples of copy-move forgery detection algorithms with SGO on GRIP, CMH, CoMoFoD, FAU, MICC-600 and SSRGFD-CMFD (from top to bottom). Here, (a1) $\sim$ (a6) denotes the forgery images; (b1) $\sim$ (b6) are the output by Cozzolino \cite{6}; (c1) $\sim$ (c6) are the output by Qi \cite{30}; (d1) $\sim$ (d6) are the output by Li \cite{22}; (e1) $\sim$ (e6) are the output by Wang \cite{27}; (f1) $\sim$ (f6) are the output by our proposed method. The
		\textit{TP}, \textit{FP}, and \textit{FN} are marked in green, red, and white, respectively.}
	\label{Fig6}
\end{figure*}

In this paper, six datasets are used for experiments: GRIP \cite{6}, CMH \cite{17}, CoMoFoD \cite{36}, FAU \cite{1}, MICC-600 \cite{16} and SSRGFD \cite{37}. Among them, the SSRGFD (2023) dataset only uses its original images and copy-move images, named SSRGFD-CMFD.

In terms of resolution, the GRIP dataset is identical resolution of $ 1024 \times 728 $; the CMH dataset has resolutions ranging from $ 845 \times 634 $ to $ 1296 \times 972 $; the CoMoFoD dataset maintains a consistent resolution of $ 512 \times 512 $; the FAU dataset has an approximate resolution of $ 3000 \times 2300 $; the MICC-600 dataset has resolutions that span from $ 800 \times 533 $ to $ 3888 \times 2592 $, and the SSRGFD-CMFD dataset also has resolutions that range from $ 1000 \times 1600 $ to $ 1600 \times 2800 $. In order to better handle images of various resolutions, this paper divides these images into three categories: small, medium, and large. Subsequently, the training is conducted separately for each category under the following conditions:
\begin{equation}
	\left\{ \begin{array}{l}
		{\mathop{\rm small}\nolimits} ,\max (h,w) < 1024\\
		{\mathop{\rm medium}\nolimits},1024 \le \max (h,w) < 2048\\
		{\mathop{\rm large}\nolimits} ,\max (h,w) \ge 2048
	\end{array} \right..
\end{equation}

In terms of attacks involved, they are divided into pre-processing attacks and post-processing attacks. For pre-processing attacks, GRIP and FAU only involve translation; other datasets add rotation and scaling attacks. In particular, SSRGFD contains partial flipping attacks; CoMoFoD contains partial distortion attacks. For post-processing attacks, only CoMoFoD and SSRGFD-CMFD are involved, mainly including contrast adjustment, brightness change, and color reduction.

\textbf{Experimental details on the dataset}: In this paper, GRIP, CMH, CoMoFoD-BASE, and FAU are used to \textbf{train} parameters. MICC-600 and SSRGFD-CMFD are used to \textbf{test} our scheme. Among them, CoMoFoD-BASE is composed of $ \rm{001\_O} \sim \rm{200\_O} $ and $ \rm{001\_F} \sim \rm{200\_F} $ from the CoMoFoD dataset. Since CMH dataset lacks original images, this paper combines it with the original image part of the GRIP dataset to form a new dataset CMH+GRIPori, which contains 188 images.

\subsection{Evaluation Metrics}
Three metrics are used to evaluate performances, defined as True Positive Rate (\textit{TPR}), False Positive Rate (\textit{FPR}), and \textit{$ F_{1} $} score:
\begin{equation}
	\textit{TPR} = \frac{{\textit{TP}}}{{\textit{TP} + \textit{FN}}},
\end{equation}
\begin{equation}
	\textit{FPR} = \frac{{\textit{FP}}}{{\textit{TN} + \textit{FP}}},
\end{equation}
\begin{equation}
	\textit{$ F_{1} $} = \frac{{2\textit{TP}}}{{2\textit{TP} + \textit{FN} + \textit{FP}}}
\end{equation}
where, the meanings of \textit{TP}, \textit{FP}, \textit{FN} and \textit{FP} are shown in Table \ref{Tab1}.
\begin{table}[ht]
	\caption{The meanings of \textit{TP}, \textit{FP}, \textit{FN} and \textit{FP}}\label{Tab1}
	\centering
	\begin{tabular}{|c|c|c|}\hline
		\diagbox{Actual}{Detected} & tampered & original \\ \hline
		tampered & \textit{TP} & \textit{FN} \\ \hline
		original & \textit{FP} & \textit{TN} \\ \hline
	\end{tabular}
\end{table}

\textbf{Details of the evaluation metrics}: In the development of the past few years, it has been unanimously believed that the pixel level is more important than the image level, causing various algorithms to ignore the negative impact of image-level false alarms on practical applications. In this paper, a comparison of image level performance is added. Subsequently, \textbf{image level} \textit{TPR}, \textit{FPR}, and \textit{$ F_{1} $} are denoted as \textit{TPR}, \textit{FPR}, and \textit{F-i}. \textbf{Pixel level} \textit{$ F_{1} $} will be denoted as \textit{F-p} and \textit{F-measure} according to two methods. Among them, \textit{F-p} is the result of aggregating the entire dataset. \textit{F-measure} firstly calculates the \textit{F-p} score of each image, and then reports their average score. Usually, \textit{F-measure} is used to evaluate datasets that do not contain authentic images.

\subsection{Detection Results on Datasets}
In this subsection, six datasets are evaluated separately in terms of training and testing. Several advanced methods are used for performance comparison, includings 1) \textit{block-based algorithms}, such as those proposed by Cozzolino \cite{6} (2015) and Qi \cite{30} (2022); 2) \textit{keypoint-based algorithms}, such as the methods proposed by Li \cite{22} (2018), Wang \cite{27} (2023).

\textbf{\textit{The comparison of SGO image detection examples}}. As illustrated in Fig. \ref{Fig6}, it can be seen that all forgery examples contain SGOs such as windows, railings, etc. This is a great challenge for existing algorithms. Overall, our method has significant advantages in detecting images with SGOs. This is primarily due to the excessive matching of keypoints, which allows the new iterative forgery localization algorithm to distinguish between SGOs and tampering using the minimum number of inliers. For \textit{block-based algorithms}, the methods proposed by Cozzolino \cite{6} and Qi \cite{30} almost always have multiple small false detection blocks, which indicates that the compared block-based algorithms cannot effectively handle SGO images. The \textit{keypoint-based algorithms} exhibit opposite results. Once these algorithms have false detections, large areas of false detection blocks appear. This is because these algorithms rely solely on a small number of matches to determine geometric relationships, lacking further exploration of matching information. Compared with the method proposed by Wang, Li's method, due to the training of contrast thresholds, can use stable keypoints for matching, thereby reducing false detections.

\textit{\textbf{Evaluation performance on datasets involved in training}}. Most of the existing algorithms aim to maximize the number of \textit{TP}, which often results in a higher number of \textit{FP}. This also has negative impact for forensic investigators. The \textit{FPR} metric in Table \ref{Tab2} effectively supports this claim.

\begin{table*}[htbp]
	\caption{\textit{TPR}(\%), \textit{FPR}(\%), \textit{F-i}(\%) (image level) and \textit{F-p} (\%) (pixel level) on training datasets}\label{Tab2}
	\centering
	\begin{tabular}{cccccccccc} \hline
		\multicolumn{1}{c}{\multirow{2}{*}{Methods}} & \multicolumn{4}{c}{GRIP (medium)} & & \multicolumn{4}{c}{CMH+GRIPori (medium)} \\ \cline{2-5} \cline{7-10}
		& \textit{TPR} & \textit{FPR} & \textit{F-i} & \textit{F-p} & & \textit{TPR} & \textit{FPR} & \textit{F-i} & \textit{F-p} \\ \hline
		Cozzolino \cite{6} & 98.75 & 8.75 & 95.18 & 92.99 & & 92.59 & 8.75 & 93.02 & 88.10 \\
		Qi \cite{30} & \textbf{100} & 20 & 90.91 & 91.81 & & 97.22 & 20 & 91.70 & 88.59 \\
		Li \cite{22} & \textbf{100} & \textbf{0} & \textbf{100} & 94.66 & & 96.30 & \textbf{0} & \textbf{98.11} & 90.61 \\
		Wang \cite{27} & \textbf{100} & 11.25 & 94.67 & 85.71 & & 99.07 & 11.25 & 95.54 & 91.09 \\
		Proposed & \textbf{100} & \textbf{0} & \textbf{100} & \textbf{95.57} & & \textbf{100} & \textbf{0} & \textbf{100} & \textbf{94.39} \\ \hline
		\multicolumn{1}{c}{\multirow{2}{*}{Methods}} & \multicolumn{4}{c}{CoMoFoD-BASE (small)} & & \multicolumn{4}{c}{FAU (large)} \\ \cline{2-5} \cline{7-10}
		& \textit{TPR} & \textit{FPR} & \textit{F-i} & \textit{F-p} & & \textit{TPR} & \textit{FPR} & \textit{F-i} & \textit{F-p} \\ \hline
		Cozzolino \cite{6} & 53.50 & 8 & 66.25 & 77.11 & & 97.92 & 8.33 & 94.95 & 93.79 \\
		Qi \cite{30} & 93 & 21.50 & 86.71 & 86.25 & & \textbf{100} & 14.58 & 93.20 & 87.66 \\
		Li \cite{22} & 74 & 8.50 & 81.10 & 73.79 & & \textbf{100} & \textbf{2.08} & \textbf{98.97} & \textbf{94.28} \\
		Wang \cite{27} & 92 & 26 & 84.40 & 73.95 & & \textbf{100} & 6.25 & 96.97 & 92.10 \\
		Proposed & \textbf{94.50} & \textbf{6.5} & \textbf{94.03} & \textbf{86.44} & & \textbf{100} & \textbf{2.08} & \textbf{98.97} & 88.96 \\ \hline
	\end{tabular}
\end{table*}

1) \textit{The runtime of our method}: overall, the time complexity of our proposed method is acceptable in most situations. For small images, our training in CoMoFoD-BASE shows that an average runtime of about 20 seconds; For medium images, the average runtimes for GRIP and CMH+GRIPori are 23.7 and 28.1 seconds, respectively; For large images, the runtime for FAU is approximately 43 seconds.

2) \textit{Performance evaluation}: for GRIP, Li's method \cite{22} was firstly reported to achieve the best performances in image-level metrics. In this paper, our method also employs it for training, further improving the localization metric \textit{F-p} while ensuring image-level performances. This is primarily due to the robust grayscale statistical information helping to overcome the localization of various complex scenarios. For CMH, all tampered images are detected by our method, and only one image is mislocalized due to inability to distinguish the source. No matter in terms of image-level or pixel-level metrics, it is the best result that has been reported so far. For CoMoFoD-BASE, it is evident that the methods recently proposed by Qi (2022) and Wang (2023) have made significant advancements in \textit{TPR}. This is mainly attributed to their application of a larger scaling factor, which is important for forensic tasks on low-resolution images. However, as the scaling factor increases, more unstable keypoints will be obtained. Recent methods do not deal with this aspect, resulting in higher \textit{FPR}. In contrast, our method considers the discrimination of homography sources in its design and achieves excellent performance in comparative experiments. As a large (high-resolution) dataset, tampered areas in FAU can generate keypoints very well. Therefore, the existing algorithms and our proposed method have superior image-level performance in this dataset. However, our algorithm exhibits some disadvantages in terms of \textit{F-p}. This is mainly because excessive keypoints under the G2NN test lead to matching holes in the tampered regions. These holes prevent the complete construction of suspicious regions through the scale information of inliers, resulting in degraded localization performance.

\textit{\textbf{Evaluation performance on datasets involving testing}}. 1) The MICC-600 is firstly used for testing. It is an extension of the FAU dataset, comprising 160 tampered images and 440 authentic images. Table \ref{Tab3} lists the performance of the comparative algorithms and our proposed algorithm on the MICC-600 dataset. In the comparison, our method achieves the top ranking in terms of \textit{F-i}. This success is attributed to our method significantly improving the \textit{FPR} metric while only slightly reducing the \textit{TPR}. This greatly reduces the cost for forensic investigators. Regarding the localization metric \textit{F-p}, our method achieved 90.39\%, ranking third among the compared the state-of-the-art algorithms. Considering that our proposed method improves the image-level metric \textit{F-i} by 5.66\%, the decrease of 1.41\% in the pixel-level metric \textit{F-p} is entirely acceptable. 2) The SSRGFD-CMFD dataset (2023) is used for further \textbf{generalization} testing. It contains 609 tampered images and 922 authentic images, with some of the tampered images undergoing pre-processing or post-processing attacks. The source of this dataset differs from all the datasets mentioned above. As shown in Table \ref{Tab4}, our proposed method ranks first in terms of \textit{FPR}, \textit{F-i}, and \textit{F-p}. Regarding \textit{TPR}, our algorithm scores 76.68\%, ranking second, only about 1\% lower than the highest score.

\begin{table}[htbp]
	\caption{\textit{TPR}(\%), \textit{FPR}(\%), \textit{F-i}(\%) (image level) and \textit{F-p}(\%) (pixel level) on MICC-600 datasets}\label{Tab3}
	\centering
	\begin{tabular}{ccccc} \hline
		Methods & \textit{TPR} & \textit{FPR} & \textit{F-i} & \textit{F-p} \\ \hline
		Cozzolino \cite{6} & 96.25 & 5.91 & 90.59 & 91.40 \\
		Qi \cite{30} & \textbf{98.13} & 9.09 & 87.96 & 89.30 \\
		Li \cite{22} & 97.50 & 5.68 & 91.50 & \textbf{91.80} \\
		Wang \cite{27} & 97.50 & 6.36 & 90.70 & 90.24 \\
		Proposed & 96.25 & \textbf{0.68} & \textbf{97.16} & 90.39 \\ \hline
	\end{tabular}
\end{table}

\begin{table}[htbp]
	\caption{\textit{TPR}(\%), \textit{FPR}(\%), \textit{F-i}(\%) (image level) and \textit{F-p}(\%) (pixel level) on SSRGFD-CMFD datasets}\label{Tab4}
	\centering
	\begin{tabular}{ccccc} \hline
		Methods & \textit{TPR} & \textit{FPR} & \textit{F-i} & \textit{F-p} \\ \hline
		Cozzolino \cite{6} & 73.89 & 13.02 & 76.31 & 66.26 \\
		Qi \cite{30} &  \textbf{77.67} & 17.25 & 76.23 & 65.69 \\
		Li \cite{22} &  63.22 & 4.99 & 74.04 & 56.03 \\
		Wang \cite{27} & 72.41 & 5.53 & 80.11 & 53.22 \\
		Proposed &  76.68 & \textbf{0.54} & \textbf{86.40} & \textbf{67.77} \\ \hline
	\end{tabular}
\end{table}

\subsection{Robustness Experiments}
\textbf{\textit{Robustness against pre-processing attacks}}. Twenty authentic images from the FAU dataset were used to construct images that have undergone pre-processing attacks. Their parameter settings are given by Table \ref{Tab5}.

\begin{table}[htbp]
	\caption{Parameters setting of the pre-processing attacks}\label{Tab5}
	\centering
	\begin{tabular}{ccc} \hline
		Attack & Range & Unit \\ \hline
		Rotation & 10, 30, 50, 70, 90 & Degree \\
		Scaling & 50, 70, 80, 90, 110, 120, 150, 200 & \% \\ \hline
	\end{tabular}
\end{table}

\begin{figure}[ht]
	\centering
	\begin{tabular}{cc}
		\includegraphics[width=0.45\linewidth]{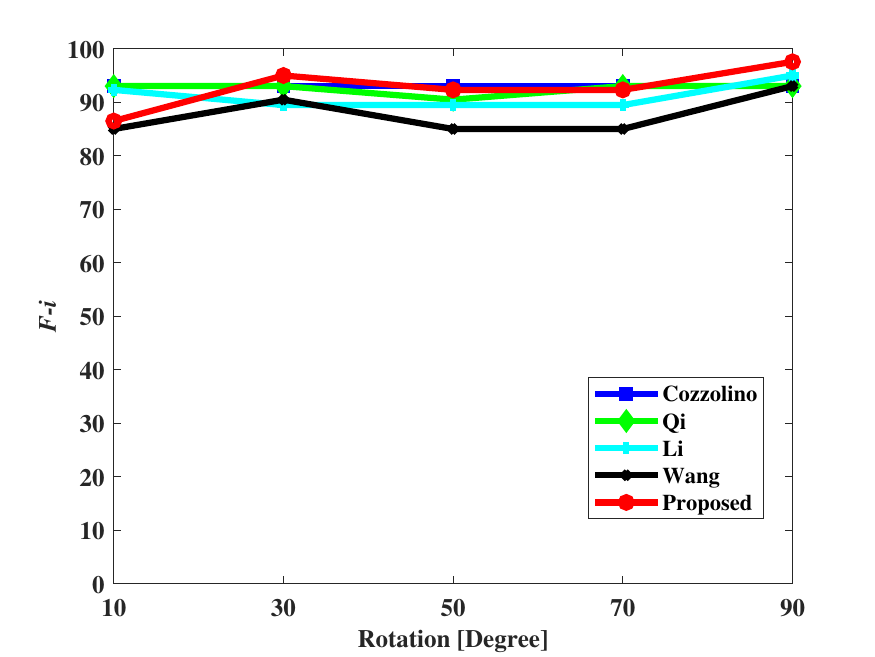} &
		\includegraphics[width=0.45\linewidth]{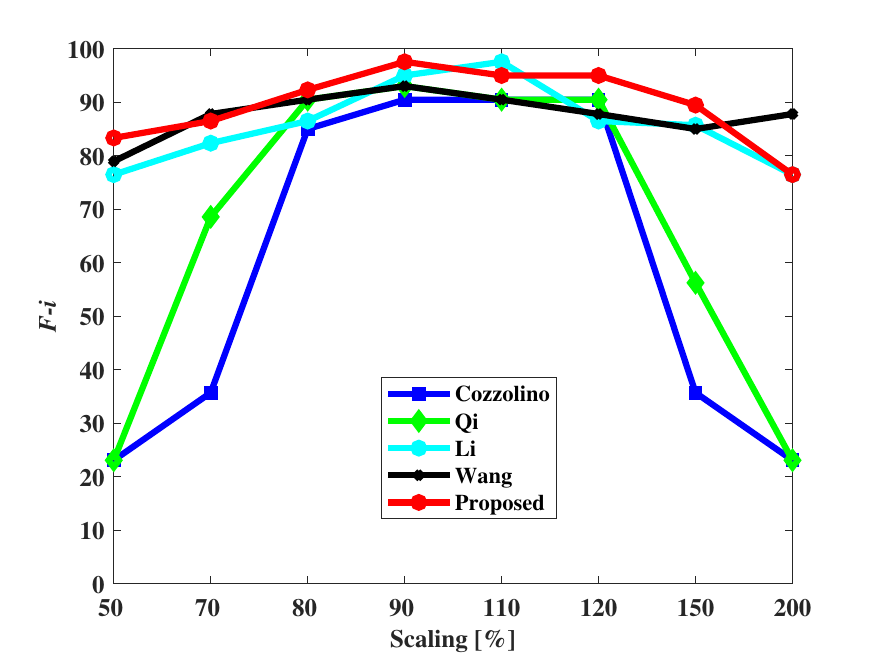} \\
		(a) & (b) \\
		\includegraphics[width=0.45\linewidth]{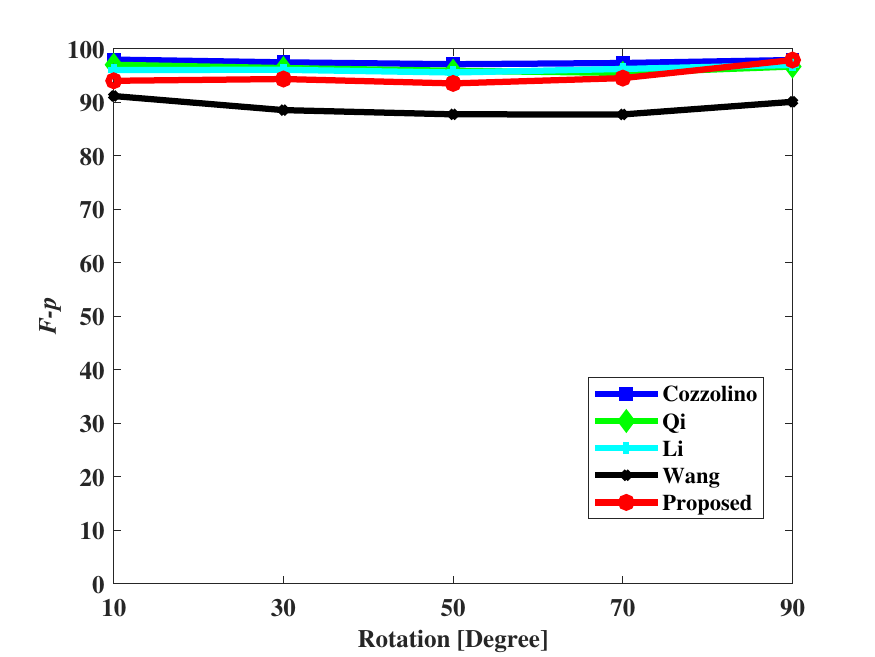} &
		\includegraphics[width=0.45\linewidth]{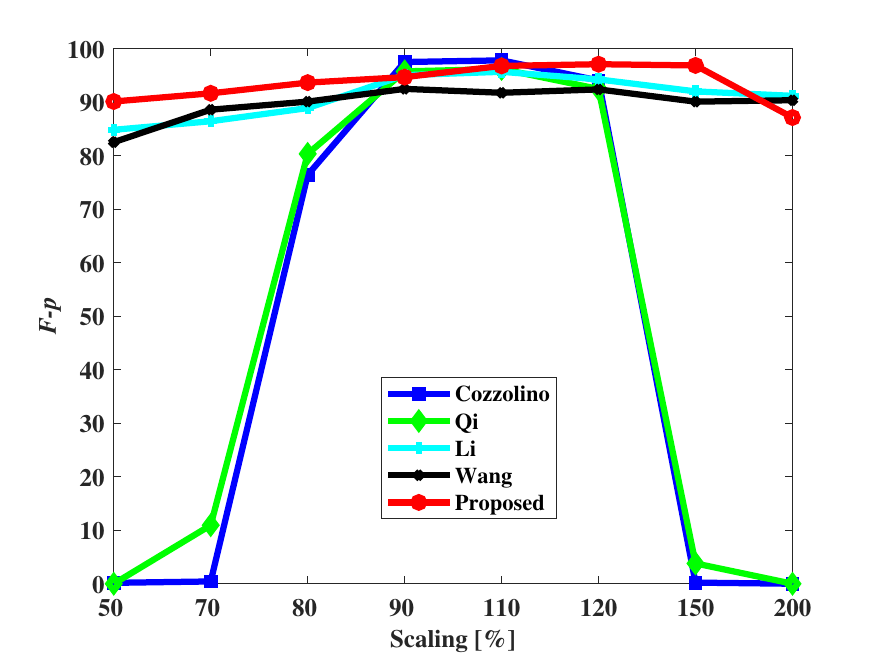} \\
		(c) & (d)
	\end{tabular}
	\caption{\textit{F-i} and \textit{F-p} curves for different algorithms. (a) \textit{F-i} curve of rotation attack; (b) \textit{F-i} curve of scaling attack; (c) \textit{F-p} curve of rotation attack; (d) \textit{F-p} curve of scaling attack.}
	\label{Fig7}
\end{figure}

The results of pre-processing attacks for competitive algorithms and our proposed method are illustrated in Fig. \ref{Fig7}. Both the\textit{ F-i} and \textit{F-p} metrics of our method are in a leading position in most cases, which is primarily attributed to the use of SIFT in our method. The method proposed by Li also ranks highly for the same reason. For the block-based algorithm proposed by Cozzolino, it leads in rotation robustness. However, due to the fixed-size windows for feature representation, it has poor scale robustness. Recently, Qi proposed Dense Invariant Representation (DIR) to enhance the scale robustness of dense features. This is why the method proposed by Qi outperforms the method proposed by Cozzolino in term of \textit{F-i}. As for the method proposed by Wang, although the Bessel-Fourier moments themselves do not possess large-scale robustness, they achieve good scale robustness through variable windows informed by the scale information of SIFT keypoints.

\textbf{\textit{Robustness against post-processing attacks}}. Recently, more and more deep learning methods have appeared in ICMFDL. They perform well against post-processing attacks. Therefore, some deep learning methods are added to test post-processing attacks, such as those proposed by Wu \cite{38} (2018), Chen \cite{39} (2020) and Xiong \cite{40} (2023). For the sources of post-processing attacks, this paper directly utilizes contrast adjustment, brightness alteration, and color reduction attacks present in CoMoFoD.

In Table \ref{Tab6}, the performances against post-processing attacks of various competitive algorithms and our proposed method are listed. It can be seen that our method has huge advantages compared with the comparative methods. There are two main reasons for this: 1) A larger scaling factor was adopted during the feature extraction stage, ensuring that there are sufficient keypoints within the tampered block. This can be demonstrated by the results of the method proposed by Qi \cite{30} and Wang \cite{27}, which typically use a larger scaling factor. 2) The identification of homography sources is considered in the forgery localization stage, which makes our method more advantageous in CoMoFoD which contains more SGO images. This is precisely the real reason why our method outperforms those proposed by Qi and Wang.

\begin{table*}[ht]
	\caption{\textit{F-measure} for different algorithms on CoMoFoD with various post-processing attack}\label{Tab6}
	\centering
	\begin{tabular}{ccccccccccccc} \hline
		\multirow{2}{*}{Method} & \multicolumn{3}{c}{Contrast Adjustment} & & \multicolumn{3}{c}{Brightness Change} & & \multicolumn{3}{c}{Color Reduction} \\ \cline{2-4} \cline{6-8} \cline{10-12}
		 & (0.01,0.95) & (0.01,0.9) & (0.01,0.8) & & (0.01,0.8) & (0.01,0.9) & (0.01,0.95) & & 32 & 64 & 128 \\ \hline
		 Cozzolino \cite{6} & 43.5 & 42.4 & 42.5 & & 39.2 & 41.9 & 41.6 & & 43.3 & 42.5 & 41.6 \\
		 Qi \cite{30} & 76.8 & 76.7 & 76.7 & & 76.2 & 74.4 & 71.2 & & 77.4 & 76.8 & 75.4 \\
		 Li \cite{22} & 49.8 & 49.2 & 49.1 & & 47.4 & 48.2 & 48.5 & & 49.8 & 48.4 & 47.9 \\
		 Wang \cite{27} & 70.6 & 73.8 & 73.3 & & 72.2 & 69.7 & 64.8 & & 72.1 & 73.9 & 69.5 \\
		 Wu \cite{38} & 49.4 & 49.4	& 49.4 & & 46.2 & 47.8 & 48.9 & & 49.2 & 48.4 & 47.9 \\
		 Chen \cite{39} & 51.6 & 52.0 & 51.5 & & 49.7 & 50.5 & 50.4 & & 51.1 & 50.8 & 51.3 \\
		 Xiong \cite{40} & 53.9 & 53.4 & 53.3 & & 51.9 & 52.6 & 52.8 & & 53.1 & 52.4 & 53.2 \\
		 Proposed & \textbf{80.6} & \textbf{80.1} & \textbf{80.4} & & \textbf{80.3} & \textbf{77.9} & \textbf{74.5} & & \textbf{81.0} & \textbf{79.6} & \textbf{75.2} \\ \hline
	\end{tabular}
\end{table*}

\subsection{Evaluation on Different Stages}
In this subsection, an evaluation of the core content of this paper will be conducted. This includes the coverage rate of keypoints in the feature extraction stage, the ablation experiments of various modules in the feature matching stage and the effectiveness of forgery localization stage.

\begin{figure}[ht]
	\centering
	\begin{tabular}{c}
		\includegraphics[width=0.5\linewidth]{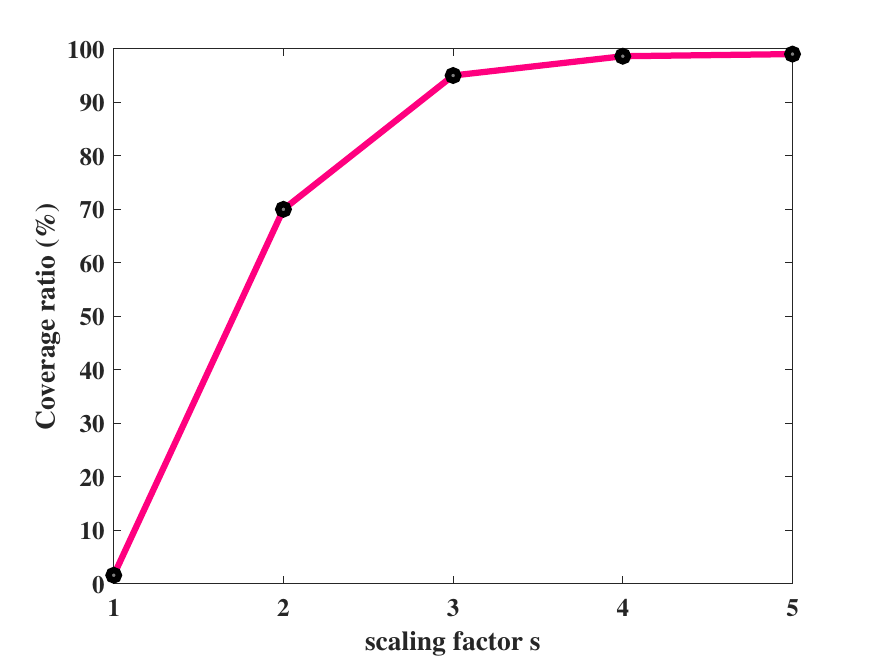}
	\end{tabular}
	\caption{The coverage rate of pixels that meet our suggestion.}
	\label{Fig8}
\end{figure}

\textit{\textbf{Evaluation the coverage rate of keypoints in the feature extraction stage}}. In Section \ref{proposed method1}, this paper suggests that the excessive keypoint strategy should have at least 4 keypoints within a $ 16 \times 16 $ patch. As illustrated in Fig. \ref{Fig8}, the coverage rate of pixels that meet this suggestion under different scaling factor $ s $ is provided. Here, experiments are conducted on GRIP, which includes smooth images. All images are converted to grayscale. As the scaling factor increases, the coverage rate continuously increases. When $ s > 4 $, almost no increase in coverage rate is observed. Therefore, this paper commonly adopts $ s = 4 $.

\textit{\textbf{The ablation experiments of the feature matching stage}}. 
As shown in Table \ref{Tab7}, ablation experiments using various modules are conducted during the feature matching stage. It includes the overlapped gray level clustering (G) \cite{22}, the overlapped entropy clustering (E), and LG. Additionally, Brute Force Matching (BFM) is used for comparison. Here, 20 tampered images from the GRIP dataset are used, and the average number of keypoints reaches 99084. It can be seen that compared to BFM, the use of these three modules significantly reduces the time complexity. In particular, LG has the lowest time complexity. However, while the LG module reduces the time complexity, the accuracy also decreases significantly, which will have a negative impact on the subsequent stages. To overcome this problem, G+E (grayscale-entropy clustering) is introduced. The ablation experiment results show that the concatenation of G and E can further reduce time complexity while maintaining accuracy. This indicates that grayscale-entropy clustering enhances the similarity within each group. Therefore, using G+E+LG can ensure a balance between efficiency and accuracy, which is well demonstrated by our ablation experiments.

\begin{table}[htbp]
	\caption{The ablation experiments of various modules in the feature matching stage}\label{Tab7}
	\centering
	\begin{tabular}{ccc} \hline
		Method & \# correct matches & time \\ \hline
		BFM & 1156 & 545.2s \\
		G & 1178 & 155.6s \\
		E & 1163 & 176.1s \\
		LG & 818 & 7.9s \\
		G+E & 1172 & 78.1s \\
		G+E+LG (Proposed) & 1086 & 14.8s \\ \hline		
	\end{tabular}
	\end{table}

\textit{\textbf{The effectiveness of forgery localization stage}}. To evalute the new iterative forgery localization, the currently advanced methods proposed by Li \cite{22} and Wang \cite{27} are compared. Two sets of experiments are conducted: 1) \textit{Evaluate localization performance}. 80 tampered images from the GRIP dataset are used. In this set of experiments, all images involve only one-to-one copy-move, and all tampering is correctly detected. Therefore, this performance represents the localization performance of various algorithm. From Table \ref{Tab8}, it can be seen that our localization performance is excellent in both \textit{F-measure} and worst case \textit{F-p}. This is mainly attributed to the robust grayscale statistics verification under the excessive keypoint strategy, which overcomes the greedy growth of false positives. 2) \textit{Evaluate the performance of one-to-many copy-move forgery detection}. Twenty tampered images with one-to-many copy-move are established using CoMoFoD, which are named as OtMFD. These images can be accessed through our code. The results of one-to-many forgery detection are listed in Table \ref{Tab9}. Compared with the method proposed by Li (the input is adjusted to $ 1024 \times 1024 $), our method has a huge advantage in \textit{F-measure}, which is mainly attributed to the fact that the selection and removal of the same homography transformation overcomes the separation problem of models that are close to each other. Compared with the method proposed by Wang, our method has an absolute advantage in time complexity, which is mainly attributed to the fact that our method avoids complex multiplication operations during the verification process.

\begin{table}[htbp]
	\caption{Evaluation localization performance}\label{Tab8}
	\centering
	\begin{tabular}{cccc} \hline
		Method & \textit{F-measure} & Max. \textit{F-p} & Min. \textit{F-p} \\ \hline
		Li \cite{22} & 92.22  & \textbf{98.23} & 71.76 \\
		Wang \cite{27} & 90.86 & 98.01 & 28.04 \\
		Proposed & \textbf{94.71} & 97.72 & \textbf{81.74} \\ \hline
	\end{tabular}
\end{table}

\begin{table}[htbp]
	\caption{Evaluation the performance of one-to-many copy-move forgery detection}\label{Tab9}
	\centering
	\begin{tabular}{ccccc} \hline
		Method & \textit{F-measure} & Max. \textit{F-p} & Min. \textit{F-p} & time \\ \hline
		Li \cite{22} & 79.96 & 94.63 & 58.08 & 2.82s \\
		Wang \cite{27} & 89.68 & \textbf{98.91} & 45.08 & 62.7s \\
		Proposed & \textbf{92.70} & 98.28 & \textbf{74.35} & \textbf{1.4s} \\ \hline
	\end{tabular}
\end{table}

\section{Conclusion}\label{conclusion}
To avoid missed detections and false alarms, this paper investigates three stages of the ICMFDL. In the feature extraction stage, the limit of keypoint-based algorithms is analyzed, and an excessive keypoint strategy is proposed. This strategy is central to reducing missed detections. In the feature matching stage, the calculation amount of excessive keypoint matching is reduced through grouping. The core of this stage is to ensure the balance between accuracy and efficiency. In the forgery localization stage, a new iterative forgery localization algorithm is improved. The performance degradation problem of locating one-to-many copy-move forgeries in the iterative forgery localization \cite{22} is addressed by adopting density-based sampling and inlier removal strategy. Subsequently, the minimum number of inliers is used to distinguish SGO and tampering. It greatly reduces the false alarm rate. Finally, a verification strategy based on robust grayscale statistical information is proposed, further leveraging excessive matching information to better adapt to various complex scenarios. Extensive experimental results have been provided to demonstrate the superior performance of our proposed scheme.


\bibliographystyle{ieeetr}
\bibliography{References}

\vspace{11pt}

\vfill

\end{document}